\documentclass{article}

\usepackage{microtype}
\usepackage{longtable}
\usepackage{multirow}
\usepackage{graphicx}
\usepackage{subcaption}
\usepackage{booktabs} 

\usepackage{hyperref}

\newcommand*\diff{\mathop{}\!\mathrm{d}}

\usepackage[accepted]{icml2024}

\usepackage{amsmath}
\usepackage{amssymb}
\usepackage{mathtools}
\usepackage{amsthm}
\usepackage{bm}

\usepackage[capitalize]{cleveref}

\theoremstyle{plain}
\newtheorem{theorem}{Theorem}[section]
\newtheorem{proposition}[theorem]{Proposition}

\theoremstyle{definition}
\newtheorem{definition}[theorem]{Definition}

\theoremstyle{remark}

\newsavebox{\mybox}
\newenvironment{conjecture}
{\begin{lrbox}{\mybox}\begin{minipage}{0.98\columnwidth} \textbf{Insight:}}
{\end{minipage}\end{lrbox}\fbox{\usebox{\mybox}}}

\usepackage[textsize=tiny]{todonotes}

\icmltitlerunning{Connecting the Dots: Feasible Sample-Based Inference in Bayesian Neural Networks}

\begin{document}

\twocolumn[
\icmltitle{Connecting the Dots: Is Mode-Connectedness the Key to Feasible \\Sample-Based Inference in Bayesian Neural Networks?}



\icmlsetsymbol{equal}{*}

\begin{icmlauthorlist}
\icmlauthor{Emanuel Sommer}{equal,lmu,mcml}
\icmlauthor{Lisa Wimmer}{equal,lmu,mcml}
\icmlauthor{Theodore Papamarkou}{uom}
\icmlauthor{Ludwig Bothmann}{lmu,mcml}
\icmlauthor{Bernd Bischl}{lmu,mcml}
\icmlauthor{David R\"ugamer}{lmu,mcml}
\end{icmlauthorlist}

\icmlaffiliation{lmu}{Department of Statistics, LMU Munich, Munich, Germany}
\icmlaffiliation{mcml}{Munich Center for Machine Learning, Munich, Germany}
\icmlaffiliation{uom}{Department of Mathematics, The University of Manchester, Manchester, UK}

\icmlcorrespondingauthor{David R\"ugamer}{david@stat.uni-muenchen.de}

\icmlkeywords{Machine Learning, ICML}
\vskip 0.3in
]



\printAffiliationsAndNotice{\icmlEqualContribution}
\begin{abstract}
A major challenge in sample-based inference (SBI) for Bayesian neural networks is the size and structure of the networks' parameter space.
Our work shows that successful SBI is possible by embracing the characteristic relationship between weight and function space, uncovering a systematic link between overparameterization and the difficulty of the sampling problem.
Through extensive experiments, we establish practical guidelines for sampling and convergence diagnosis. 
As a result, we present a deep ensemble initialized approach as an effective solution with competitive performance and uncertainty quantification.
\end{abstract}

\section{Introduction}

Bayesian neural networks (BNNs) represent a principled solution to the problem of probabilistic deep learning.
In the absence of analytically tractable solutions, Bayesians traditionally rely on sample-based inference (SBI) as it can in theory recover the true posterior---up to a Monte Carlo error---and requires no (restrictive) assumptions on the posterior distributional family.
Consequently, SBI has enormous potential for uncertainty quantification in BNNs \citep{farquhar_2020_LibertyDeptha, izmailov_2021_WhatArea,ecml}.
However, recent research has focused on local approximations that may not fully capture the multimodality of BNN posteriors, as noted in \citet{alexos_2022_StructuredStochastic, arbel_2023_PrimerBayesiana}. Such approaches potentially overlook significant portions of the posterior density, precluding a comprehensive quantification of uncertainty.
The main reason behind the reluctance to adopt SBI appears to be its perceived computational demands, as discussed in  \citet{papamarkou_2022_ChallengesMarkova, sharma_2023_BayesianNeural}.
Indeed, \citet{izmailov_2021_WhatArea} show that SBI is possible for large state-of-the-art (SOTA) networks but comes at a high cost. 
Yet, scalable software solutions \citep[e.g., JAX,][]{jax2018github} and methodological adaptations \citep{nemeth2021stochastic} have made SBI increasingly accessible. 

While technological progress continues unabated, work studying the peculiarities of SBI remains limited.
One reason for this research gap, and possibly the most fundamental problem for BNNs, is their large number of parameters. 
The resulting challenges are uncharted territory for established Bayesian workflow routines such as the sampling strategy, the choice of prior, and convergence monitoring. 
It is tempting to conclude that parameter space inference should simply be discarded in favor of addressing uncertainties directly in the function space \citep[see, e.g.,][]{tran2022all}, but this perspective risks suspending valuable insights.

\begin{figure}[!t]
    \centering
    \includegraphics[width = 0.91\columnwidth]{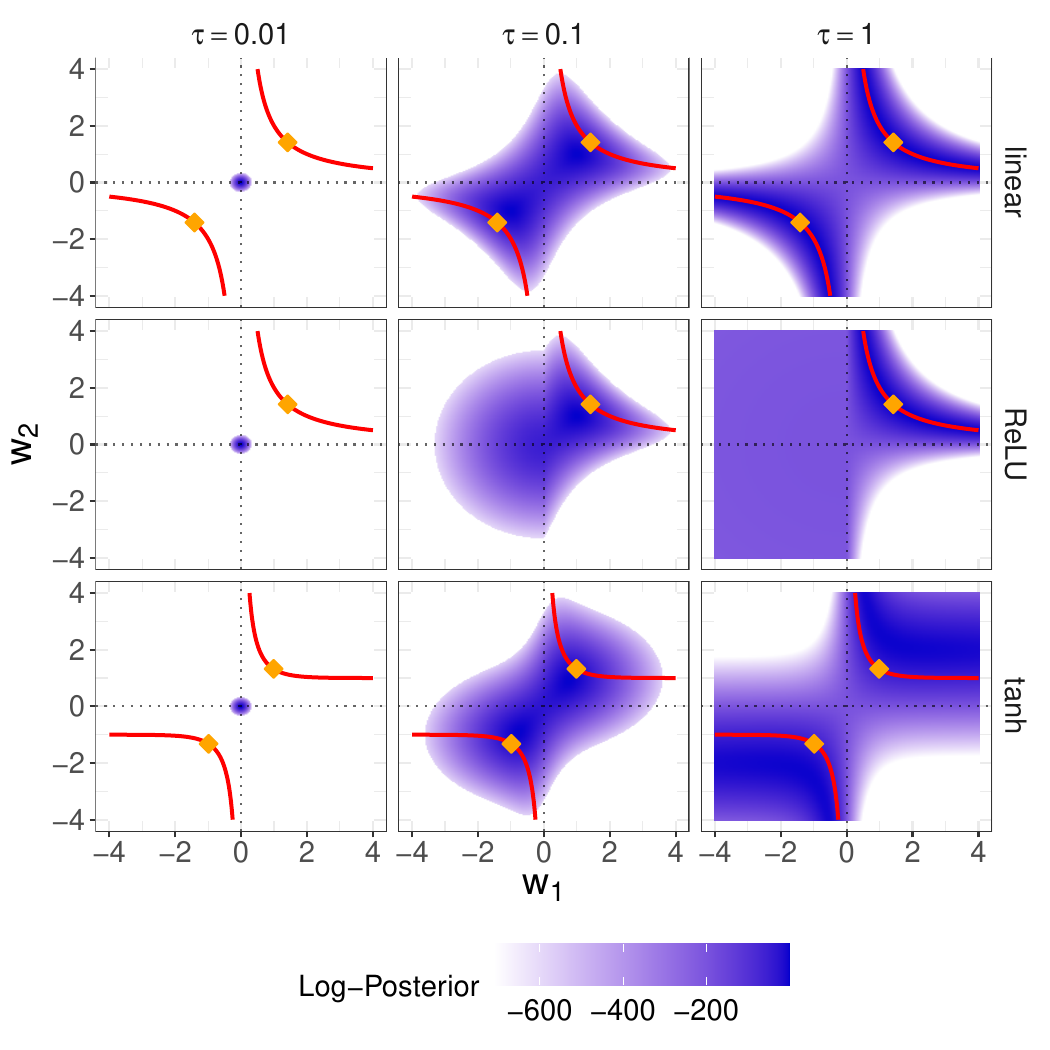}
    \vskip -0.15in
    \caption{Log-posterior of a 1-1-1 neural network for different activation functions $h$ (rows) using a varying $\mathcal{N}(0,\tau)$ prior (columns). 
    \textit{Red lines:} set of maximum likelihood solutions for the non-identifiable solution of $h(w_1 x) \cdot w_2$ with $x=1$. 
    \textit{Orange diamonds:} minimum-norm solutions among the maximum likelihood set.}
    \label{fig:1-1-1}
    \vskip -0.15in
\end{figure}

In fact, the complex posterior topology contains structures that can be unraveled and accounted for. Most notably, overparameterization in BNNs leads to many symmetrical modes in the posterior density \citep{chen_1993_GeometryFeedforwarda}.
It is less clear whether these modes are ``connected'' in the parameter space.
\citet{yao_2022_StackingNonmixing} point out that BNN posterior surfaces contain many high-density regions separated by basins of low probability, making it difficult for samplers to traverse between modes in finite time.
As \cref{fig:1-1-1} shows, high-density regions of the posterior can behave quite differently depending on the influence of the prior. 
Compared to the maximum likelihood solutions (red lines), priors can induce unfavorable starting points (left column) and prevent the sampler from reaching parameter values of high predictive capacity. 
On the other hand, priors are able to link seemingly disconnected symmetry regions (center and right column).
This observation highlights a fundamental aspect of the Bayesian paradigm:
regions like the non-activated areas of ReLU networks (center row, left quadrants), which do not affect the optimization objective but are relevant to posterior uncertainty, should attain a positive probability of being sampled. 
In summary, the connectedness of posterior modes---and thus the difficulty of the sampling problem---strongly depends on the interplay between prior information, data, and overparameterization.

\textbf{Our Contributions}. 
Among the discussions about the merits of parameter space inference (see also~\cref{app:discussweight}), our paper sheds light on how SBI can be successfully leveraged for BNNs.
We argue that accounting for the idiosyncratic relationship between weight and function space is key to affordable SBI. 
From a broad range of experiments, we obtain insights into SBI's successes and failures. 
Key findings include the \emph{importance of chain initialization} and the \emph{increasing connectedness of modes in deeper layers}. 
Moving beyond classical measures unfit to assess BNN inference, we develop novel \emph{convergence diagnostics that incorporate heterogeneity of layer-specific variances}.
As a result, our work can be seen as an extension of recent investigations into BNN symmetries \citep{ecml}, prior influence \citep{fortuin2022bayesian}, and convergence diagnostics \citep{vetharirhat2021}. 
Our investigations reveal what BNN posteriors demand from practical SBI, leading us to propose \emph{Deep Ensemble Initialized MCMC (DEI-MCMC)} as a straightforward approach.
DEI-MCMC achieves competitive performance and uncertainty quantification in a number of tasks.

\section{Background and Related Literature}


We focus on multi-layer perceptrons (MLPs). 
Let $f: \mathcal{X} \to \mathcal{Y}$, $\mathcal{X} \subseteq \mathbb{R}^p, \mathcal{Y} \subseteq \mathbb{R}^m$, represent a fully-connected network. 
We collect all weights and biases in $\bm{\theta} \in \Theta \subseteq \mathbb{R}^d$.
Under the Bayesian paradigm, $\bm{\theta}$ is a random variable that admits a prior density $p(\bm{\theta})$.
Updating the prior with evidence from the observed data $\mathcal{D} \in (\mathcal{X} \times \mathcal{Y})^n$ leads to a posterior density 
    $p(\bm{\theta} | \mathcal{D}) = p(\mathcal{D} | \bm{\theta}) p(\bm{\theta}) / p(\mathcal{D})$,
allowing us to express epistemic uncertainty about $\bm{\theta}$.
This epistemic uncertainty governs the posterior predictive density (PPD) over the label $\bm{y}^\ast$ for a new observation $\bm{x}^\ast \in \mathcal{X}$: 
\begin{equation} \label{eq:ppd}
    p(\bm{y}^\ast | \bm{x}^\ast, \mathcal{D}) = \int_\Theta p(\bm{y}^\ast | \bm{x}^\ast, \bm{\theta}) p(\bm{\theta} | \mathcal{D}) \diff\bm{\theta}.
\end{equation}
The dispersion of $p(\bm{y}^\ast | \bm{x}^\ast, \mathcal{D})$ can be used to quantify predictive uncertainty about $\bm{y}^\ast$ \citep[see, e.g.,][]{murphyProbabilisticMachineLearning2022}.
Since the integral in~\cref{eq:ppd} usually lacks a closed-form solution, we rely on Monte Carlo estimates of the form
\begin{equation} \label{eq:mc_integral}
    p(\bm{y}^\ast | \bm{x}^\ast, \mathcal{D}) \approx \frac{1}{S} \sum_{s = 1}^S p\left(\bm{y}^\ast | \bm{x}^\ast, \bm{\theta}^{(s)}\right),
\end{equation}
with $S$ posterior samples $\bm{\theta}^{(s)} \sim p(\bm{\theta} | \mathcal{D})$ \citep{andrieu_2003_IntroductionMCMC}.
Markov chain Monte Carlo (MCMC) methods, the workhorse of SBI, construct a Markov chain whose stationary distribution is the posterior density, meaning that samples $\bm{\theta}^{(s)}$ from the chain are samples from $p(\bm{\theta} | \mathcal{D})$ after the chain has converged \citep{gelman_2013_BayesianData}.

\subsection{Multimodality in the Posterior Landscape}

The non-identifiability of deep neural networks (DNNs) makes inference highly challenging \citep{wei_2023_DeepLearning}.
The weight space contains multiple \emph{equioutput} states, i.e., different parameter vectors leading to the same functional mapping \citep{hecht-nielsen_1990_ALGEBRAICSTRUCTUREa}.
This phenomenon generates \emph{symmetries}\footnote{
We focus on what \citet{villar_2023_FullyCovariant} call \emph{passive symmetries}, arising purely from modeling choices.
In contrast, a separate field of research studies \emph{active symmetries}, typically in the quest of making functions equivariant to symmetric properties of the physical world \citep[e.g.,][]{cohen_2014_LearningIrreducible}.
} in the BNN posterior: given equal prior probability, all equioutput parameters have equal posterior density, which leads to strong multimodality of the posterior
\citep{ecml}.
Being equioutput induces an equivalence relation \citep[e.g.,][]{kurkova_1994_FunctionallyEquivalenta}: 
$$\bm{\theta} \sim \bm{\theta}^\prime \Longleftrightarrow f_{\bm{\theta}}(\bm{x}) = f_{\bm{\theta}^\prime}(\bm{x}) \,\forall\,\bm{x}\in\mathcal{X}, ~~ \bm{\theta}, \bm{\theta}^\prime \in \Theta.$$

\begin{definition}[\textbf{Equioutput parameters}] \label{def:eqioutput_params}
    We say that two parameters $\bm{\theta}, \bm{\theta}^\prime \in \Theta$ are equioutput iff $\bm{\theta} \sim \bm{\theta}^\prime$. 
\end{definition}
Equioutput parameters are related by homeomorphisms $\mathcal{F}: \Theta \rightarrow \Theta$ \citep{grigsby_2023_HiddenSymmetriesa}.
Depending on the network architecture, there is a huge number of (nontrivial) transformations $\mathcal{F}$ that preserve the input-output mapping, mainly arising from neuron permutability in hidden layers and certain activation functions \citep{chen_1993_GeometryFeedforwarda, kunin_2021_NeuralMechanics}.
Symmetries have been studied at length for DNNs optimized by empirical risk minimization (ERM).
Building on pioneering work on MLPs with odd activation functions \citep{sussmann_1992_UniquenessWeightsa, chen_1993_GeometryFeedforwarda, albertini_1994_UniquenessWeightsa, kurkova_1994_FunctionallyEquivalenta}, the rise of ReLU sparked follow-up research for homogeneous activations \citep[e.g.,][]{neyshabur_2015_PathSGDPathNormalizeda, freeman_2017_TopologyGeometrya, bona-pellissier_2021_ParameterIdentifiability, grigsby_2023_HiddenSymmetriesa}.
More recently, symmetries have been rediscovered in efforts to understand the learning dynamics of DNN optimization \citep{brea_2019_WeightspaceSymmetryb, ainsworth_2023_GitReBasin, entezari_2022_RolePermutationa}, where the closely related phenomenon of overparameterization---enabling equioutput states in the first place---plays an important role \citep{simsek_2021_GeometryLossa, bubeck_2022_UniversalLaw}.
Interestingly, symmetries have scarcely been discussed from a Bayesian viewpoint \citep[with a few notable exceptions, e.g.,][]{moore_2016_SymmetrizedVariationala, pourzanjani_2017_ImprovingIdentifiabilityb}.
Since there is a well-established link between ERM with regularized objectives and Bayesian inference,
the above phenomena translate from \textit{loss} to \textit{posterior} landscapes \citep[e.g.,][]{mackay1992_bayesBackprop}.
In particular, we consider the following notion:
\begin{definition}[\textbf{Posterior symmetry}] \label{def:posterior_symm}
    The posterior density exhibits a symmetry w.r.t. $ \bm{\theta}, \bm{\theta}^\prime \in \Theta$ iff $p(\bm{\theta} | \mathcal{D}) = p(\bm{\theta}^\prime | \mathcal{D})$.
\end{definition}
\begin{proposition} \label{lem:eqioutput_symm}
    For two parameters $\bm{\theta}, \bm{\theta}^\prime \in \Theta$ with $p(\bm{\theta}) = p(\bm{\theta}^\prime)$, being equioutput implies a posterior symmetry, et vice versa: $\bm{\theta} \sim \bm{\theta}^\prime \Longleftrightarrow p(\bm{\theta} | \mathcal{D}) = p(\bm{\theta}^\prime | \mathcal{D}).$
\end{proposition}
\cref{lem:eqioutput_symm} follows immediately from Bayes' theorem and the fact that equioutput parameters share the same likelihood by definition.

\subsection{Mode Connectivity}
The existence of multiple modes in the posterior (or loss) landscape begs the question of how inference (or optimization) is affected.
Loosely speaking, a multimodal surface with strong curvature makes inference harder because the sampler will take a long time to travel between modes through areas of low posterior probability, or even infinitely long in the worst case of wholly disconnected modes.
The same intuition holds for standard gradient-based optimizers that risk getting stuck in isolated local optima.
As a consequence, the phenomenon of \emph{mode connectivity} \citep{garipov_2018_LossSurfacesb} has received increased attention in the past few years.
Fortunately, local optima generally seem to be connected by low-loss areas for sufficiently benign curvature of the loss hypersurface \citep{
kuditipudi_2019_ExplainingLandscape, pittorino_2022_DeepNetworksa, sharma_2023_SimultaneousLinear, farrugia-roberts_2023_FunctionalEquivalencea}.
\citet{draxler_2018_EssentiallyNo} even conjectured that there might in fact be a single manifold containing all optima.
However, it is crucial to understand that connectivity is a relative notion in three regards.
First, modes will rarely be disconnected in the actual sense of the word, meaning that areas between them might be assigned very low, but not zero, posterior probability.
Second, as discussed in the introductory case of \cref{fig:1-1-1}, the curvature of the landscape is modulated by the relationship between prior belief \citep[or inductive biases; e.g.,][]{garipov_2018_LossSurfacesb} and observed data \citep[e.g.,][]{draxler_2018_EssentiallyNo}.
Weak priors together with a small amount of training data, for instance, will produce a rather smooth surface with less pronounced modes in absence of strong evidence for any particular hypothesis.
Moreover, \citet{ainsworth_2023_GitReBasin} suggest that mode connectivity is inextricably tied to the implicit regularization of gradient-based optimizers.
Third, and perhaps most importantly for the present work, theoretical mode connectivity is not all that matters.
The behavior and mobility of a sampler depend directly on its design (e.g., the propensity of accepting proposals for sampling locations).
If the sampler fails to traverse low-density areas, modes become effectively disconnected even if the theory suggests the existence of a non-zero-probability link between them. 
In addition to the data, inductive biases and regularization, algorithmic uncertainty of the sampling process thus also plays a role. 

\subsection{Inference Methods}

\textbf{Ensemble Approaches.}
To account for multiple basins of attraction in the posterior landscape, several inference methods using \emph{ensembles} have been proposed.
Inspired by explicit ensembling of DNNs \citep{lakshminarayanan_2017_SimpleScalablea}, approaches grounded more firmly in a Bayesian mindset include modifications of the forward pass to (asymptotically) obtain samples from the Gaussian process posterior \citep{he_2020_BayesianDeep} and ensembles of Gaussian posterior approximators \citep[MultiSWAG;][]{wilson_2020_BayesianDeepa}.

\textbf{Sampling-Based Inference.}
Research on SBI, on the other hand, is largely disregarding symmetries.
Hamiltonian Monte Carlo \citep[HMC;][]{neal_2011_MCMCUsing}, struggling with multimodality just like any other MCMC method, is the \emph{de facto} gold standard \citep{farquhar_2020_LibertyDeptha, izmailov_2021_WhatArea}.
Combinations of SBI with classical optimization elements have been proposed for scalability, such as 
stochastic-gradient MCMC variants \citep{mandt_2017_StochasticGradientb, zhang_2020_CyclicalStochastic, cobb_2021_ScalingHamiltonian}, minibatch MCMC with blockwise sampling \citep{papamarkou_2023_ApproximateBlocked}, subsampling of likelihoods \cite{goan_2023_PiecewiseDeterministic}, or MCMC boosted with momentum-based information \citep{bieringer_2023_AdamMCMCCombining}.
A separate strand of research attempts to find more informative priors \citep{fortuin2022bayesian, kim_2023_InverseReferencePriors}.

However, as of yet, truly feasible SBI for BNNs is still lacking.
Setting out to change that, we study in great detail the state of SBI---with particular attention on the symmetry-induced multimodality of the posterior---and derive a practical solution that addresses the uncovered issues.

\newpage

\section{Experimental Setup} \label{sec:expset}

\begin{table}[!t]
    \centering
    \small
    \setlength{\tabcolsep}{2pt}
        \caption{Average RMSE for different models over different data sets. All neural networks have two hidden layers with 16 neurons each. The best method per data set is highlighted in bold.}
    \label{tab:rmse_bench}
    \vskip 0.15in
       \begin{center}
    \begin{sc}
    \resizebox{0.85\columnwidth}{!}{
    \begin{tabular}{lrrrrrr}
        \toprule
        \multicolumn{1}{l}{data set} & LM & RF & DNN & DE & RS & BNN \\ 
        \midrule\addlinespace[2.5pt]
airfoil & 0.716 & 0.255 & 0.252 & 0.239 & 0.250 & \textbf{0.182} \\
bikesharing & 0.790 & \textbf{0.231} & 0.374 & 0.365 & 0.362 & 0.253 \\
concrete & 0.630 & 0.304 & 0.317 & 0.282 & 0.554 & \textbf{0.258} \\
energy & 0.274 & 0.050 & 0.048 & 0.043 & 0.062 & \textbf{0.037} \\
protein & 0.863 & \textbf{0.581} & 0.804 & 0.803 & 1.077 & 0.716 \\
yacht & 0.612 & 0.072 & 0.108 & 0.103 & 0.032 & \textbf{0.022} \\
        \bottomrule
    \end{tabular}
    }
    \end{sc}
    \end{center}
\vskip -0.3 in
\end{table}

In Sections~\ref{sec:general}--\ref{sec:dying_bde}, we highlight findings from an extensive grid of experiments with more details and results in~\cref{app:additional_exp}.
Our code is available  \href{https://github.com/EmanuelSommer/bnn_connecting_the_dots}{under this URL}.
~\cref{sec:general} examines the general feasibility of SBI, focusing on whether SBI can achieve SOTA performance.~\cref{sec:multimodality} investigates how the multimodality of posteriors influences SBI performance, after which~\cref{sec:practicalsbi} discusses ways of handling multimodality to make SBI more practical.~\cref{sec:dying_bde} summarizes the findings and proposes DEI-MCMC as a competitive solutions for SBI in BNNs.

\textbf{Data sets and models}. We consider several regression data sets from the UCI benchmark \citep{Dua.2019}: \texttt{airfoil}, \texttt{concrete}, \texttt{energy} and \texttt{yacht}, as well as two larger data sets, \texttt{bikesharing} and \texttt{protein}.
The MLPs considered vary in width (up to 64 hidden units) and depth (up to seven layers). We investigate not only differences in model architecture, the activation function and the weight priors, but also in the choice of the sampler and its configuration.

\textbf{Sampling}. In particular, we employ HMC and the No-U-Turn Sampler \citep[NUTS;][]{hoffman_2014_NoUTurnSampler}.
The latter implements HMC with auto-tuning of the trajectory length and step size, promising a considerable speed-up over optimizing these critical hyperparameters separately. In all experiments, we run up to 12 chains with 8,000 samples each for the smaller and 4,000 samples for the larger data sets. We use a fixed target acceptance probability of 0.8 and warmup phases of 10,000 steps unless stated otherwise. While we also examined the effect of a longer warmup by running up to 100,000 steps, this did not improve performance in almost all cases.

\textbf{Performance metrics}. BNN predictions are computed via (empirical) Bayesian model averaging. 
In particular, each posterior sample induces a model from which we obtain the parameterization of a PPD conditioned on a given test observation from a test set $\mathcal{D}_{\text{test}} \in (\mathcal{X} \times \mathcal{Y})^{n_{\text{test}}}$ (20\% of the data in all experiments).
As a point estimate, we compute the expectation over those conditional distributions via Monte Carlo integration. 
Since we focus on regression, we use the test root mean squared error (RMSE) to assess predictive performance.
The quality of uncertainty quantification can be measured by the log-PPD (LPPD; ~\cref{eq:lppd} in~\cref{app:conv_diag}) on test data\footnote{
Meaningful inference should yield (L)PPDs under which the true label $\bm{y}^\ast$ has high posterior density.
Intuitively, this can only be the case for appropriate PPD dispersion (too concentrated---$\bm{y}^\ast$ will frequently not coincide with the PPD mode and thus have low density; too diffuse---no value is assigned high density).
}. We also assess the calibration of the samples by comparing nominal and empirical coverage of credibility intervals of the PPD.




\section{General Feasibility} \label{sec:general}
\begin{table}[!t]
    \centering
    \small
    \setlength{\tabcolsep}{2pt}
        \caption{Proportion of chains with RMSE performance better than an LM. Each proportion is the average of 72 experiments with 3 different train-test splits, either 1,000 or 10,000 warmup iterations, and 12 chains each with 8,000 posterior samples.}
    \label{tab:prop_blm}
    \begin{center}
    \begin{sc}
    \resizebox{1.0\columnwidth}{!}{
    \begin{tabular}{lrrrrrrrrrr}
        \toprule
        \multicolumn{1}{l}{} & \multicolumn{2}{c}{2} & \multicolumn{2}{c}{8} & \multicolumn{2}{c}{16-16} & \multicolumn{2}{c}{64} & \multicolumn{2}{c}{32-32-32} \\ 
        \cmidrule(lr){2-3} \cmidrule(lr){4-5} \cmidrule(lr){6-7} \cmidrule(lr){8-9} \cmidrule(lr){10-11}
        \multicolumn{1}{l}{data set} & relu & tanh & relu & tanh & relu & tanh & relu & tanh & relu & tanh \\ 
        \midrule\addlinespace[2.5pt]
        airfoil & $0.97$ & \bm{$1.00$} & $0.89$ & \bm{$1.00$} & $0.14$ & \bm{$0.97$} & $0.36$ & \bm{$0.64$} & $0.00$ & \bm{$0.67$} \\ 
        concrete & $0.89$ & \bm{$1.00$} & $0.69$ & \bm{$1.00$} & $0.00$ & \bm{$0.92$} & $0.08$ & \bm{$0.56$} & $0.00$ & \bm{$0.31$} \\ 
        energy & $0.65$ & \bm{$0.88$} & $0.88$ & \bm{$1.00$} & $0.06$ & \bm{$0.97$} & $0.17$ & \bm{$0.53$} & $0.00$ & \bm{$0.39$} \\ 
        yacht & $0.75$ & \bm{$0.82$} & $0.83$ & \bm{$1.00$} & $0.14$ & \bm{$0.97$} & $0.33$ & \bm{$0.64$} & $0.00$ & \bm{$0.47$} \\ 
        \bottomrule
    \end{tabular}
    }
    \end{sc}
    \end{center}
\vskip -0.25 in
\end{table}
%
%

\begin{figure*}[!ht]
    \centering
    \includegraphics[width=\textwidth]{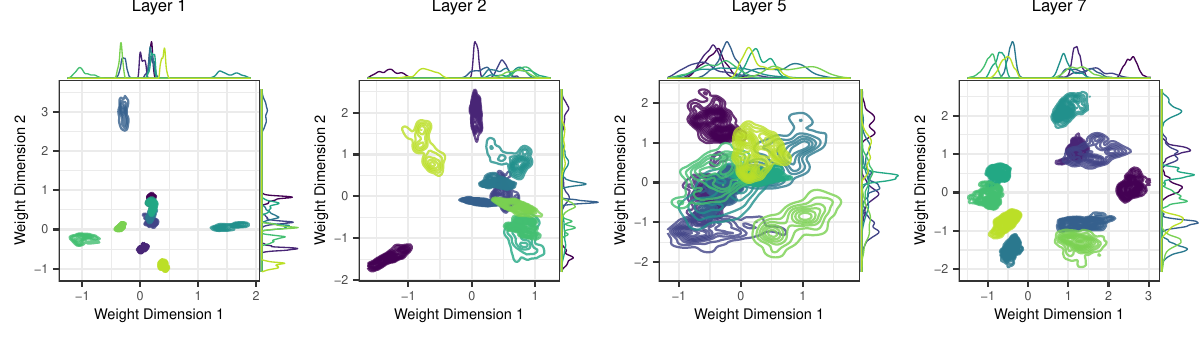}
    \vskip -0.1in
    \caption{Bivariate density plot of 10 chains (colors) of two randomly selected weights in the first, second, fifth, and final layer (from left to right) of a seven-layer BNN showing the varying degree of mode connectedness.}
    \label{fig:disconnectedmodes}
\end{figure*}

We start by investigating the general feasibility of SBI with HMC and NUTS. 
We initially focus on predictive performance, studying whether it is at all possible
to run these samplers and perform on par with the non-Bayesian SOTA. 
We also analyze in \cref{sec:practicalsbi} whether the successful configurations provide useful uncertainty quantification.
While various studies reported on the performance of BNNs based on SBI \citep[e.g.,][]{izmailov_2021_WhatArea}, we are not aware of any other systematic investigation beyond performance. 

\begin{conjecture}
    For certain architectures and sampling algorithms, BNNs using SBI can achieve SOTA performance.
\end{conjecture}

We hypothesize that published results obscure the fact that samplers cannot deal with all existing BNN posterior problems off-the-shelf.
As a sanity check, we match the RMSE performance of SBI against that of a linear model (LM). 
Using the sampled chains that outperform the LM, we then compare the resulting BNN performance, i.e., the ensemble of all chains and their 8,000 samples, to a random forest (RF; tuned with optuna \citep{optuna2019}) as a stronger baseline, a (non-Bayesian) DNN of the same architecture, and a deep ensemble \citep[DE;][]{lakshminarayanan_2017_SimpleScalablea} with 12 members. 
The DNN and DE models are trained with Adam \citep[][further details in~\cref{app:setup}]{kingma_2015_AdamMethod} and the BNNs are sampled using NUTS with unit Gaussian prior and tanh activation. 
In order to show the ensemble effect of the BNN, we also report the performance of the model induced by a single random sample (RS) from the chains of the BNN.

\textbf{General feasibility and tanh SOTA performance}. 
Using chains that outperform the LM, samplers indeed yield SOTA performance across different experimental settings for tanh activation (cf.~\cref{tab:rmse_bench}). 
While individual samples (RS) cannot outperform the RF, DNN or DE, their ensemble supersedes the baselines in performance. We discuss the benefits of multi-chain SBI in Sections~\ref{sec:multimodality} and \ref{sec:dying_bde}. 

\textbf{The unboundedness problem}. In contrast to the results of tanh networks, off-the-shelf SBI does not produce meaningful results for ReLU-, SiLU- or LeakyReLU-activated BNNs across different architectures, prior choices, and data sets. 
Instead, we observe what we call \emph{dying samplers}, getting stuck almost immediately and never leaving the area of the starting value (cf.~\cref{app:additional_exp}).
This becomes evident in \cref{tab:prop_blm} reporting the share of chains that perform better than an LM in terms of RMSE for ReLU and tanh activation. The number of successful chains with ReLU is notably smaller compared to tanh, which holds true also for other (unbounded) activation functions (cf.~\cref{fig:allactiv} in the Appendix). As is known from classical DNN optimization, weight variances in networks with unbounded activation functions can explode if not properly initialized \citep[see, e.g.,][]{he2015delving}. Similarly, if priors remain constant across layers with variance independent of the size of each layer, we observe that SBI will not produce chains better than a simple baseline.
Hence, choosing bounded activation functions might help to achieve good performance. 
However, even for tanh, \cref{tab:prop_blm} shows that the more complex the architecture, the less reliable the samplers become.
We elaborate on a potential solution of this problem in~\cref{sec:dying_bde}.

\textbf{Superiority of NUTS}.
NUTS outperforms HMC across different data sets and activation functions (cf.~\cref{tab:comparison_nuts_hm} in the Appendix). 
Overall, HMC with fixed hyperparameters can only produce chains with better performance than the LM in 1\% of all experiments, suggesting that the refined handling of sampling trajectories by NUTS is indispensable for BNNs. Tuning HMC's hyperparameters, in turn, might be less efficient than NUTS, with the latter performing well out of the box in many cases.
%

%

\section{Multimodality of Posteriors} \label{sec:multimodality}
For a deeper understanding of the previous results, we now turn to the peculiarities of BNNs and how they affect SBI. 
A key insight in this respect refers to the symmetries caused by overparameterization (cf.~\cref{fig:1-1-1}). 
We observe that this redundancy becomes more pronounced with network depth.

\begin{conjecture}
    In BNNs, the uncertainty of the weight distributions progressively increases in deeper layers. 
    This means that more mode connectivity is observed in deeper layers.
\end{conjecture}

\phantom{foo}

\textbf{Multimodality of the posterior}. 
\cref{fig:1-1-1} suggests that prior influence can induce a \emph{merging of modes}. This means that practically irrelevant likelihood regions between two posterior modes increase in posterior probability through the prior contribution, allowing samplers to traverse from one mode to another. In apparent contrast to this initial hypothesis on mode connectivity, we do observe multimodality across various settings for different prior variances.
Even in larger networks and smaller data sets, the model complexity and data signal may outweigh prior information and (initially) induce disconnected modes in the posterior (see \cref{fig:disconnectedmodes}, left, visualizing the densities of weights in different layers marginalized for two dimensions). 
Thus, the general existence of connected mode surfaces is not guaranteed.

\begin{figure}[!ht]
    \centering
    \includegraphics[width = 0.85\columnwidth]{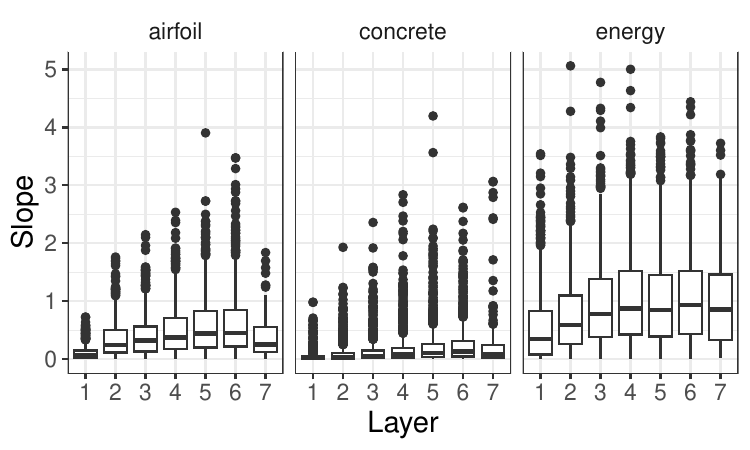}
    \caption{Range of ``motion'' of chains per layer (x-axis) and selected data sets (facets). Boxplots show the distribution of absolute values of slopes obtained by fitting a linear regression for each chain to capture chain trends during the sampling process, clearly indicating more movement in deeper layers.}
    \label{fig:slopes}
\end{figure}

\begin{figure}[!ht]
    \centering
    \vskip 0.1in
    \includegraphics[width = 0.95\columnwidth]{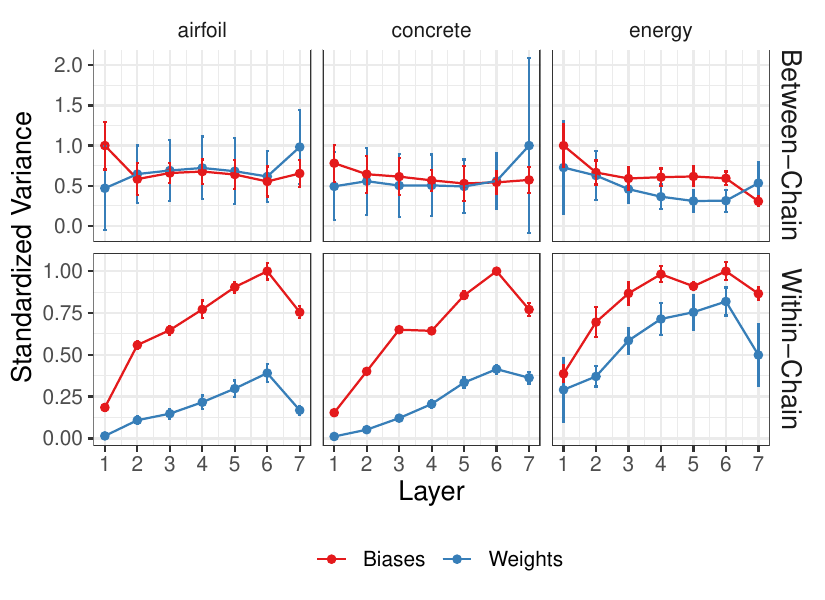}
    \vskip -0.2in
    \caption{Mean and standard deviation of the between- and within-chain variance (in different rows) separated by layer (x-axis) for different data sets (columns) of a seven-layer BNN.}
    \label{fig:withinvariance}
\end{figure}

\textbf{More movement deeper down.}
It turns out, however, that this phenomenon requires a more nuanced analysis.
As shown in \cref{fig:disconnectedmodes}, the posterior density is heavily concentrated on disconnected modes in the first layer, but then becomes more diffuse and thus connected in deeper parts of the BNN before concentrating again in the output layer. 
Our observation is reflected by chains growing notably more variable in deeper layers (cf. \cref{fig:slopes} and~\ref{fig:withinvariance}); only in the last layer does the variance decrease again.
The explanation for this behavior can be found in the varying degree of overparameterization throughout the network.
We conjecture that disconnected modes occur mainly in layers with limited flexibility, i.e., where the role of the neurons is controlled by their close connection to the input (or output) that populates a certain manifold.
Hidden-layer weights, by contrast, have vastly more degrees of freedom in assuming diverse values and swapping their roles in creating input embeddings as long as the output is realized in an appropriate range.


\textbf{Role of network layers.}
Our intuition about the roles of different layers echoes recent findings about the varying robustness of layers to re-initialization during training \citep[where the initial layer, in accordance with our results, proves much more sensitive to resetting the parameter values mid-training than intermediate ones;][]{zhang_2022_AreAll}.
The layerwise investigation also has interesting relations to the mode connectivity studies in \citet{izmailov_2021_WhatArea}.
Taking the notion of mode connectivity to a per-layer level, work concurrent to ours \citep{adilova_2024_LayerwiseLineara} suggests that intermediate layers play a different role for the success of layerwise neuron alignment \citep[interpolating between multiple independent solutions; cf.][]{ainsworth_2023_GitReBasin} than the first or last one.
We further believe our results could shed light on the effectiveness of methods such as subspace inference \citep{izmailov_2020_SubspaceInferencea,dold2024} that move or sample in directions of quasi-constant loss. 
More immediately, as we discuss in the next section, they have important consequences for practical SBI.

\section{Practical SBI} \label{sec:practicalsbi}

%


\subsection{How to Handle Multimodality in Practice?} \label{sec:multiplechains}

Given the existence of multimodal posterior surfaces that samplers do not seem to be able to traverse, the question arises of how SBI can still achieve useful results. 
One possibility \citep[also mentioned in][]{riou-durand_2023_AdaptiveTuning, ecml} is to run multiple chains to cover as many modes as possible. 
Covering all modes seems to be an impossible endeavor given that even in small BNNs the number of symmetric modes easily exceeds $10^{200}$ \citep{ecml}. 
However, our results in the previous section give reason for hope and suggest that the number of disconnected modes (and thus required chains) does not simply scale with the complexity of the network. 
As overparametrization induces smoother posterior surfaces for deeper layers that samplers can explore much better (cf.~\cref{fig:slopes}), it is likely only necessary to cover the few distinct modes in earlier layers.

%

We now investigate the joint influence of the number of chains and the number of samples on the predictive and uncertainty performance of SBI.

\begin{conjecture}
    Multiple samples from multiple chains improve predictive performance and uncertainty quantification.
\end{conjecture}

\begin{figure}[!ht]
\centering
\includegraphics[width=0.9\columnwidth]{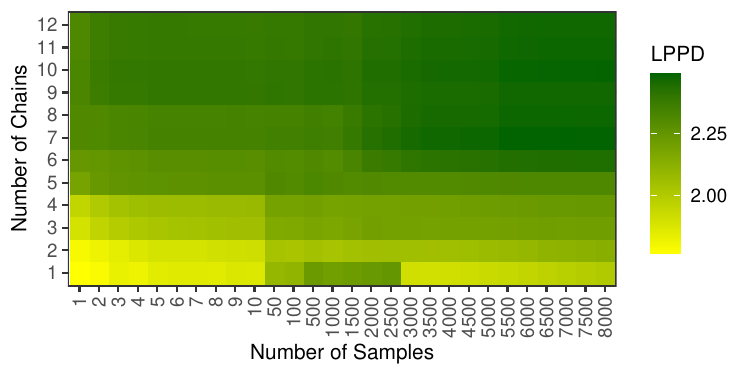}
\caption{Grid for the \texttt{energy} data set showing the change in LPPD with more samples and chains.}
\label{fig:chainssamples}
\end{figure}

\textbf{Performance and uncertainty}. 
\cref{fig:chainssamples} (further examples in~\cref{app:experiments_practical}) shows a clear upward trend of LPPD in both the number of chains and the number of samples. The same general pattern can be observed for the RMSE (Appendix \ref{app:experiments_practical}).
Notably, the LPPD increases more strongly for multiple chains than the RMSE, suggesting the importance of covering multiple modes, but we also see performance gains by exploring the region around modes with more samples. 
It seems promising that we obtain good results with a maximum of 12 chains despite the large number of symmetries.
This is in line with findings from \citet{ecml} who ran more than 1,200 chains but also observed early saturation of performance metrics.

\begin{figure}[!ht]
    \centering
    \includegraphics[width=\columnwidth]{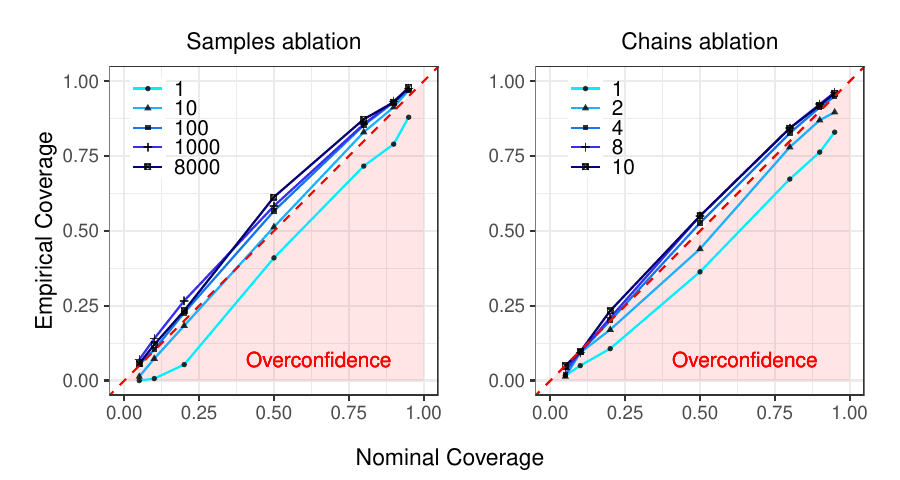}
    \caption{Nominal vs.~empirical coverage of posterior credibility intervals for different numbers of samples using 10 chains everywhere (left) and different numbers of chains using 100 samples each (right) for the \texttt{airfoil} data set, NUTS with 10,000 warmup, two hidden layers of 16 neurons each, tanh activation, and unit Gaussian priors.}
    \label{fig:coverage}
\end{figure}

\textbf{Predictive coverage}. 
Bayesian inference enables the computation of predictive credibility intervals which should be calibrated and cover the true prediction.
\cref{fig:coverage} (further results in~\cref{app:experiments_practical}) shows the coverage of credibility intervals for varying amounts of samples (left) and chains (right). 
For a calibrated model (represented by the diagonal), the observed labels fall within the $\alpha \%$ credibility interval in $\alpha \%$ of cases.
With increasing chains and samples, we can see a trend from overconfidence to close-to-nominal coverage. 
These results confirm that the use of multiple chains and samples indeed produces well-calibrated posteriors.

\subsection{Convergence Diagnostics}


The previous experiments investigated the performance of different numbers of samples per chain. 
However, in practice, it is often not obvious when to stop the sampling process. 
While there are well-established convergence diagnostics for Bayesian statistics, it is unclear whether these work well for BNNs. 
In contrast to the existing literature, which tends to disregard convergence analysis of the parameter space, we will discuss both parameter and function space diagnostics. 
For both, the \textit{de facto} standard for measuring convergence is the \emph{rank-normalized split} $\widehat{R}^{(2)}$ as introduced in \citet[][see~\cref{app:conv_diag} for a definition and functional extension]{vetharirhat2021}. 

\begin{conjecture}
    Due to symmetries and large differences in within-chain variance across layers, classical diagnostics are not useful for checking convergence in BNNs.
\end{conjecture}

\textbf{Parameter space convergence}. 
The classical $\widehat R$ diagnostic assumes identifiable parameters, which is not given in BNNs due to the aforementioned symmetries. Our experiments in~\cref{app:conv_diag} also confirm empirically that $\widehat R$ is not a suitable metric. Another pitfall we can infer from our results is the aggregation of $\widehat R$ over all model parameters. 
As $\widehat R$ is normalized using the within-chain variance, it decreases in the later layers of the BNN due to the increase in variance (\cref{fig:withinvariance}), leading to the false conclusion of convergence of the whole network when averaging across all weights. 

\begin{figure}[!ht]
    \centering
    \begin{subfigure}{0.43\columnwidth}
    \centering
        \includegraphics[width=\columnwidth]{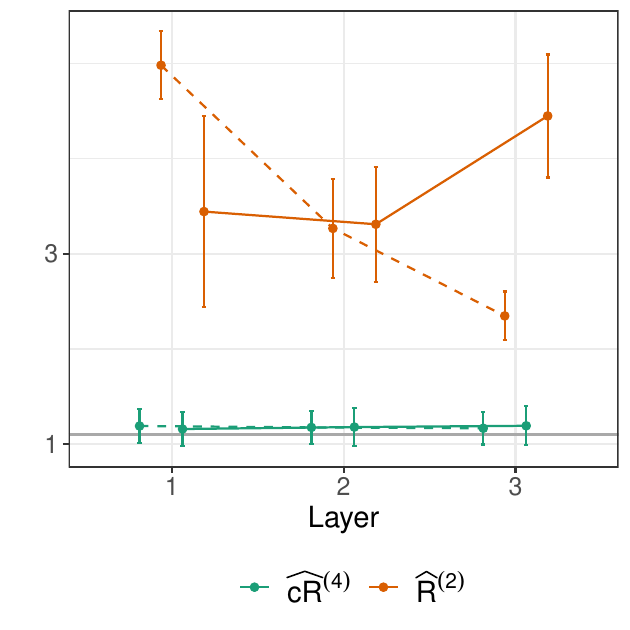}
    \end{subfigure}
    \begin{subfigure}{0.43\columnwidth}
    \centering
        \includegraphics[width=\columnwidth]
        {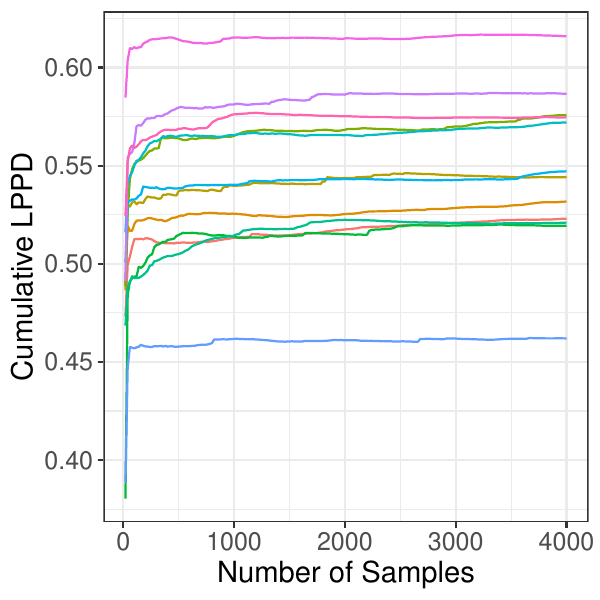}
    \end{subfigure}
    \caption{\emph{Left:} Comparison between $\widehat{R}^{(2)}$ and $\widehat{cR}^{(4)}$ for weights and biases (in different linetypes). \emph{Right:} Cumulative LPPD across the process of collecting 4,000 samples and different chains (colors). Both plots are based on the \texttt{bikesharing} data set. }
    \label{fig:bike_cum_lppd}
    \vskip -0.2in
\end{figure}

Thus, we argue that parameter space convergence should be measured both chain- and layer-wise. 
For the chain-wise convergence, we propose the $\widehat{cR}^{(\kappa)}$ diagnostic (defined in~\cref{app:conv_diag}), which splits a chain's sample path into $\kappa$ subchains of equal length as the basis for the $\widehat R$ calculation. Empirically, we observe  $\widehat{cR}^{(\kappa=4)}$ values close to or lower than $1.1$ for well-performing and calibrated models, indicating chain-wise convergence
(cf.~\cref{fig:bike_cum_lppd}, left, and \ref{fig:param_rhats}).

\textbf{Function space convergence}. Various post-hoc methods exist to evaluate the function space convergence of BNNs, all using a hold-out test data set (see~\cref{app:conv_diag}). While some report the pointwise $\widehat R$ values for the PPD samples of each test data point \citep{izmailov_2021_WhatArea} and visually analyze their distribution (cf.~\cref{fig:rhat_hist}), others calculate the $\widehat R$ for a goodness-of-fit measure over the test data set (e.g., log-likelihood) to obtain a single diagnostic \citep{fortuin2022bayesian}. Both approaches penalize disagreement between chains. 
Taking into account the disconnectedness of modes in earlier layers,
we cannot, however, expect the convergence of all chains to a common function outcome. In particular, our experiments show the existence of substantial between-chain variance in the function space for a well-calibrated model (e.g., \cref{fig:func_lpl}).
We argue that a proper function space convergence metric should not penalize these chain-wise differences as visiting different modes is essential for well-working SBI. 

Therefore, we propose to monitor the convergence of each chain individually by the cumulative LPPD, as defined in~\cref{eq:lppdcum} in the Appendix and  shown in, e.g., \cref{fig:bike_cum_lppd} (right). By running multiple chains, we can compare the different cumulative LPPD paths and thereby obtain a better understanding of the convergence of each chain and the difference in chain performance. This can also be used as an early-stopping criterion of the sampling process for potentially converged or diverging chains, freeing resources and allowing to start new chains for more efficiency. 
%





\section{Dying Sampler \& Deep Ensemble Initialization} \label{sec:dying_bde}

Equipped with a clearer understanding of BNN posteriors and practical strategies to obtain calibrated SBI, we now address the unresolved problem of chains that perform worse than a simple baseline, leading to samplers ``dying" soon after initialization (cf.~\cref{sec:general} and \cref{fig:stuck_traces} in the Appendix).

\subsection{The Dying Sampler Problem}
Although~\cref{sec:general} established that the use of bounded activation functions largely avoids the dying sampler problem,
it would be unsatisfactory to simply dismiss ReLU BNNs as infeasible or refrain from using larger architectures. 
While previously discussed results on prior variance offer a possible explanation for this phenomenon, a---if not the most important---cause is already evident in \cref{fig:1-1-1}: \emph{Starting values for chains might prompt the sampler to get stuck or not explore meaningful hypotheses}. 

For example, as a consequence of the ReLU activation function being zero for negative input signals, many neurons---just by chance---will not be activated, creating low-density regions at the locations of the weights associated with those neurons (the white area in \cref{fig:1-1-1}). 
Subsequent proposals will then also end up in regions with a posterior probability of practically zero. 
We can thus draw an analogy to the ``dying ReLU problem'' \citep{dyingrelu} caused by zero gradients of non-activated ReLU neurons in optimization.

\subsection{Deep Ensemble Initialized MCMC} \label{sec:bde}
\begin{table*}[!t]
    \centering
    \small
    \setlength{\tabcolsep}{2pt}
    \caption{Average RMSE and LPPD values ($\pm$ their standard deviations) of the LM, a classic DNN, a DE, and our method (i.e., DEI-MCMC) using different amounts of samples (in brackets) for all benchmark data sets (in rows; abbreviated by their first letter). All networks have two hidden layers with 16 neurons each and ReLU activations. The best method per data set is highlighted in bold.}
    \vskip 0.1in
    \label{tab:bdes}
    \begin{sc}
    \resizebox{\textwidth}{!}{
    \begin{tabular}{clcccccccccccc}
        \toprule
        &  & \multicolumn{6}{c}{RMSE ($\downarrow$)} & \multicolumn{6}{c}{LPPD ($\uparrow$)} \\
        \cmidrule(lr){3-8} \cmidrule(lr){9-14}
        & & LM & DNN & DE & Ours (10) & Ours (100) & Ours (1000) & LM & DNN & DE & Ours (10) & Ours (100) & Ours (1000) \\ 
        \midrule
        & A\, & 0.72$\pm$0.02 & 0.27$\pm$0.01 & 0.24$\pm$0.01 & {0.21$\pm$0.01} & {0.21$\pm$0.01} & \textbf{0.21$\pm$0.01} & 
        -1.09$\pm$0.03 & -0.17$\pm$0.08 & \phantom{-}0.19$\pm$0.05 & \phantom{-}0.50$\pm$0.01 & \phantom{-}0.53$\pm$0.02 & \textbf{\phantom{-}0.58$\pm$0.03} \\
        & B\,    & 0.77$\pm$0.01 & 0.31$\pm$0.00 & 0.30$\pm$0.00 & 0.26$\pm$0.01 & {0.25$\pm$0.01} & \textbf{0.25$\pm$0.01} & 
        -1.16$\pm$0.01 & \phantom{-}0.35$\pm$0.00 & \phantom{-}0.42$\pm$0.00 & \phantom{-}0.60$\pm$0.02 & \phantom{-}0.61$\pm$0.02 & \textbf{\phantom{-}0.63$\pm$0.02}\\
        & C\,   & 0.63$\pm$0.02 & 0.35$\pm$0.04 & 0.32$\pm$0.05 & 0.40$\pm$0.19 & 0.70$\pm$0.71 & \textbf{0.31$\pm$0.05} & 
       -0.96$\pm$0.03 & -0.99$\pm$0.38 & -0.08$\pm$0.20 & \phantom{-}0.05$\pm$0.27 & \phantom{-}0.19$\pm$0.11 & \textbf{\phantom{-}0.23$\pm$0.12} \\
        & E\,     & 0.27$\pm$0.02 & 0.05$\pm$0.00 & {0.04$\pm$0.01} & {0.04$\pm$0.00} & {0.04$\pm$0.00} & \textbf{0.04$\pm$0.00} & 
       -0.13$\pm$0.06 & \phantom{-}1.60$\pm$0.13 & \phantom{-}1.93$\pm$0.05 & \phantom{-}1.91$\pm$0.28 & \phantom{-}1.94$\pm$0.25 & \textbf{\phantom{-}2.06$\pm$0.20} \\
        & P\,    & 0.85$\pm$0.01 & 0.77$\pm$0.01 & 0.77$\pm$0.01 & 0.71$\pm$0.00 & 0.71$\pm$0.00 & \textbf{0.70$\pm$0.00} & 
        -1.25$\pm$0.01 & -1.06$\pm$0.01 & -1.05$\pm$0.01 & -0.78$\pm$0.02 & -0.77$\pm$0.02 & \textbf{-0.75$\pm$0.02} \\
        & Y\,      & 0.61$\pm$0.07 & 0.05$\pm$0.00 & {0.03$\pm$0.01} & {0.03$\pm$0.01} & {0.03$\pm$0.01} & \textbf{0.03$\pm$0.01} & 
        -0.94$\pm$0.13 & \phantom{-}1.74$\pm$0.40 & \phantom{-}2.55$\pm$0.05 & \phantom{-}2.97$\pm$0.17 & \phantom{-}3.00$\pm$0.20 & \textbf{\phantom{-}3.08$\pm$0.20} \\
        \bottomrule
    \end{tabular}
    }
    \end{sc}
\end{table*}
%

As already elaborated in~\cref{sec:general}, using a different initialization solves convergence issues such as the dying sampler problem. 
This motivates our concluding proposal for practically useful SBI, which we call \emph{Deep Ensemble Initialized MCMC (DEI-MCMC)}. Inspired by their similarity to DEs one could also think of them as a \emph{Bayesian Deep Ensemble}. 
Specifically, we suggest running a standard optimization procedure for $M$ networks in a non-Bayesian fashion, and using the resulting $M$ sets of network weights as initial proposals for the samplers.
Note that this does not require changing the prior distribution.
We combine the information of all chains in the classical Bayesian sense by merging their empirical distributions.
DEI-MCMC has several attractive properties, which we discuss in the following.

\textbf{Valid starting values}. 
As priors with support on the entire domain do not per se restrict the admissible values of weights
(but activation functions in the model might), we can expect initialization with the $M$ sets of weights to yield valid results. 
In other words, the starting values will produce non-zero posterior values and, hence, solve the problem of the dying sampler.

\textbf{Expected improvement}. DEI-MCMC is equivalent to DEs in the edge case where we only obtain one sample from the posterior (the initial weight). As DEs have repeatedly shown top-notch quantifying quantification, this means that DEI-MCMC is likely at least as good as DEs in this respect.
In our experiments, DEI-MCMC indeed provides better LPPD for as few as ten samples (cf.~\cref{tab:bdes}).

\textbf{A modular Bayesian toolbox}. Similar to the Laplace approximation \citep{daxberger_2021_LaplaceReduxa}, DEI-MCMC can be used post-optimization, allowing to combine prior knowledge of any kind with a non-Bayesian network into a BNN. In particular, DEI-MCMC is applicable to pre-trained networks.

\textbf{Flexibility}. While DEI-MCMC is as expensive as DEs by design and requires further computation for additional sampling, this second process can be flexibly adjusted to the availability of resources and computing time. 
Furthermore, since our previous results suggest that a few samples already yield very good performance when using multiple chains, it might be possible to shorten the sampling step considerably by saving on warmup iterations.  

\subsection{Numerical Experiments}
In order to investigate the effectiveness of DEI-MCMC, we run an additional benchmark study. 
Proving that the proposed warm start can solve the dying sampler problem, we use BNNs with ReLU activation and compare the performance of DEI-MCMC using all chains to those of an equivalent DNN and DE. In other words, we do not need to filter chains for the comparisons as done in~\cref{sec:general}. 

\textbf{Results}. We summarize our findings in \cref{tab:bdes}. As becomes evident from the comparisons with the LM and DE, the warm starts of samplers using the $M$ different network solutions avoid any dying sampler problems (no model shows worse performance than the baseline LM). Furthermore, we see the added benefit of sampling around the DE solutions when continuing the sampling from these solutions after a very short warmup phase of 100 steps, suggesting that this process supplies valuable information to the model both for predictive (RMSE) and uncertainty (LPPD) performance. 
It should be noted that, although DEI-MCMC tends to benefit from longer sampling, there is a diminishing return to extending the sampling phase beyond the 1,000 samples reported in the last column of~\cref{tab:bdes}, with only marginal improvements from drawing even up to 10,000 samples.
While the best cold-started BNNs required 10,000 warmup steps, DEI-MCMC achieves competitive performance with a 100$\times$ shorter warmup phase.
We thus conclude that DEI-MCMC, though conceptually simple, offers a consistent and robust improvement over classical BNNs.



\section{Discussion}

In this work, we discussed sampling-based inference in BNNs. We argue that sampling from a posterior in these networks is not only feasible but can achieve SOTA results when accounting for the nature of posterior landscapes. 
Through extensive experiments with different BNN architectures, we present insights into the successes and limitations of SBI in this context. Critical findings include performance differences of samplers for bounded and unbounded activation functions and the increasing connectedness of modes in deeper layers. Following these results, we recommend a multi-chain and multi-sample strategy using NUTS, with a convergence diagnostic that accounts for heterogeneous variances in different layers. In order to allow samplers to also explore highly multimodal landscapes, particularly those induced by ReLU networks, we propose a novel approach, \emph{Deep Ensemble Initialized MCMC (DEI-MCMC)}, which uses optimized networks as a starting point for posterior sampling. These ensembles can be seen as a modular Bayesian toolbox applicable to any network post-optimization independent of the state of prior knowledge.

\textbf{Limitations and future work}. Due to the multitude of experiments, as well as the long duration of sampling procedures, the analysis presented in this paper is limited in some aspects. In particular, only full-batch sampling routines were tested. An exciting target for future analysis, therefore, is to explore the potential of SG-MCMC samplers in the context of DEI-MCMC, as well as an extension to larger data sets. 
Likewise, the performance of DEI-MCMC in uncertainty-related downstream tasks, such as out-of-distribution detection, remains an open question.
It would further be interesting to deepen our findings in conjunction with the insights from mode connectivity \citep{garipov_2018_LossSurfacesb} and subspace research \citep{izmailov_2020_SubspaceInferencea}.
Finally, inspired by our tentative results and the work by \citet{adilova_2024_LayerwiseLineara}, all these avenues of future work should be pursued adopting a layerwise lens.

\section*{Impact Statement}
This paper presents work whose goal is to advance the field of Machine Learning. There are many potential societal consequences of our work, none which we feel must be specifically highlighted here.

\section*{Acknowledgments}
LW is supported by the DAAD programme Konrad Zuse Schools of Excellence in Artificial Intelligence, sponsored by the German Federal Ministry of Education and Research.

\bibliography{refs}
\bibliographystyle{icml2024}

\newpage
\appendix
\onecolumn

\section{Inference in Weight vs.~Function Space} \label{app:discussweight}

Some authors suggest completely disregarding the parameter space \citep{tran2022all, arbel_2023_PrimerBayesiana} since we arguably care more about the functions learned by a DNN than about single weights in a million-dimensional space. However, opting for inference over weights but then deliberately turning a blind eye to everything that happens in the weight space seems unsatisfactory. Other authors argue that expressing prior beliefs in function space \citep{tran2022all} is the more meaningful approach when researchers rarely have actual prior knowledge about BNN weights. Formulating a prior about a complex functional mapping is, however, not necessarily a simpler problem. While each side has its valid points, a general discussion of the meaningfulness of sampling and inference in the parameter space of neural networks is beyond the scope of this work. At this point, we just point out briefly that sampling-based inference will remain highly relevant for simpler models that offer some degree of interpretability (e.g., in statistics), and since there is no clear cut-off for when a model is too complex, researchers will eventually use sampling-based inference for BNNs, in particular as computational resources and performance increase. 

\section{Details on Convergence Diagnostics} \label{app:conv_diag}
\subsection{Parameter Space}

The two standard diagnostics for convergence of MCMC methods are currently the rank-normalized split-$\widehat{R}$ \citep{gelman_2013_BayesianData, vetharirhat2021} and the effective sample size \citep[ESS;][]{vetharirhat2021}. We extend the rank-normalized split-$\widehat{R}$ to a functional version. For a transformation $\psi(\cdot)$ and posterior samples $\bm{\theta}^{(k,s)}$ from chain $k \in 1,\ldots,K$, $s \in 1,\ldots,S$, we define the diagnostic as follows:

\begin{definition}[\textbf{$\widehat{R}_\psi$}] \label{def:rhat}
    Let $\psi: \mathbb{R}^d \to \Psi; \bm{\theta} \mapsto \psi(\bm{\theta})$ be a transformation function transforming the parameter samples into some metric of interest. $\bar{\bm{\theta}}^{(k, \cdot)} = \frac{1}{S} \sum_{s = 1}^S \psi(\bm{\theta}^{(k,s)})$ and $\gamma_k^2 = \frac{1}{S - 1} \sum_{s = 1}^S \left(\psi(\bm{\theta}^{(k,s)}) - \bar{\bm{\theta}}^{(k,\cdot)}\right)^2$ denote the (potentially vector-valued) empirical mean and variance of the functional of the posterior samples of chain $k$, respectively. In case $\Psi\nsubseteq \mathbb{R}$, i.e., $\psi$ is not mapping to (a subset of) $\mathbb{R}$ but results in a vector, the $(\cdot)^2$ is supposed to be interpreted element-wise applied to all dimensions of $\Psi$ individually. In all cases considered in this paper, this means that $\psi$ either maps a sample to a scalar statistic or creates a statistic for every of the $d$ dimensions of $\bm{\theta}$. Further, denote by $\bar{\bm{\theta}}^{(\cdot, \cdot)} = \frac{1}{K} \sum_{k = 1}^K \bar{\bm{\theta}}^{(k, \cdot)}$ the grand mean over all chains and samples.
    Then, we can define the within-chain variance $W$ and the between-chain variance $B$ as
    \begin{align}
        B &= \frac{S}{K-1}\sum_{k=1}^K \left(\bar{\bm{\theta}}^{(k,\cdot)} - \bar{\bm{\theta}}^{(\cdot,\cdot)}\right)^2, \\ 
        W &= \frac{1}{K} \sum_{k=1}^K \gamma_k^2,
    \end{align}
    where $(\cdot)^2$ is again supposed to be interpreted element-wise in case the means represent vectors.
    Normalizing $B$ with $W$ results in (a potentially vector-valued)
    \begin{align}
        \widehat{R}_\psi &= \sqrt{\frac{\frac{S-1}{S}W+\frac{1}{S}B}W}. \label{eq:splitrhat}
    \end{align}
    For $\psi$ being the identity we recover the $\widehat{R}$ as defined in \citet{vetharirhat2021} and omit the $\psi$ subscript in this case.
\end{definition}

\textbf{Split-$\widehat{R}^{(2)}_\psi$}.
\citet{gelman_2013_BayesianData} proposed to apply $\widehat{R}$ on a single chain split into two sub-chains, thus providing a measure for convergence for $K = 2$ sub-chains.
Intuitively, split-$\widehat{R}^{(2)}$ assesses the stationarity of a chain by comparing its mixing over the first and second half of the considered window.
This can again be generalized to $M$ simulated chains by splitting each of the $M$ chains and setting $K = 2M$. The concept is trivially applicable for the functional version to obtain Split-$\widehat{R}^{(2)}_\psi$.

\textbf{Split-$\widehat{R}^{(\kappa)}_\psi$}.
We modify Split-$\widehat{R}^{(2)}_\psi$ for a general number $\kappa$ of sub-chain splits and denote this convergence measure by $\widehat{R}^{(\kappa)}_\psi$.
Effectively, we construct $K = \kappa \cdot M$ chains from $M$ Markov chains.
Suitable values for $\kappa$ ensure the sub-chains retain a certain length so mean and variance estimates are meaningful.
Our experiments suggest that $\kappa = 4$ often produces good results. 

\textbf{Rank normalized split-$\widehat{R}^{(\kappa)}_\psi$}.
Lastly, split-$\widehat{R}^{(\kappa})$ can be improved by rank-normalizing the posterior samples to achieve standardized sample distributions \citep[see][for a detailed description]{vetharirhat2021}. The concept is again trivially applicable for the functional version by rank-normalizing the functional values $\psi(\bm{\theta}^{(k,s)})$.


While $\widehat{R}$ values $\leq 1.1$ were originally treated as a sufficient indication of convergence, \citet{vetharirhat2021} suggest that a value of $1.01$ is more appropriate.
This translates to a maximum between-chain variance of approximately 1\% of within-chain variance times the number of samples.
The diagnostics thus heavily penalize if chains converge to different modes---i.e., high between-chain variance---and implicitly assume identifiability of the parameter of interest.

\textbf{Rank normalized split-$\widehat{cR}^{(\kappa)}_\psi$}.
Since penalizing within-chain variance is not meaningful for multimodal BNN posteriors, we propose to measure the parameter space convergence both chain- and layerwise. Thus we define the chain-wise measure $\widehat{cR}^{(\kappa)}_\psi$ which refers to the application of $\widehat{R}^{(\kappa)}_\psi$ to a single chain ($M = 1$).

\begin{figure}[!ht]
    \centering
    \begin{subfigure}{0.48\textwidth}
        \includegraphics[width=\textwidth]{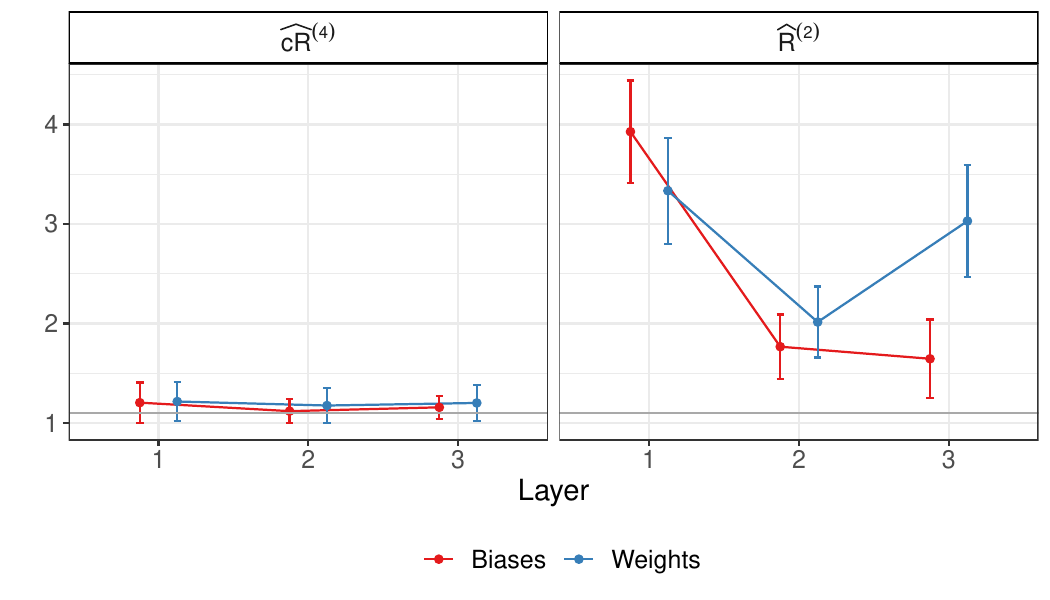}
        \caption{Airfoil}
    \end{subfigure}
    \begin{subfigure}{0.48\textwidth}
        \includegraphics[width=\textwidth]{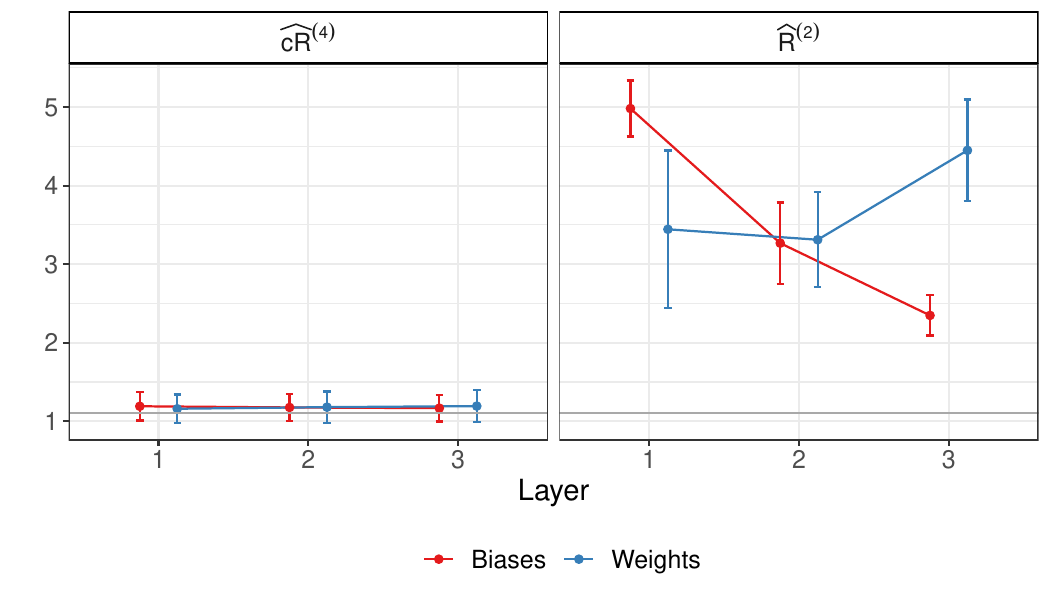}
        \caption{Bikesharing}
    \end{subfigure}

    \begin{subfigure}{0.48\textwidth}
        \includegraphics[width=\textwidth]{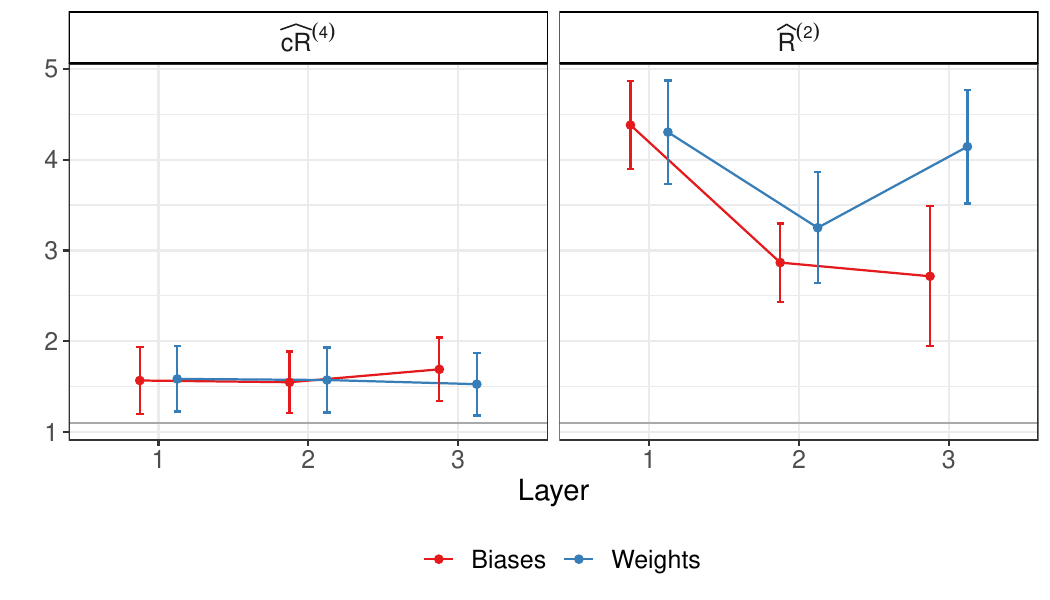}
        \caption{Concrete}
    \end{subfigure}
    \begin{subfigure}{0.48\textwidth}
        \includegraphics[width=\textwidth]{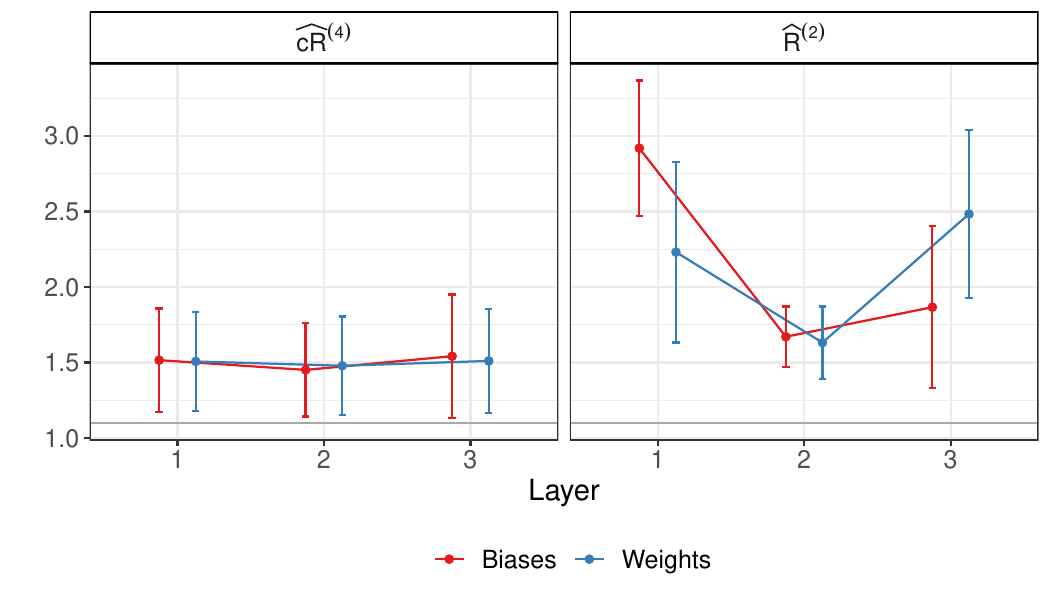}
        \caption{Energy}
    \end{subfigure}

    \begin{subfigure}{0.48\textwidth}
        \includegraphics[width=\textwidth]{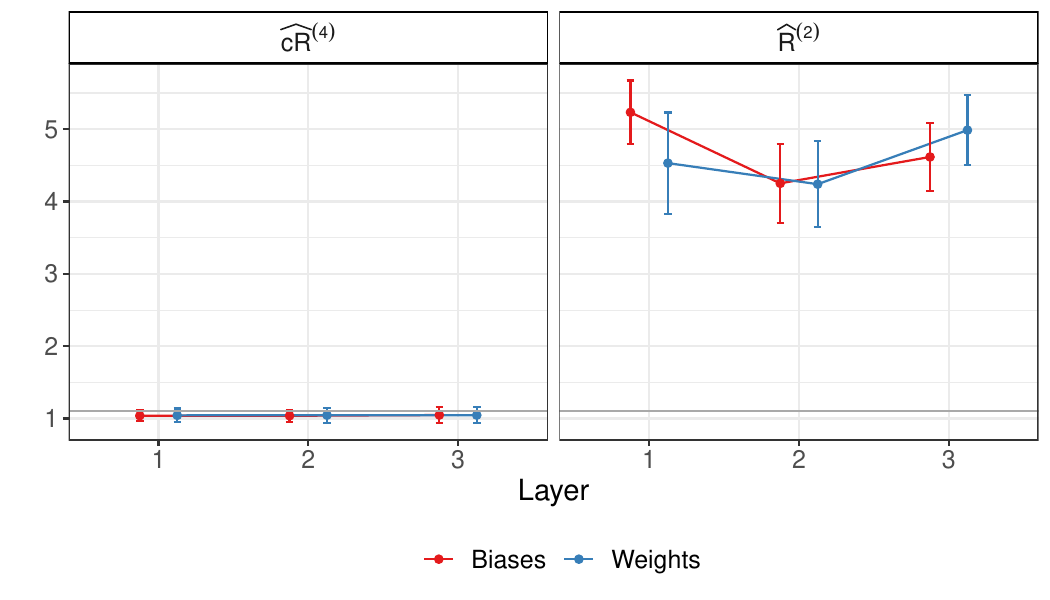}
        \caption{Protein}
    \end{subfigure}
    \begin{subfigure}{0.48\textwidth}
        \includegraphics[width=\textwidth]{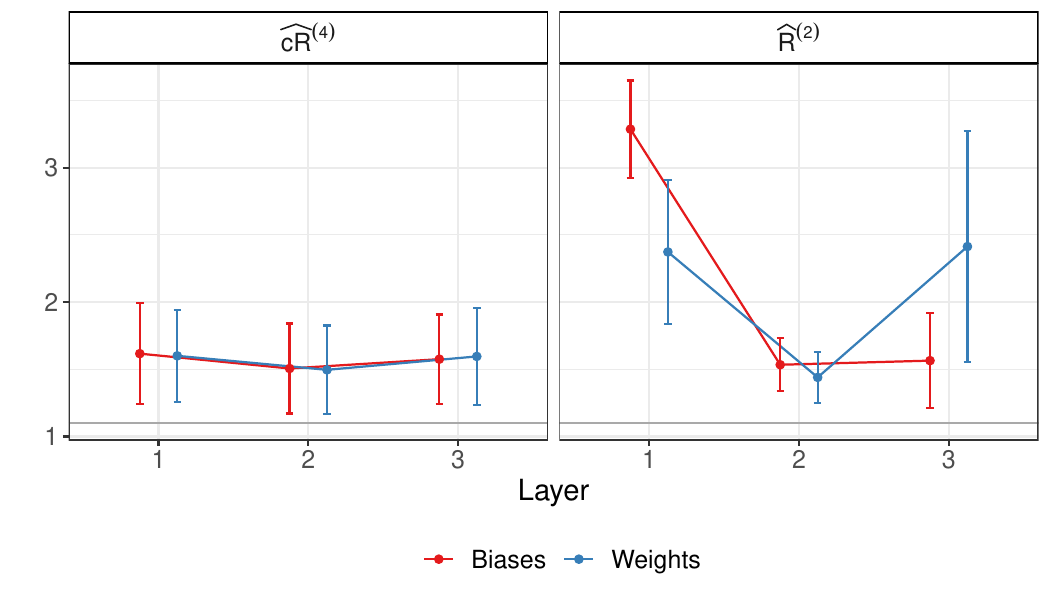}
        \caption{Yacht}
    \end{subfigure}
    \caption{Average parameter-space, log-transformed $\widehat{R}^{(2)}$ and $\widehat{cR}^{(4)}$ with standard error bars across layers. For each data set, all fitted chains with better-than-LM RMSE performance are used. In all cases, the BNN consists of two hidden layers with 16 neurons each and tanh activation. NUTS with a warmup phase of 10,000 steps and 8,000 samples per chain as well as unit Gaussian priors are used.}
    \label{fig:param_rhats}
\end{figure}

A comparison of classical $\widehat{R}^{(2)}$ and $\widehat{cR}^{(4)}$ in the parameter space is given in \cref{fig:param_rhats}.

\begin{figure}[!ht]
    \centering
    \begin{subfigure}{0.16\textwidth}
        \includegraphics[width=\textwidth]{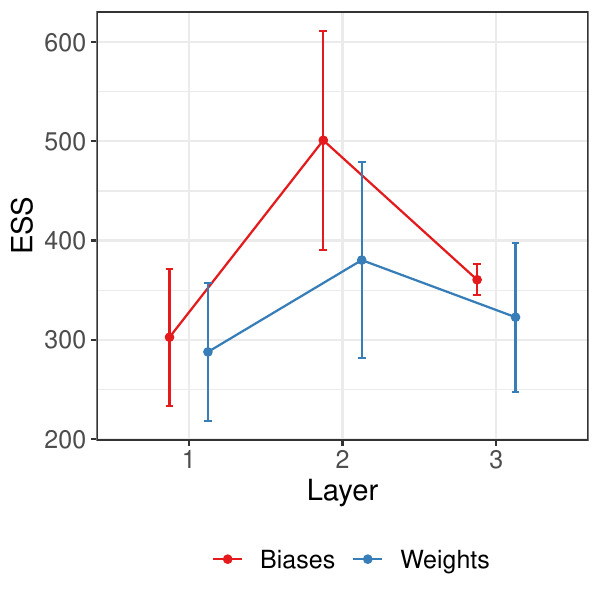}
        \caption{Airfoil}
    \end{subfigure}
    \begin{subfigure}{0.16\textwidth}
        \includegraphics[width=\textwidth]{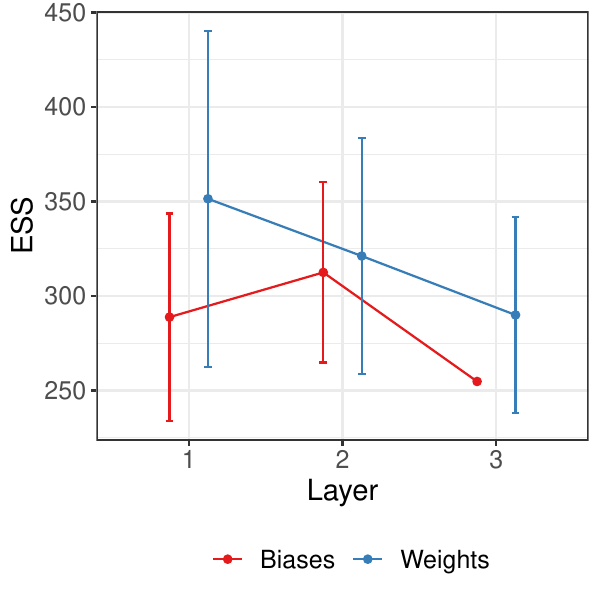}
        \caption{Bikesharing}
    \end{subfigure}
    \begin{subfigure}{0.16\textwidth}
        \includegraphics[width=\textwidth]{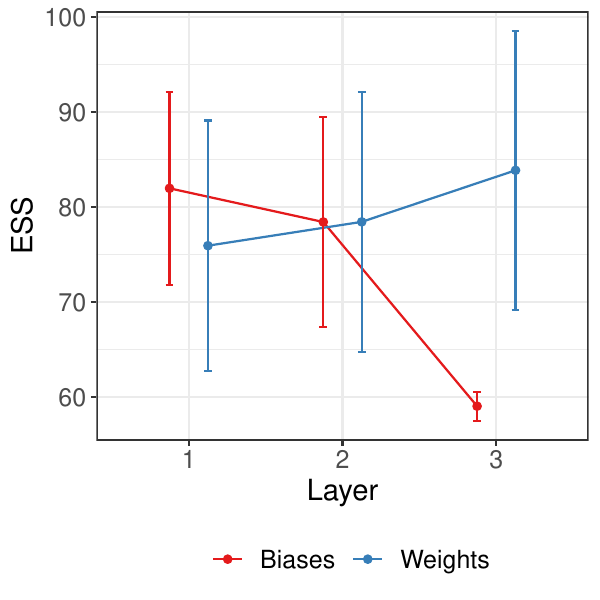}
        \caption{Concrete}
    \end{subfigure}
    \begin{subfigure}{0.16\textwidth}
        \includegraphics[width=\textwidth]{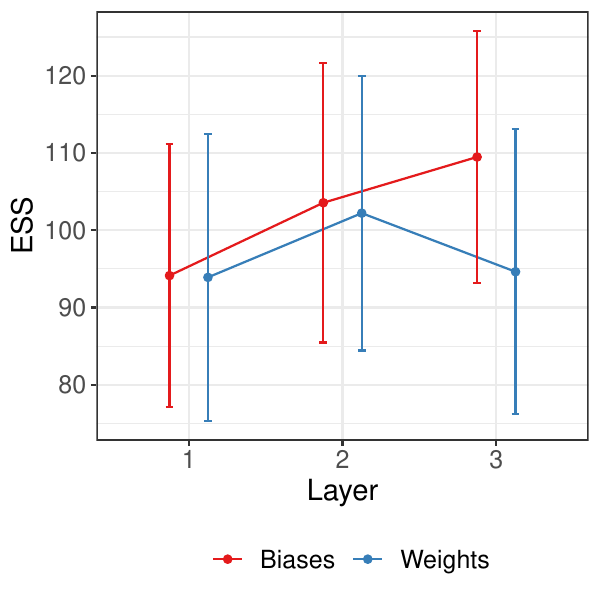}
        \caption{Energy}
    \end{subfigure}
    \begin{subfigure}{0.16\textwidth}
        \includegraphics[width=\textwidth]{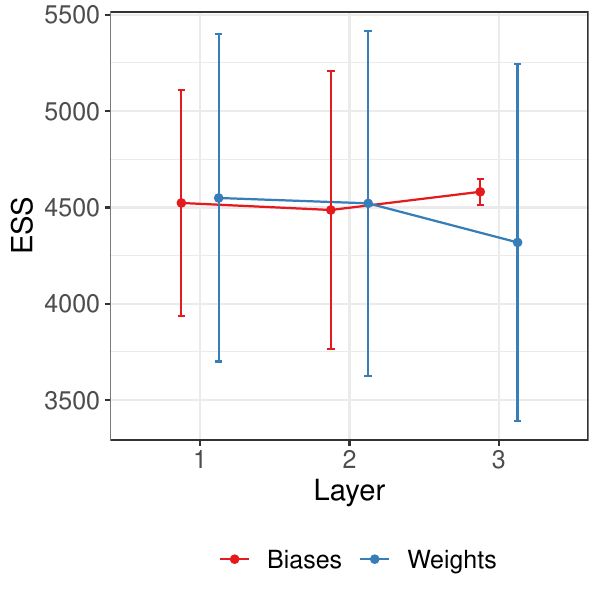}
        \caption{Protein}
    \end{subfigure}
    \begin{subfigure}{0.16\textwidth}
        \includegraphics[width=\textwidth]{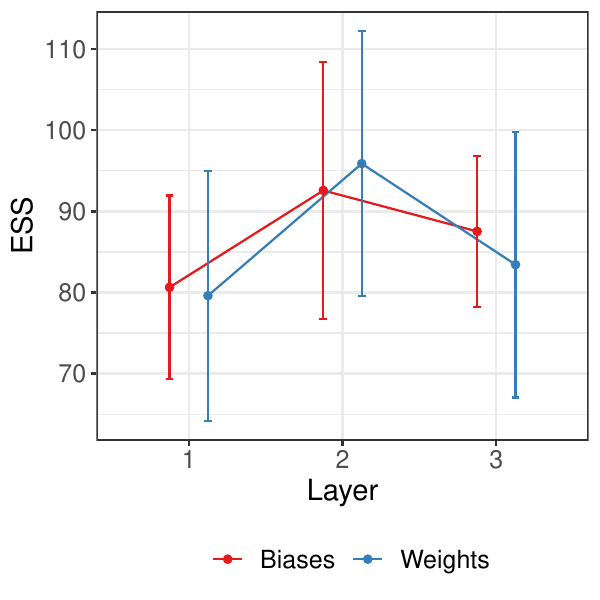}
        \caption{Yacht}
    \end{subfigure}
    \caption{The average parameter space ESS with standard error bars across layers and split by weights and bias role. For each data set, all fitted chains with better than LM RMSE performance are used. In all cases, the BNN consists of 2 hidden layers with 16 neurons each and tanh activation. NUTS with a warmup phase of 10k steps and 8k samples (4k for the two larger datasets) per chain as well as unit Gaussian priors are used.}
    \label{fig:param_ess}
\end{figure}

Effective Sample Size (ESS) is another classical MCMC convergence measure.
\begin{definition}[\textbf{Effective Sample Size (ESS)}]\label{def:ess}
    Following the \citet{stanrefmanual}, we define the effective sample size (ESS) as
    \begin{align}
        \text{ESS} = \frac{S}{1 + 2\sum_{t=1}^{\infty}\rho_t}
    \end{align}
    with $\rho_t$ being the autocorrelation between samples at time lag $t$.
\end{definition}
The ESS can be interpreted as the number of independent fictitious samples from the posterior itself that suffice to provide estimates as efficient as estimates from the $S$ MCMC samples.
Low ESS means that $S$ collected samples from the chain emulate only considerably fewer samples than $S$ from the true posterior density. 
Accordingly, we aim for large ESS values, which indicate low autocorrelation and thus a better exploration of the parameter space. 
Empirically, however, we observe that the ESS values of the parameter chains are rather small, in line with results reported in \citet{papamarkou_2022_ChallengesMarkova}. 
As there seems to be no systematic pattern across chains or layers like the ones observed for the $\widehat{R}$ case, we assume that the ESS is simply not a good measure of convergence in a highly overparametrized model where large autocorrelation of samples is to be expected. Visualizations of the ESS across layers for our benchmark data sets for a well-performing architecture of two hidden layers with 16 neurons each are displayed in Figure \ref{fig:param_ess}.

\subsection{Function Space}

Convergence can also be assessed in function space since we are ultimately interested in a good functional mapping.
For the calculation of convergence metrics like the $\widehat{R}^{(\kappa)}_\psi$ in function space, a hold-out test data set $\mathcal{D}_{\text{test}} \in (\mathcal{X} \times \mathcal{Y})^{n_{\text{test}}}$ is required. 
We first introduce two popular post-hoc approaches to measure function space convergence  \citep{izmailov_2021_WhatArea, fortuin2022bayesian} and then propose an alternative measure that can be used for online computation of diagnostics during the sampling process. 

\begin{figure}[!ht]
    \centering
    \begin{subfigure}{0.23\textwidth}
        \includegraphics[width=\textwidth]{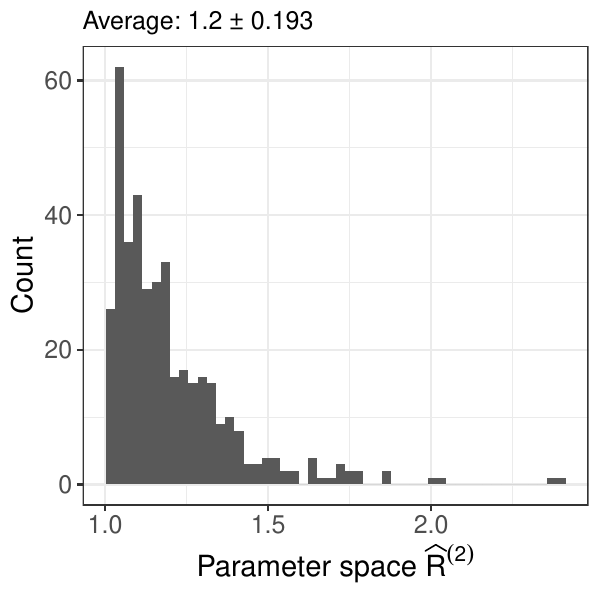}
        \caption{Airfoil (PS)}
    \end{subfigure}
    \begin{subfigure}{0.23\textwidth}
        \includegraphics[width=\textwidth]{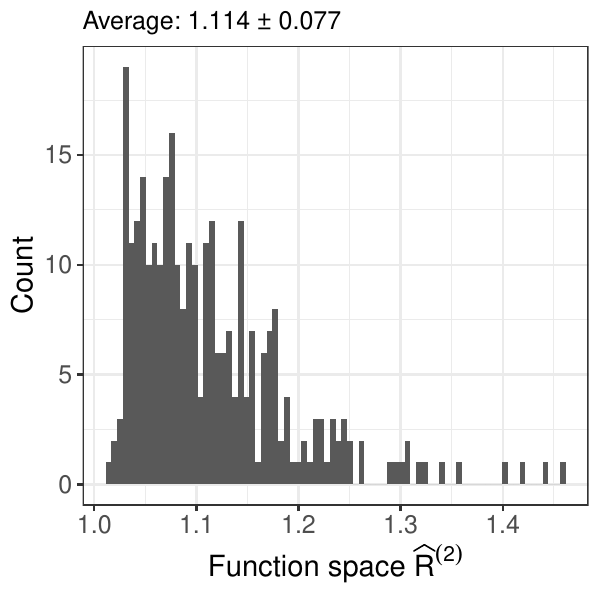}
        \caption{Airfoil (FS)}
    \end{subfigure}
    \begin{subfigure}{0.23\textwidth}
        \includegraphics[width=\textwidth]{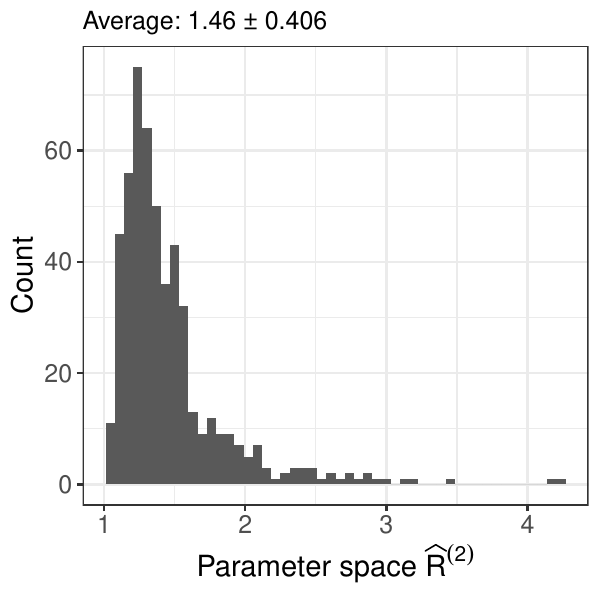}
        \caption{Bikesharing (PS)}
    \end{subfigure}
    \begin{subfigure}{0.23\textwidth}
        \includegraphics[width=\textwidth]{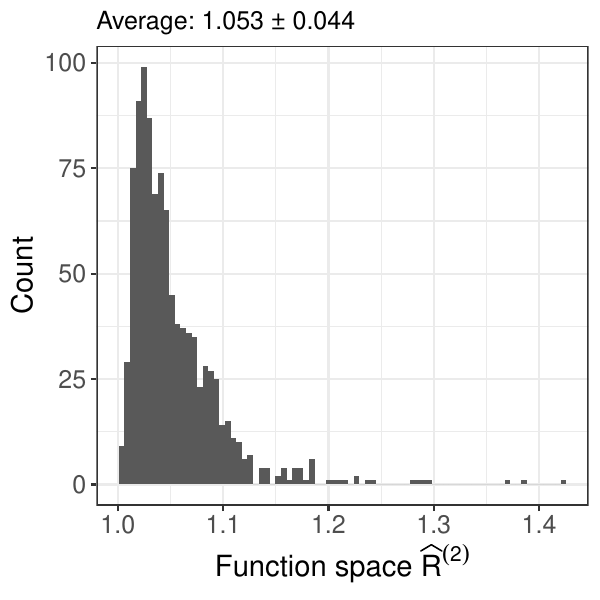}
        \caption{Bikesharing (FS)}
    \end{subfigure}

    \begin{subfigure}{0.23\textwidth}
        \includegraphics[width=\textwidth]{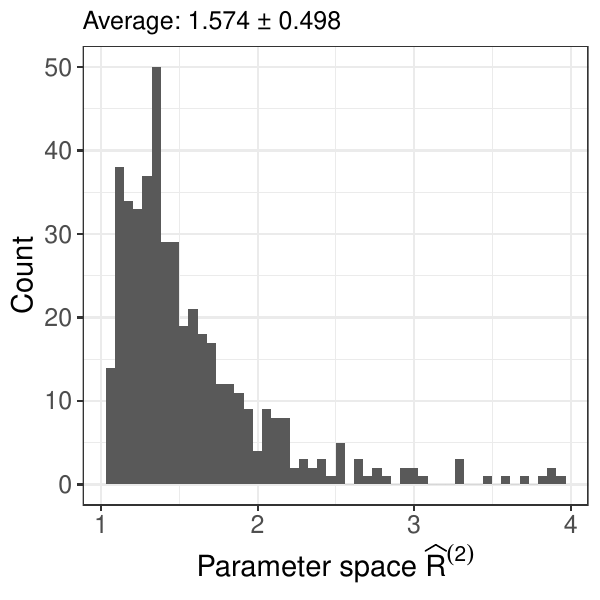}
        \caption{Concrete (PS)}
    \end{subfigure}
    \begin{subfigure}{0.23\textwidth}
        \includegraphics[width=\textwidth]{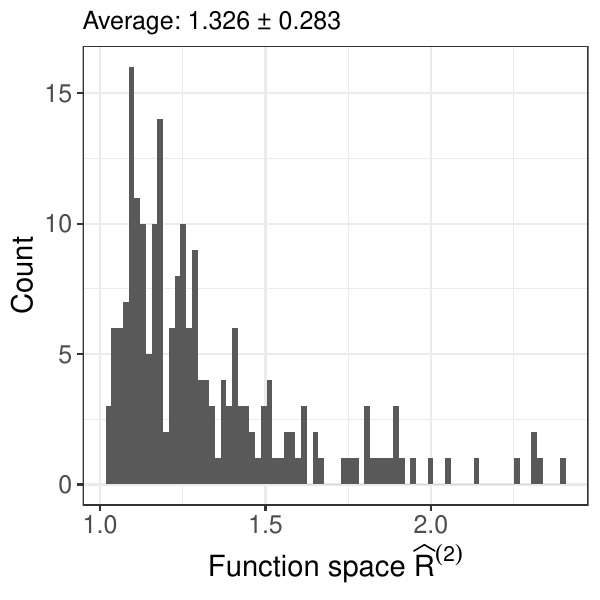}
        \caption{Concrete (FS)}
    \end{subfigure}
    \begin{subfigure}{0.23\textwidth}
        \includegraphics[width=\textwidth]{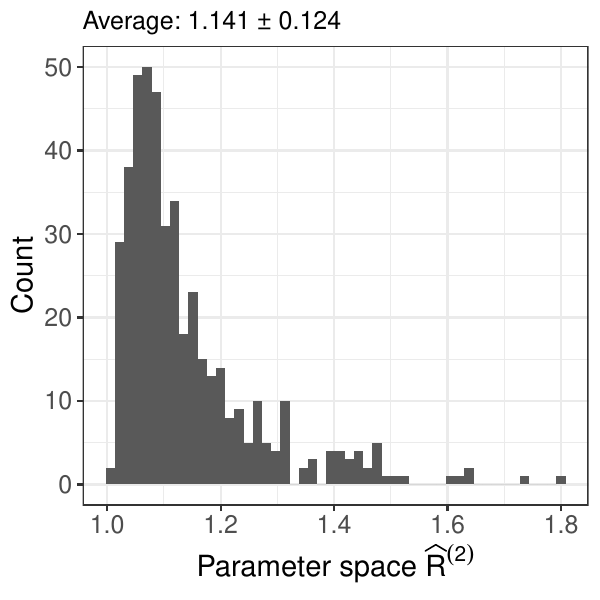}
        \caption{Energy (PS)}
    \end{subfigure}
    \begin{subfigure}{0.23\textwidth}
        \includegraphics[width=\textwidth]{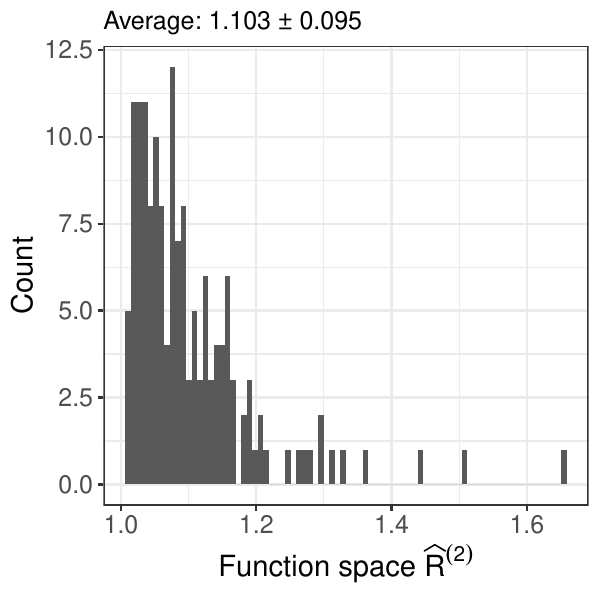}
        \caption{Energy (FS)}
    \end{subfigure}

    \begin{subfigure}{0.23\textwidth}
        \includegraphics[width=\textwidth]{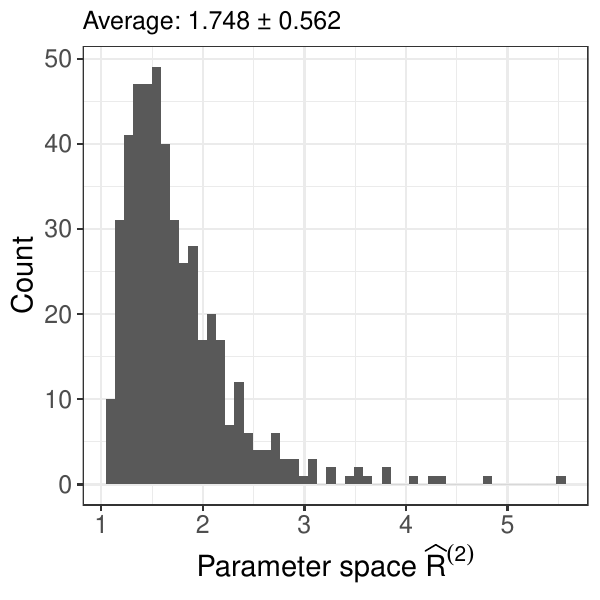}
        \caption{Protein (PS)}
    \end{subfigure}
    \begin{subfigure}{0.23\textwidth}
        \includegraphics[width=\textwidth]{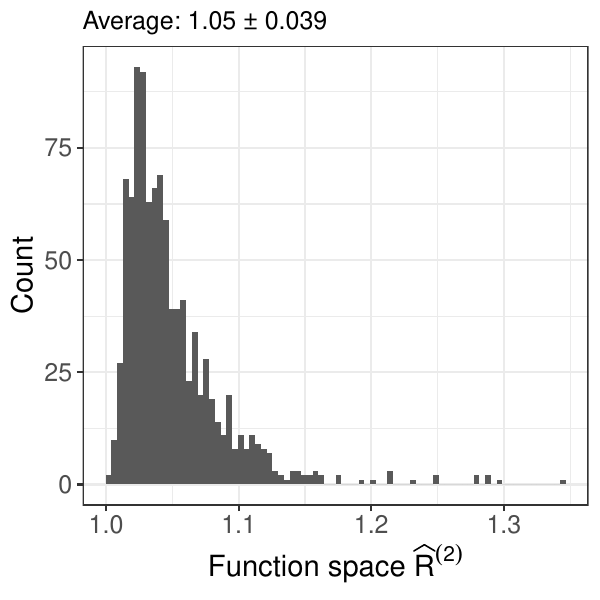}
        \caption{Protein (FS)}
    \end{subfigure}
    \begin{subfigure}{0.23\textwidth}
        \includegraphics[width=\textwidth]{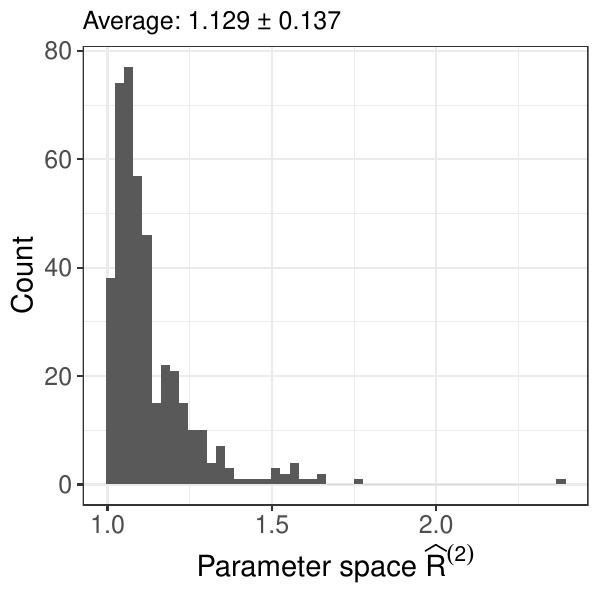}
        \caption{Yacht (PS)}
    \end{subfigure}
    \begin{subfigure}{0.23\textwidth}
        \includegraphics[width=\textwidth]{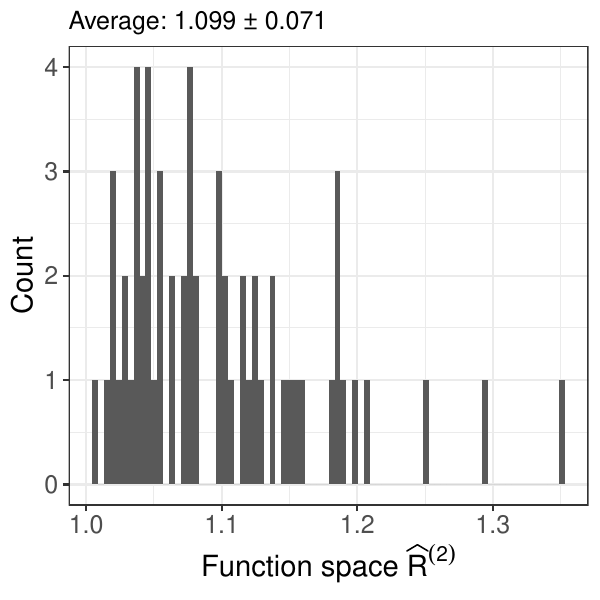}
        \caption{Yacht (FS)}
    \end{subfigure}
    \caption{Histograms, mean and standard deviation of the parameter space (PS) $\widehat{R}^{(2)}$  and function space (FS) $\widehat{R}^{(2)}_{\psi_{\text{PSC}}}$, evaluated for each parameter and test data entry respectively. For each data set, all fitted chains with better than LM RMSE performance are used. In all cases, the BNN consists of two hidden layers with 16 neurons each and tanh activation. NUTS with a warmup phase of 10k steps and 8k samples per chain as well as unit Gaussian priors are used.}
    \label{fig:rhat_hist}
\end{figure}

Each sampled parameter vector $\bm{\theta}^{(k,s)}$ and test data point $\bm{x}^\ast$ result in one corresponding conditional density function $p\left(\bm{y} | \bm{\theta}^{(k,s)}(\bm{x}^\ast)\right)$ from which we sample one observation $\bm{y}^{(k,s)}$, hence every model induces $K \cdot S$ samples of outcome values for each test data point. 
Therefore, it is straightforward to compute the $\widehat{R}^{(2)}_\psi$ in the output space diagnostic for each of the test points using the mapping $\psi_{\text{PSC}}(\bm{\theta}^{(k,s)}(\bm{x}^\ast)) \mapsto \bm{y}^{(k,s)}$ (where PSC stands for \emph{pointwise sample convergence}).
\citet{izmailov_2021_WhatArea} aggregate these $\widehat{R}^{(2)}_{\psi_{\text{PSC}}}$ values and report diagnostics using histograms or summary statistics. 
Such metrics can be interpreted as a measure of how stationary the samples are for a fixed evaluation point.
In other words, confident disagreement between chains (large difference in the expectation of the PPD and low variance within the chain) is penalized, which, as we have argued before, is unjustified for high multimodality.
We report these function space $\widehat{R}^{(2)}_{\psi_{\text{PSC}}}$  values histograms in \cref{fig:rhat_hist} across various data sets for well-performing models of the same architecture.

Another popular way of achieving a single diagnostic for function space convergence uses performance-related functions of the parameters $\bm{\theta}^{(k,s)}(\bm{x}^\ast)$ induced by the test data set \cite{fortuin2022bayesian}. We will call this approach \emph{functional convergence (FC)}. 
The performance function can, for instance, be the RMSE for a single posterior sample or the log pointwise likelihood $$\text{LPL}^{(k,s)} = \log \left(\frac{1}{n_{\text{test}}} \sum_{(\bm{x}^\ast,\bm{y}^\ast) \in \mathcal{D}_{\text{test}}} p \left(\bm{y}^\ast | \bm{\theta}^{(k,s)}(\bm{x}^\ast) \right) \right).$$
Now, we can calculate the $\widehat{R}^{(2)}_\psi$ with $\psi$ being the $\text{LPL}^{(k,s)}$, accounting for effects induced by between- and within-chain variance of the $\bm{\theta}^{(k,s)}$.
This notion of convergence can be interpreted as whether the chains have reached a stable state in terms of the performance metric expressed by the aggregation function.

\begin{figure}[!ht]
    \centering
    \begin{subfigure}{0.47\textwidth}
        \includegraphics[width=\textwidth]{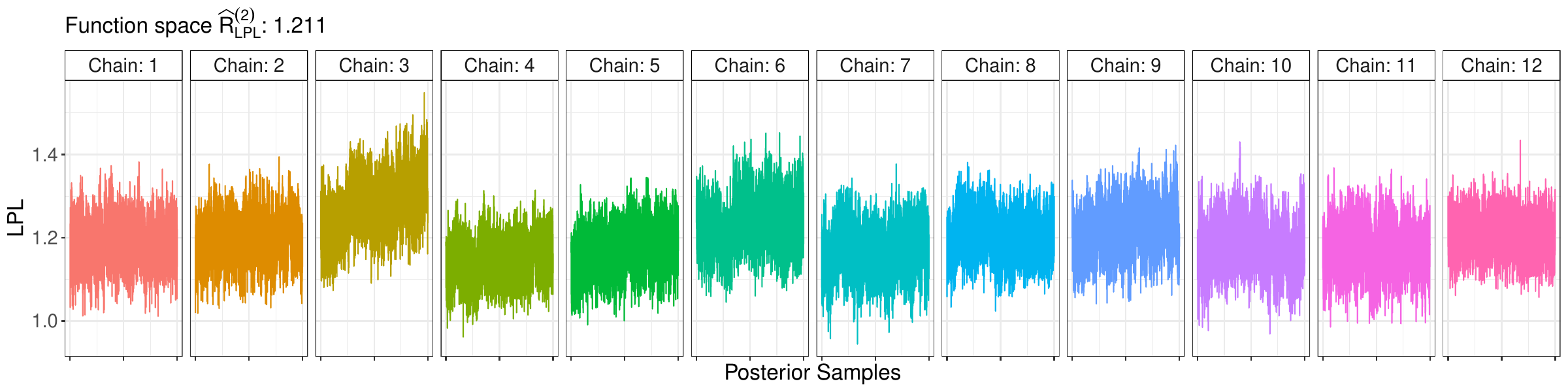}
        \caption{Airfoil}
    \end{subfigure}
    \begin{subfigure}{0.47\textwidth}
        \includegraphics[width=\textwidth]{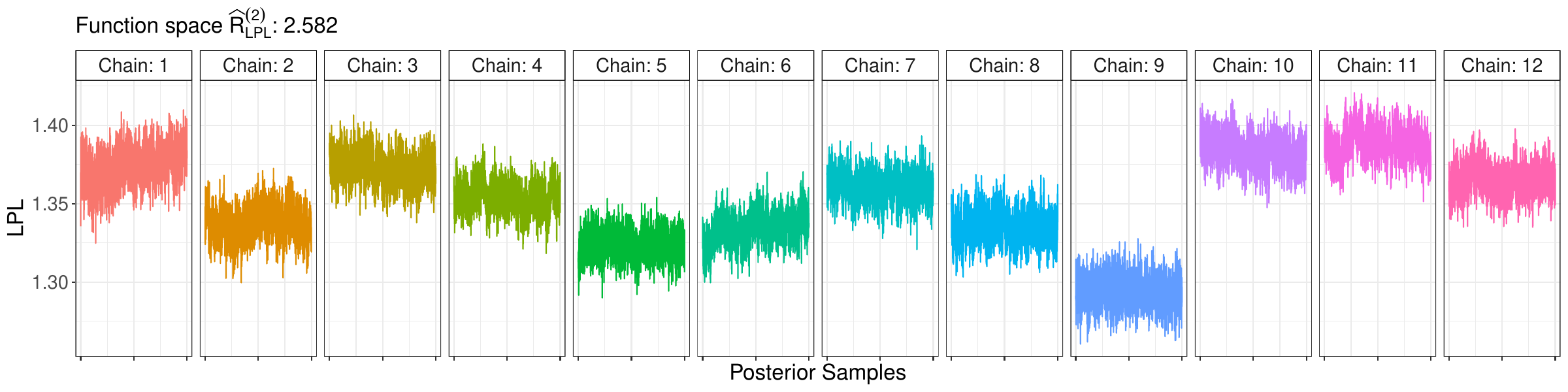}
        \caption{Bikesharing}
    \end{subfigure}

    \begin{subfigure}{0.47\textwidth}
        \includegraphics[width=\textwidth]{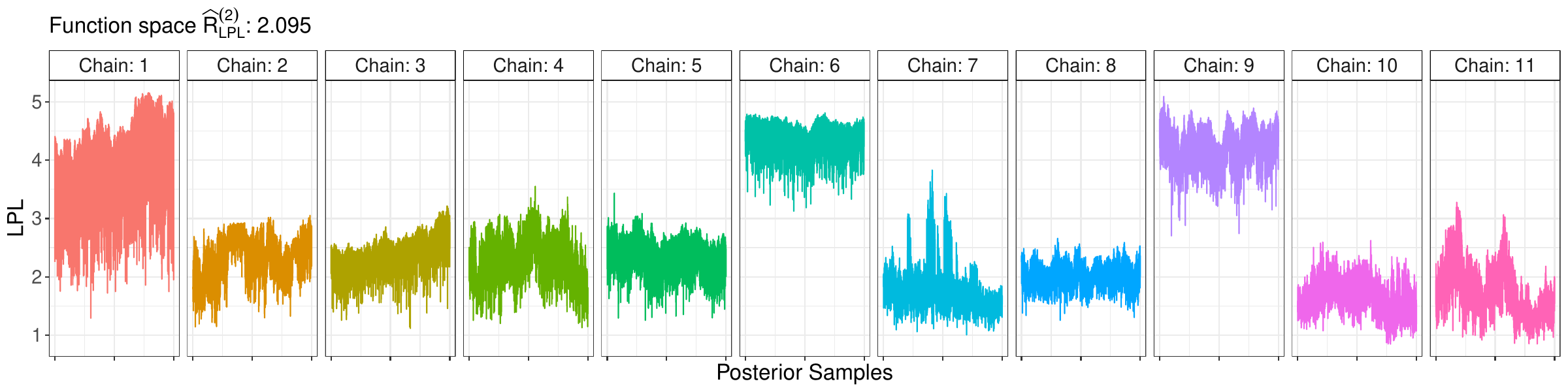}
        \caption{Concrete}
    \end{subfigure}
    \begin{subfigure}{0.47\textwidth}
        \includegraphics[width=\textwidth]{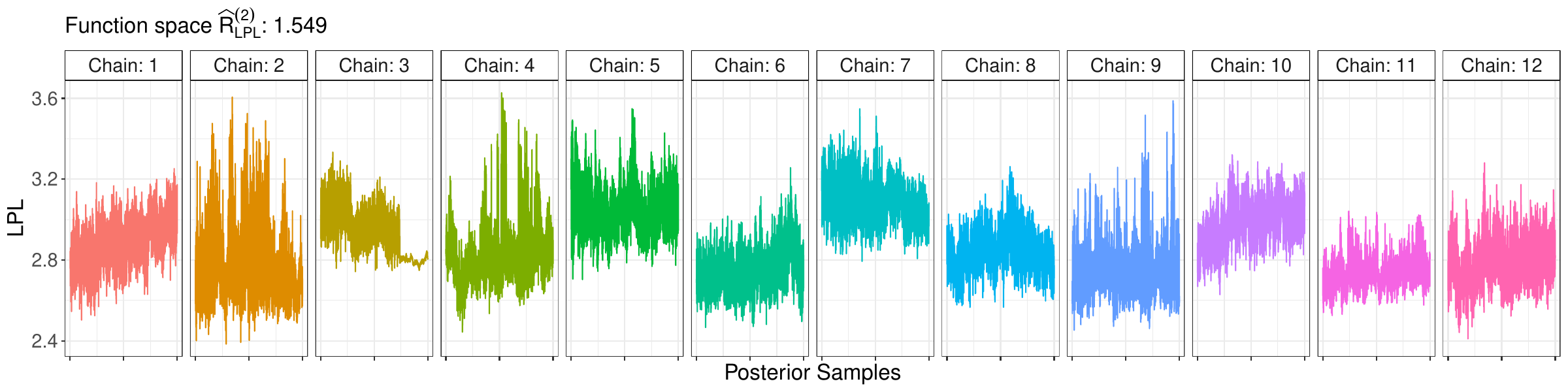}
        \caption{Energy}
    \end{subfigure}

    \begin{subfigure}{0.47\textwidth}
        \includegraphics[width=\textwidth]{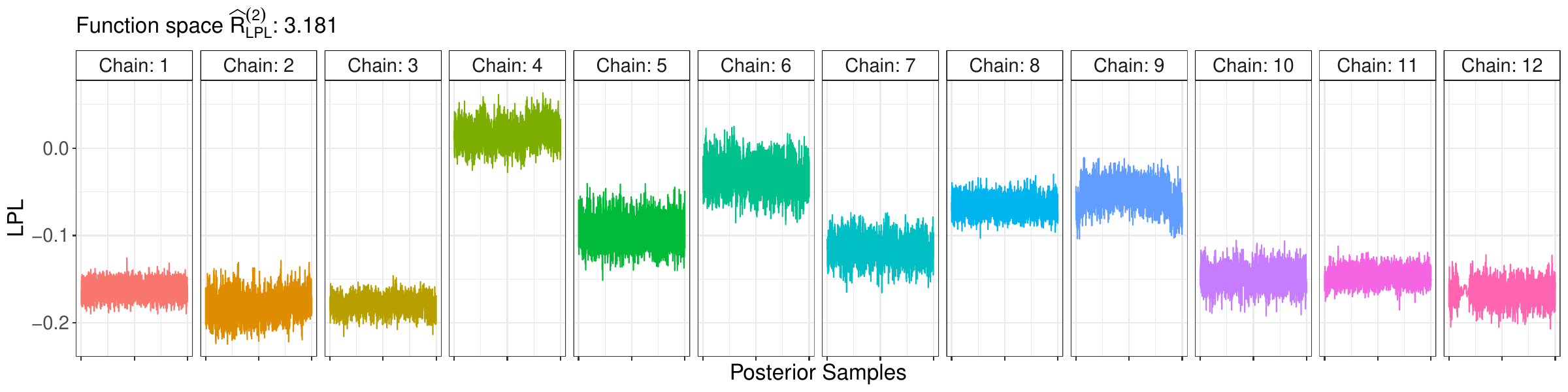}
        \caption{Protein}
    \end{subfigure}
    \begin{subfigure}{0.47\textwidth}
        \includegraphics[width=\textwidth]{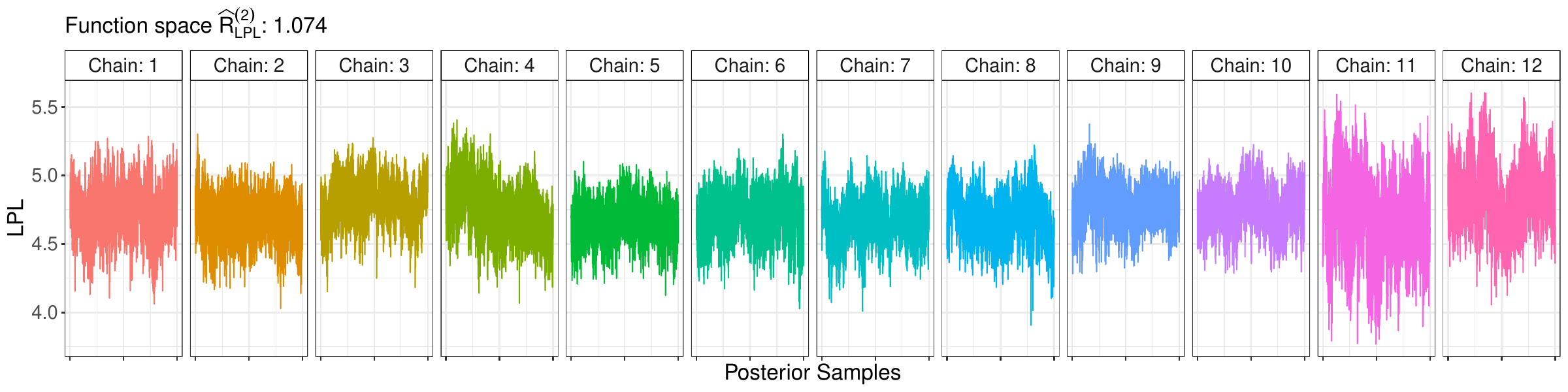}
        \caption{Yacht}
    \end{subfigure}
    \caption{The LPL chains that form the basis of the $\widehat{R}^{(2)}_{\text{LPL}}$ calculation for the functional convergence assessment. For each data set, all fitted chains with better than LM RMSE performance are used (colors and boxes). In all cases, the BNN consists of 2 hidden layers with 16 neurons each and tanh activation. NUTS with a warmup phase of 10k steps and 8k samples per chain (x-axis of the boxes) as well as unit Gaussian priors are used.}
    \label{fig:func_lpl}
\end{figure}

\begin{figure}[!ht]
    \centering
    \begin{subfigure}{0.47\textwidth}
        \includegraphics[width=\textwidth]{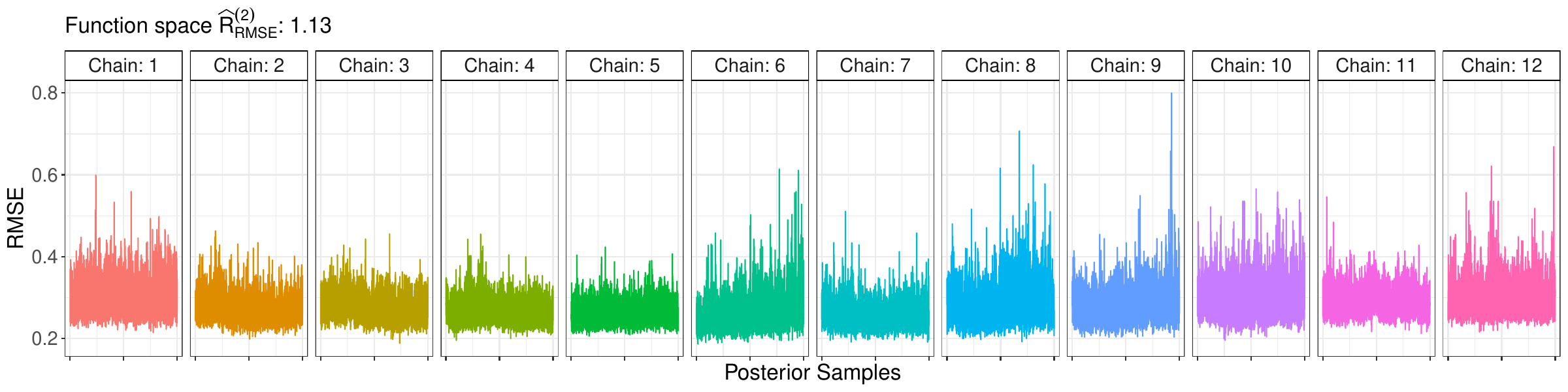}
        \caption{Airfoil}
    \end{subfigure}
    \begin{subfigure}{0.47\textwidth}
        \includegraphics[width=\textwidth]{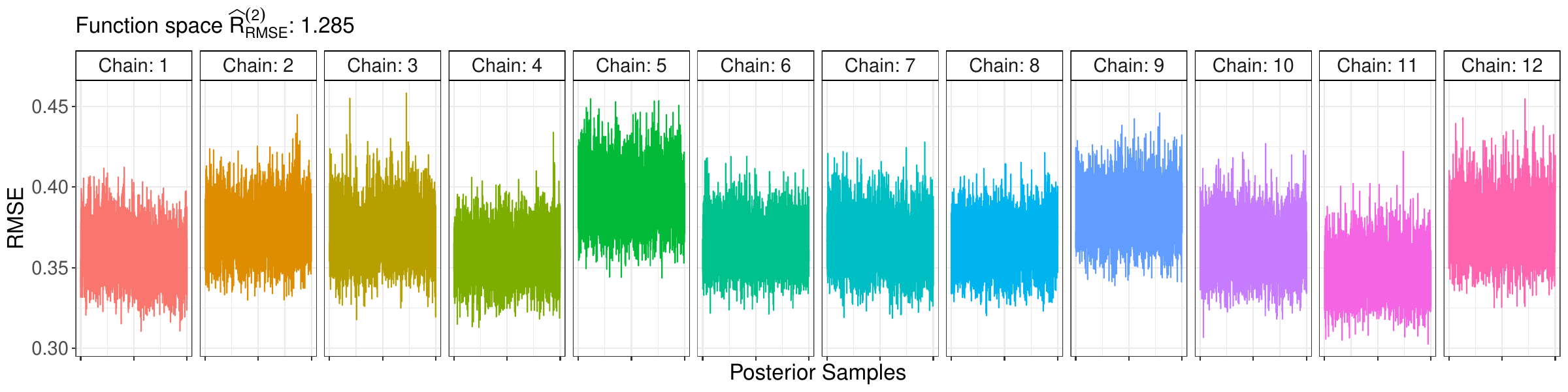}
        \caption{Bikesharing}
    \end{subfigure}

    \begin{subfigure}{0.47\textwidth}
        \includegraphics[width=\textwidth]{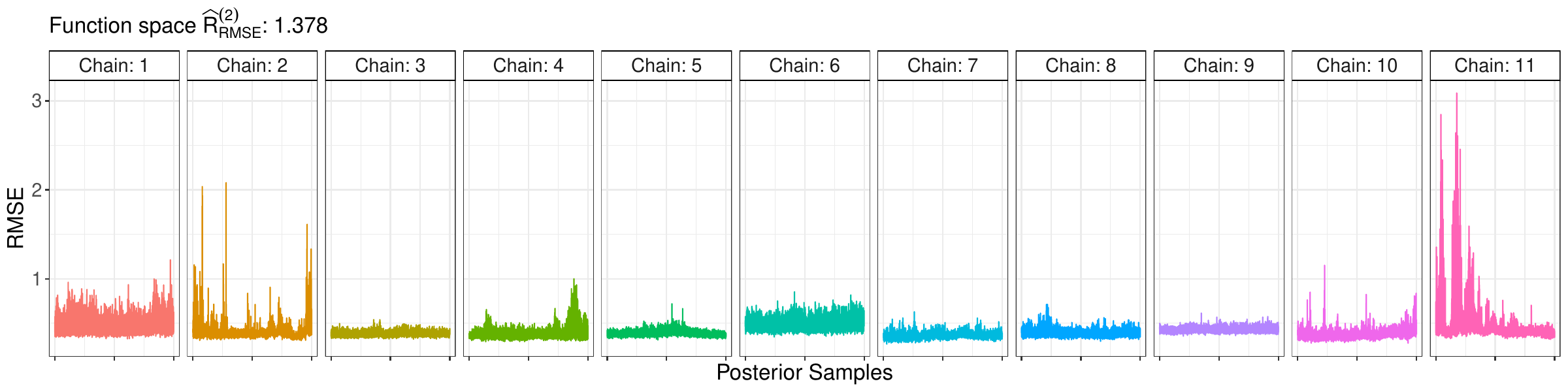}
        \caption{Concrete}
    \end{subfigure}
    \begin{subfigure}{0.47\textwidth}
        \includegraphics[width=\textwidth]{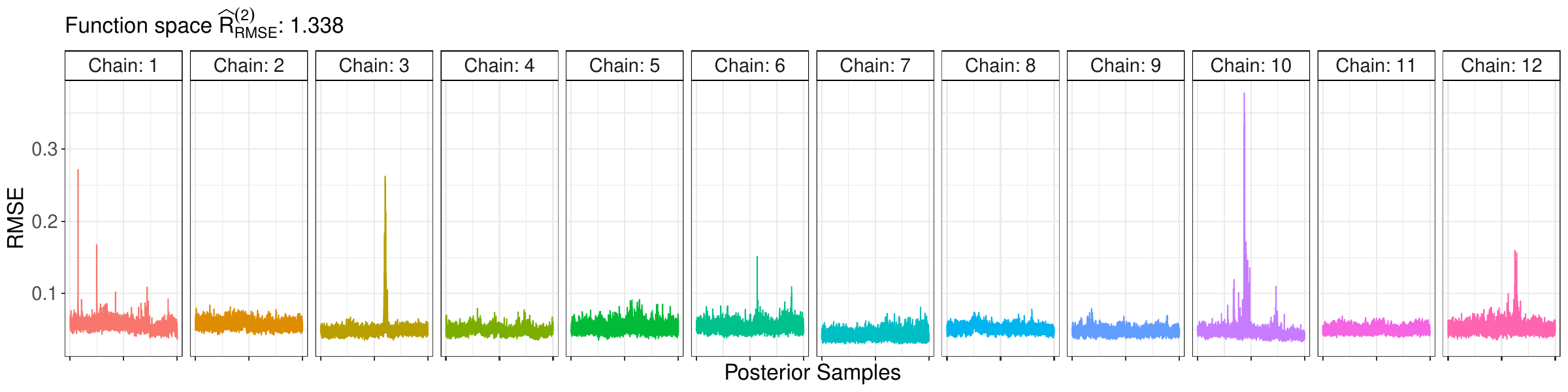}
        \caption{Energy}
    \end{subfigure}

    \begin{subfigure}{0.47\textwidth}
        \includegraphics[width=\textwidth]{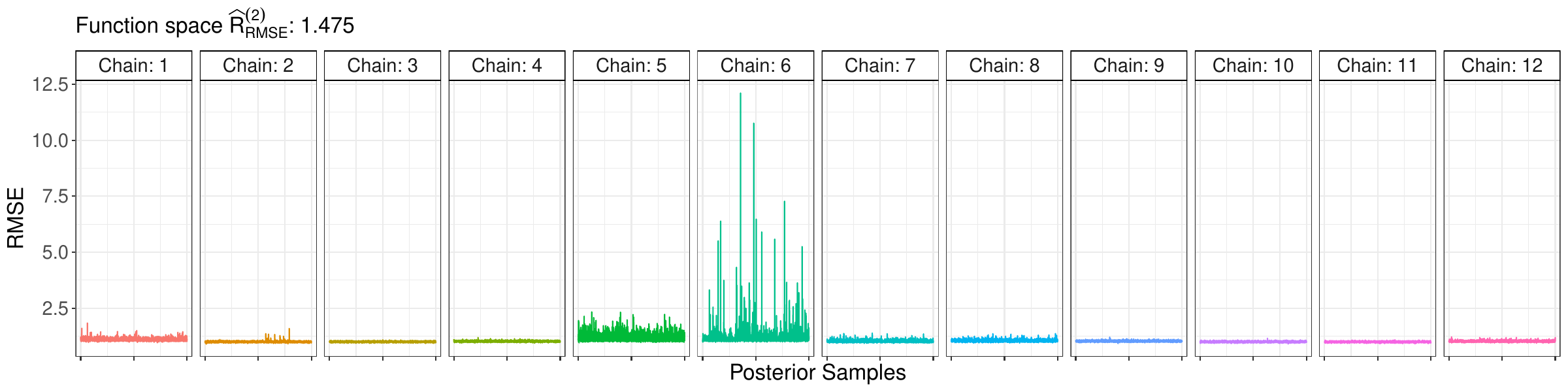}
        \caption{Protein}
    \end{subfigure}
    \begin{subfigure}{0.47\textwidth}
        \includegraphics[width=\textwidth]{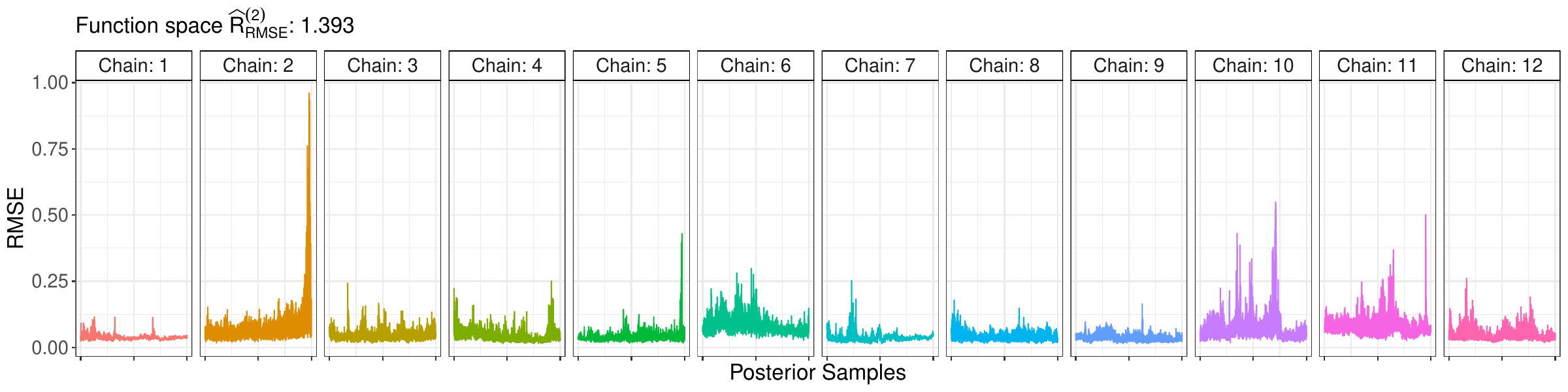}
        \caption{Yacht}
    \end{subfigure}
    \caption{The RMSE chains that form the basis of the $\widehat{R}^{(2)}_{\text{RMSE}}$ calculation for the functional convergence assessment. For each data set, all fitted chains with better than LM RMSE performance are used (colors and boxes). In all cases, the BNN consists of 2 hidden layers with 16 neurons each and tanh activation. NUTS with a warmup phase of 10k steps and 8k samples per chain (x-axis of the boxes) as well as unit Gaussian priors are used.}
    \label{fig:func_rmse}
\end{figure}

We show in our experiments ($\text{LPL}^{(k,s)}$ and $\text{RMSE}^{(k,s)}$ chains are provided in Figures \ref{fig:func_lpl} and \ref{fig:func_rmse}, resp.) that FC heavily depends on the choice of the function. RMSE-based FC, for instance, does not account for the uncertainty of predictions explicitly. In our experiments, it thus shows better $\widehat{R}^{(2)}_{\text{RMSE}}$ values (rather close to $1.1$) compared to often considerably worse $\widehat{R}^{(2)}_{\text{LPL}}$ values for most data sets. This is because the chains in function space indeed show different levels of quality in both prediction and uncertainty quantification. Thus, even with rough filtering of very bad chains before the FC calculation, we obtain low within-chain variance and high between-chain variance in many cases, resulting in a $\widehat{R}^{(2)}_\psi$ larger than even $1.1$. This clearly shows that even for overall good models there are differences between the chains concerning their quality of prediction and uncertainty quantification and, again, proves our point that current metrics are not meaningful in measuring convergence. 

\textbf{Expanding-window function space convergence.}
Convergence diagnostics usually evaluate chain mixing post-hoc.
We propose an expanding-window approach, enabled by chain-wise analysis, that allows us to assess chain-wise convergence and between-chain agreement in function space at any time point during the sampling process.
In particular, this entails the option of online resource management by early stopping of chains: when the sampler shows evidence of being stuck and risks wasting compute, the chain can be restarted.

For this, following \citet{gelman2014a} and \citet{ecml}, we choose the log posterior predictive density (LPPD) over a test set, defined as
\begin{align} \label{eq:lppd}
    \text{LPPD} = \frac{1}{n_{\text{test}}} \sum_{(\bm{y}^\ast, \bm{x}^\ast) \in \mathcal{D}_{\text{test}}} \log \left(\frac{1}{K \cdot S} \sum_{k=1}^K \sum_{s=1}^{S} p \left(\bm{y}^\ast | \bm{\theta}^{(k,s)}(\bm{x}^\ast)\right)\right).
\end{align}
The LPPD quantifies how well, on average, the predictive distribution covers the observed label.
As we can evaluate the LPPD chain-wise (using $K=1$ in \eqref{eq:lppd}) after every sample update, 
this motivates the 
chain-wise expanding-window LPPD for any time point $l \in \{1,\ldots,S\}$:
\begin{align}\label{eq:lppdcum}
    \text{LPPD}_l = \frac{1}{n_{\text{test}}} \sum_{(\bm{y}^\ast, \bm{x}^\ast) \in \mathcal{D}_{\text{test}}} \log \left(\frac{1}{l} \sum_{\tilde{l} \leq l} p\left(\bm{y}^\ast | \bm{\theta}^{(\tilde{l})}(\bm{x}^\ast)\right)\right).
\end{align}

In order to assess convergence, we can define an $\epsilon$-threshold and window $\varpi \in \mathbb{N}$ such that 
\begin{align}
    \left|\left(\frac{1}{\varpi} \sum_{l - \varpi \leq \tilde{l} < l} \text{LPPD}_{\tilde{l}}\right) - \text{LPPD}_l\right| < \epsilon.
\end{align}

As soon as convergence or a maximum number of samples is reached, the sampling of the chain can be terminated to start a new chain. In case of non-convergence, the chain is discarded. This is especially valuable if one aims for more chains rather than more samples, as suggested in Section \ref{sec:multiplechains}, given a certain computation budget. Another advantage of this metric is the possibility to compare the LPPD across chains to, e.g., detect outliers. As such, the notion of between-chain variance is not lost. In contrast to FC, our method does not only evaluate a pointwise performance but reflects the current overall performance by using the running mean.

\begin{figure}[!ht]
    \centering
    \begin{subfigure}{0.3\textwidth}
        \includegraphics[width=\textwidth]{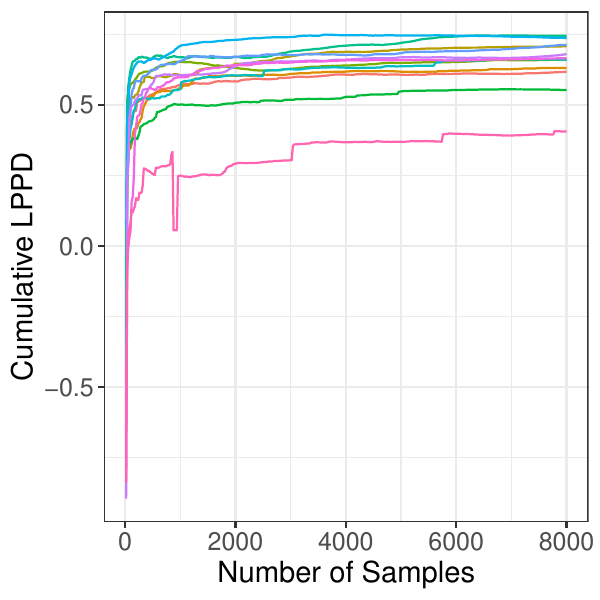}
        \caption{Airfoil}
    \end{subfigure}
    \begin{subfigure}{0.3\textwidth}
        \includegraphics[width=\textwidth]{icml2024/img/bikesharing_3layers_cumulative_lppd.pdf}
        \caption{Bikesharing}
    \end{subfigure}
    \begin{subfigure}{0.3\textwidth}
        \includegraphics[width=\textwidth]{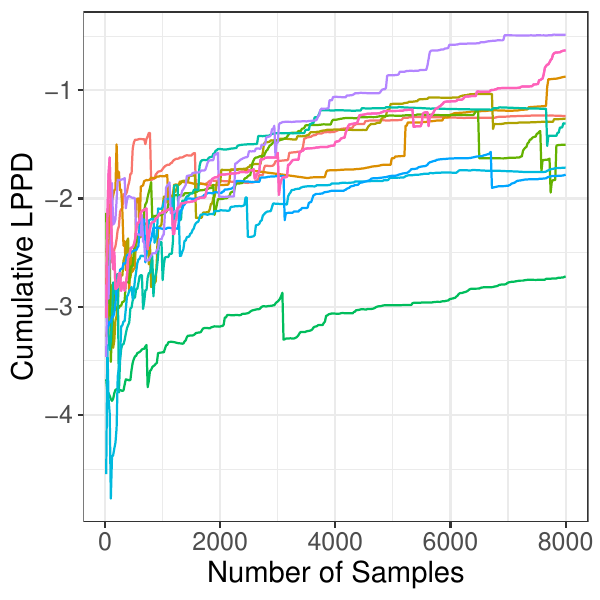}
        \caption{Concrete}
    \end{subfigure}
    \begin{subfigure}{0.3\textwidth}
        \includegraphics[width=\textwidth]{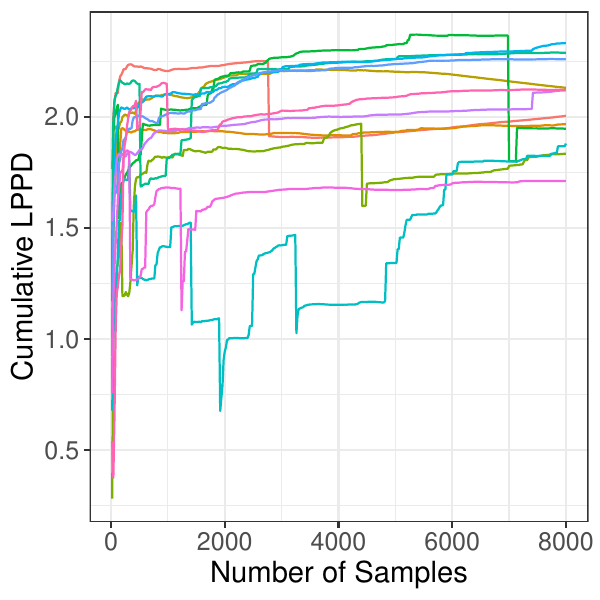}
        \caption{Energy}
    \end{subfigure}
    \begin{subfigure}{0.3\textwidth}
        \includegraphics[width=\textwidth]{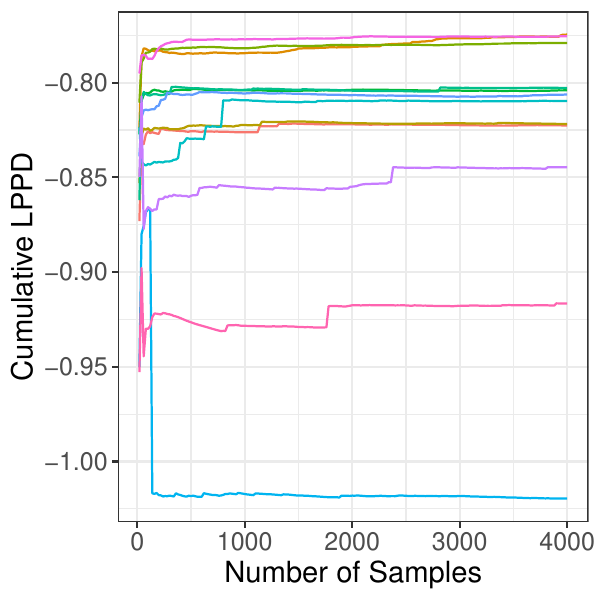}
        \caption{Protein}
    \end{subfigure}
    \begin{subfigure}{0.3\textwidth}
        \includegraphics[width=\textwidth]{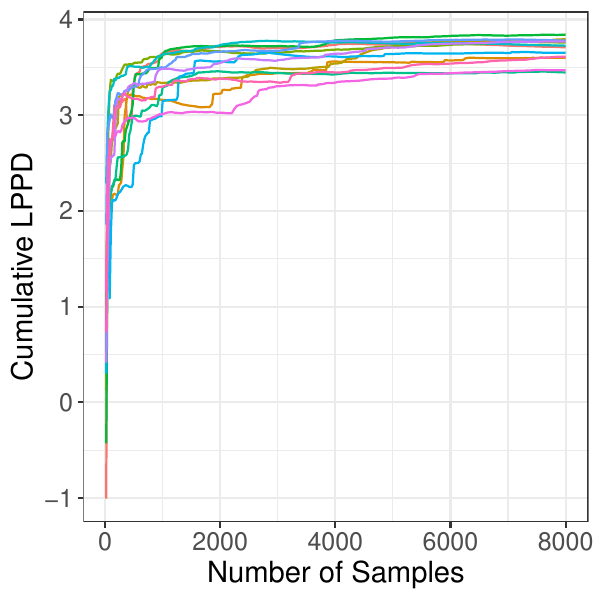}
        \caption{Yacht}
    \end{subfigure}
    \caption{Cumulative LPPD across data sets for the proposed expanding-window convergence assessment. For each data set, all fitted chains with better-than-LM RMSE performance are used (colors and boxes). In all cases, the BNN consists of two hidden layers with 16 neurons each and tanh activation. NUTS with a warmup phase of 10,000 steps and 8,000 samples per chain (x-axis) as well as unit Gaussian priors are used.}
    \label{fig:func_cum_lppd}
\end{figure}

Empirically we observe convergence for most chains, as displayed in~\cref{fig:func_cum_lppd}. Moreover, the convergence to different levels of LPPD is notably visible, especially for the larger data sets \texttt{bikesharing} and \texttt{protein}. The chain-wise expanding-window LPPD usually shows convergence already after a few samples per chain. This implies great benefits in sampling efficiency, allowing to free computational resources early on in the sampling process and to run a large number of chains (almost) in parallel.

\clearpage

\section{Additional Experiments} \label{app:additional_exp}

\subsection{General Feasibility}

\cref{tab:rmse_bench_full} contains the results reported in \cref{tab:rmse_bench} with additional standard deviations in parentheses. The architecture is a two-hidden layer MLP with 16 neurons in each hidden layer. The BNNs are sampled using 10,000 warmup steps of NUTS and 8,000 samples for each of the 12 chains. Only chains that proved to be better than the baseline LM are considered for the calculation (here always at least 10). The tanh activation function and unit Gaussian priors are applied. Experiments reported in \cref{tab:prop_blm} also use unit Gaussian priors. Note that for the Random Forest baseline an extensive HPO search using optuna \citep{optuna2019} is performed. Thus, the comparison is rather conservative as we do not perform a neural architecture search for the network-based models. The relevant yardstick in this comparison are the Deep Ensembles which share the same inductive biases as the BNNs.

\begin{table}[H] 
    \centering
    \small
    \setlength{\tabcolsep}{2pt}
        \caption{Average RMSE for different models over different data sets. All networks have two hidden layers with 16 neurons each.. The best method per data set is highlighted in bold.}
    \label{tab:rmse_bench_full}
    \vskip 0.15in
       \begin{center}
    \begin{sc}
    \resizebox{0.85\textwidth}{!}{
    \begin{tabular}{lrrrrrr}
        \toprule
        \multicolumn{1}{l}{data set} & LM & RF (tuned) & DNN & DE & BNN (RS) & BNN \\ 
        \midrule\addlinespace[2.5pt]
airfoil & 0.716 (0.022) & 0.255 (0.005) & 0.252 (0.024) & 0.239 (0.023) & 0.25 (0.009) & \textbf{0.182} (0.009) \\
bikesharing & 0.790 (0.028) & \textbf{0.231} (0.007) & 0.374 (0.002) & 0.365 (0.003) & 0.362 (0.013) & 0.253 (0.036) \\
concrete & 0.630 (0.018) & 0.304 (0.021) & 0.317 (0.015) & 0.282 (0.018) & 0.554 (0.158) & \textbf{0.258} (0.020) \\
energy & 0.274 (0.020) & 0.051 (0.009) & 0.048 (0.006) & 0.043 (0.007) & 0.062 (0.006) & \textbf{0.037 }(0.006) \\
protein & 0.863 (0.028) & \textbf{0.581} (0.004) & 0.804 (0.004) & 0.803 (0.004) & 1.077 (0.085) & 0.716 (0.035) \\
yacht & 0.612 (0.068) & 0.072 (0.015) & 0.108 (0.032) & 0.103 (0.031) & 0.032 (0.007) & \textbf{0.022} (0.007) \\
        \bottomrule
    \end{tabular}
    }
    \end{sc}
    \end{center}
\end{table}

\begin{figure}[H]
    \centering
    \includegraphics[width = 0.4\textwidth]{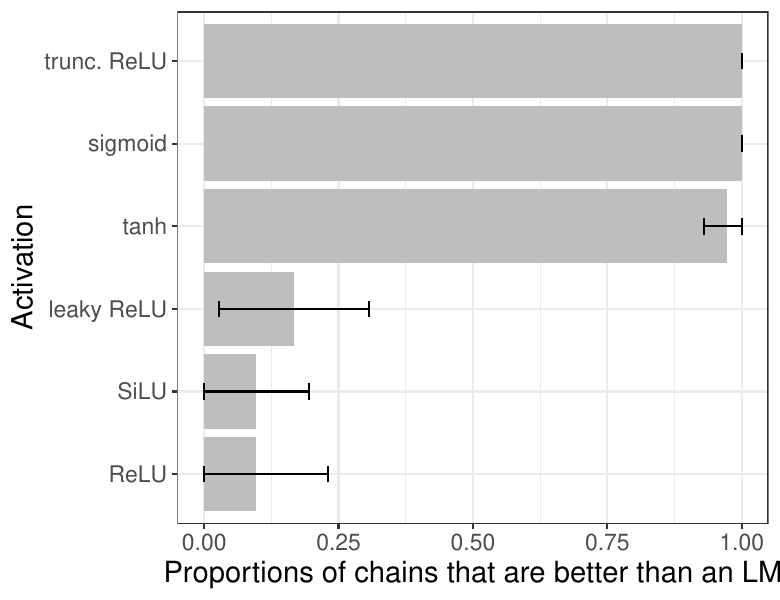}
    \caption{Proportions of BNN chains with better performance than an LM separated for different activation functions. Each proportion is the average of 72 experiments with 3 different train-test splits of the two data sets \texttt{airfoil} and \texttt{energy}, 1,000 warmup iterations of NUTS and 12 chains each with 1,000 posterior samples.}
    \label{fig:allactiv}
\end{figure}

\cref{fig:allactiv} shows the share of chains that are able to outperform the lower baseline of an LM across different activation functions, clearly pointing towards a problem of unbounded activation functions. 
Notably, only the truncated ReLU is able to avoid the dying sampler problem, whereas smoothed ReLU versions, such as SiLU and leaky ReLU, offer very little mitigation.

In \cref{tab:comparison_nuts_hm}, reflecting the setting of Table 5 in \citet{izmailov_2021_WhatArea}, we contrast the proportion of LM-beating chains produced by HMC and NUTs for ReLU, SiLU and tanh activations. We consider a single-hidden layer network with 50 neurons and a Gaussian prior with standard deviation 0.1
HMC is used with fixed step size of $10^{-5}$, trajectory length $\frac{\pi * 0.1}{2}$ and 10 warmup samples as suggested in \citet{izmailov_2021_WhatArea}. The NUTS sampler runs with default settings and a warmup length of 5k steps. For both samplers 400 samples are drawn from 12 independent chains over 3 random train-test splits.
Again, tanh activations produce fewer below-baseline chains when NUTS is used, whereas HMC with fixed setting essentially fails in all settings.

\begin{figure}[!ht]
    \centering
    \begin{subfigure}{0.45\textwidth}
        \includegraphics[width=\textwidth]{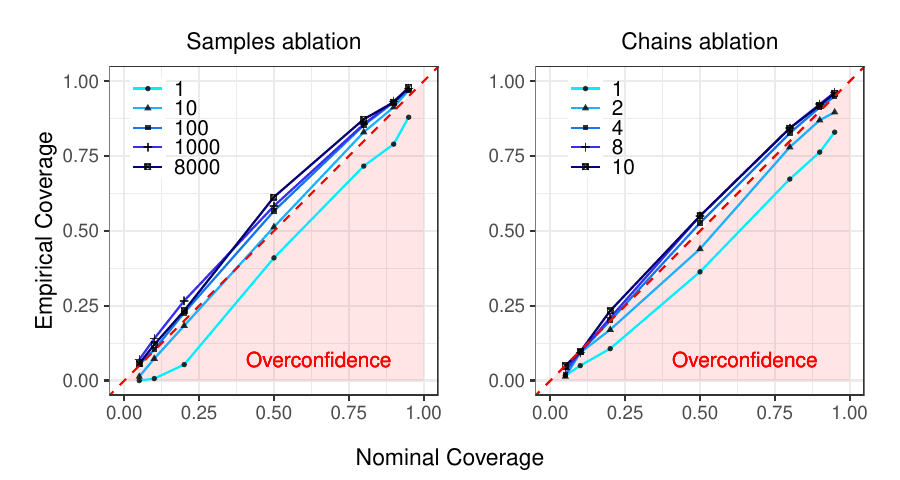}
        \caption{Airfoil}
    \end{subfigure}
    \begin{subfigure}{0.45\textwidth}
        \includegraphics[width=\textwidth]{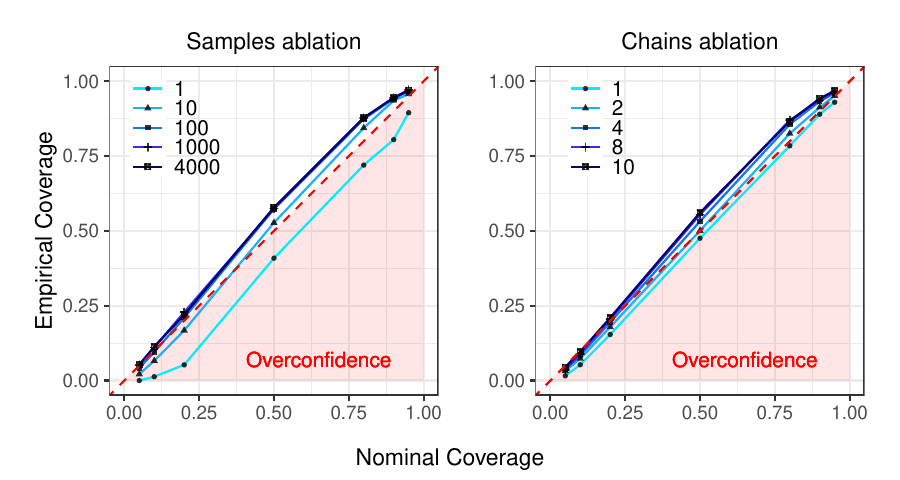}
        \caption{Bikesharing}
    \end{subfigure}
    \begin{subfigure}{0.45\textwidth}
        \includegraphics[width=\textwidth]{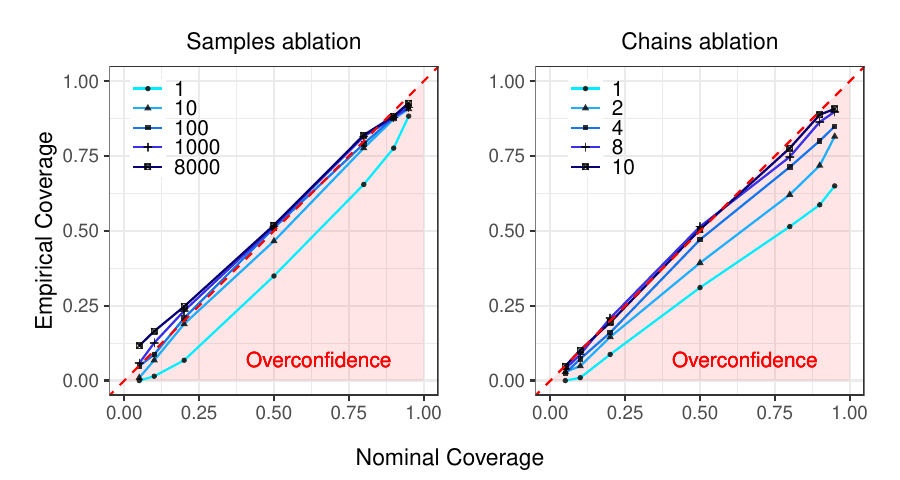}
        \caption{Concrete}
    \end{subfigure}
    \begin{subfigure}{0.45\textwidth}
        \includegraphics[width=\textwidth]{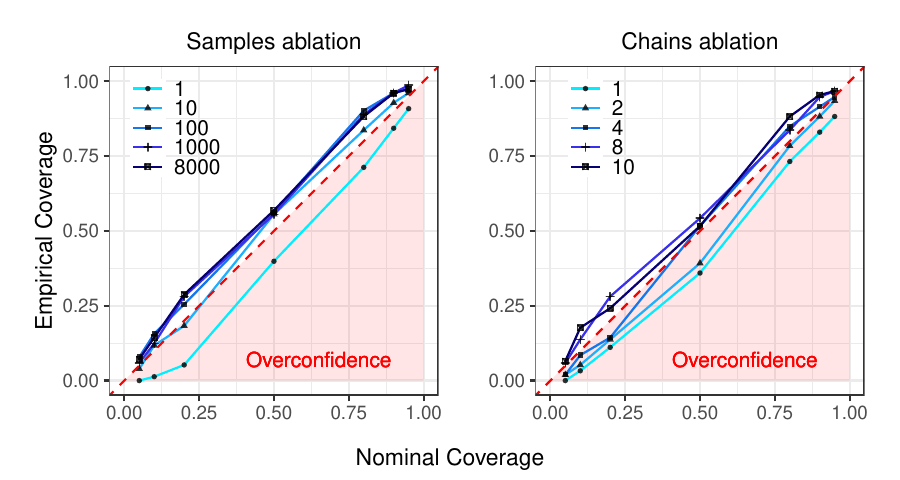}
        \caption{Energy}
    \end{subfigure}
    \begin{subfigure}{0.45\textwidth}
        \includegraphics[width=\textwidth]{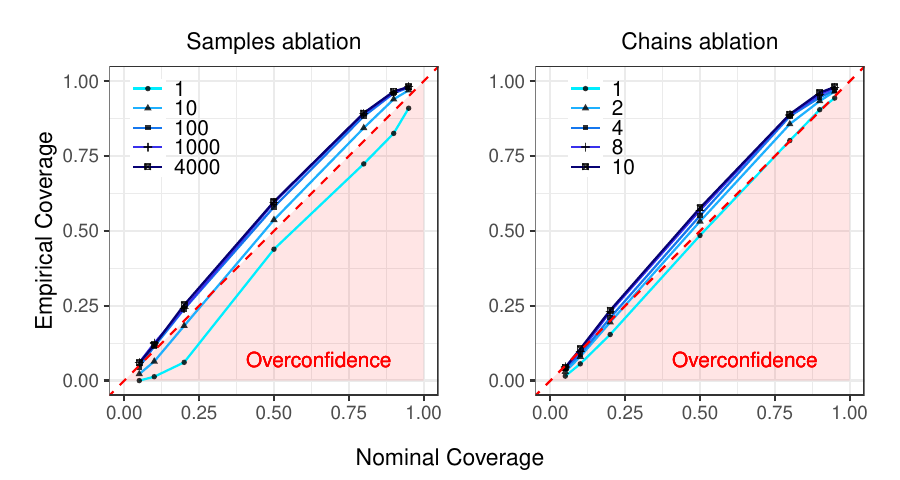}
        \caption{Protein}
    \end{subfigure}
    \begin{subfigure}{0.45\textwidth}
        \includegraphics[width=\textwidth]{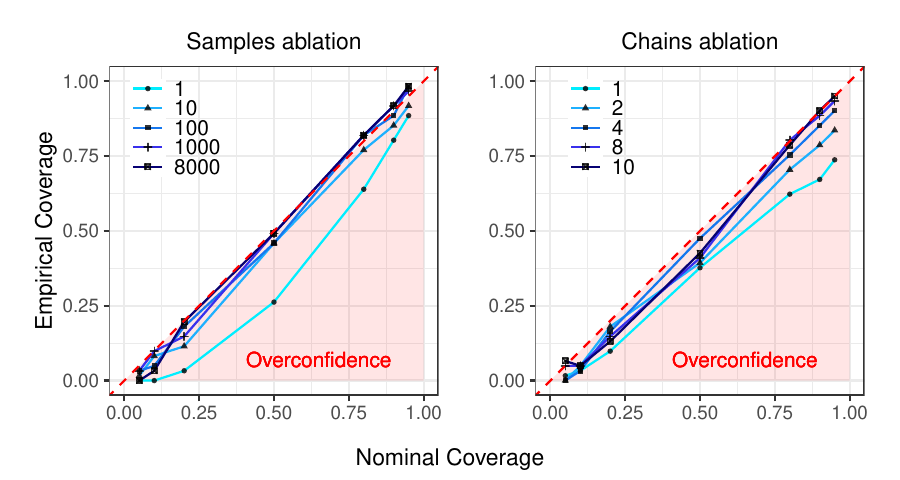}
        \caption{Yacht}
    \end{subfigure}
    \caption{Nominal vs.~empirical coverage of posterior credibility intervals for different numbers of samples using 10 chains everywhere (left) and different numbers of chains using 100 samples each (right) across various data set, NUTS with 10,000 warmup, two hidden layers of 16 neurons each, tanh activation, and unit Gaussian priors.}
    \label{fig:coverage_all}
\end{figure}

\textbf{Predictive coverage}. 
Bayesian inference enables the computation of predictive credibility intervals, which should be calibrated and cover the true prediction.
\cref{fig:coverage_all} shows the coverage of credibility intervals for varying amounts of samples (left) and chains (right). 
For a calibrated model (represented by the diagonal), the observed labels fall within the $\alpha \%$ credibility interval in $\alpha \%$ of cases.
With increasing chains and samples, we can see a trend from overconfidence to close-to-nominal coverage for all data sets. While \cref{fig:coverage_all} covers the nominal coverage levels $(0.05, 0.1, 0.2, 0.5, 0.8, 0.9, 0.95)$ for a broad overview, the trend observed is even more accentuated for very small and large coverage levels.

\begin{figure}[!ht]
    \centering
    \begin{subfigure}{\textwidth}
        \includegraphics[width=\textwidth]{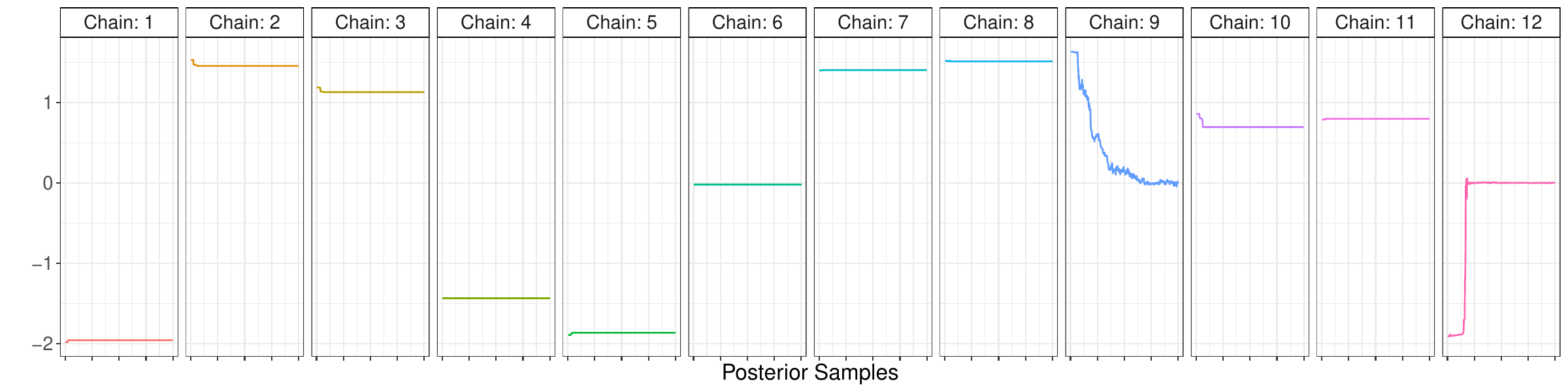}
        \caption{Random first layer weight}
    \end{subfigure}
    \begin{subfigure}{\textwidth}
        \includegraphics[width=\textwidth]{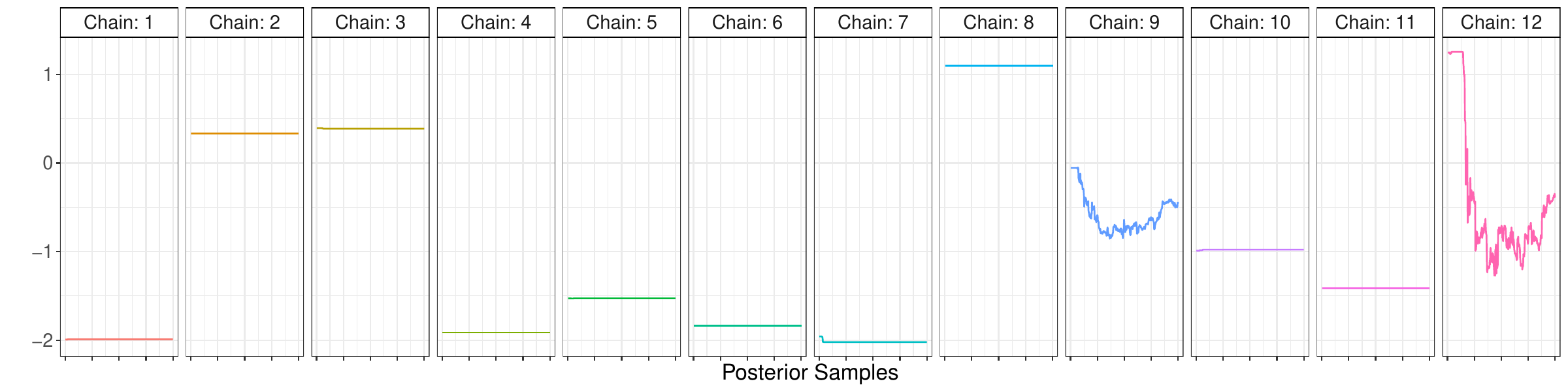}
        \caption{Random second layer weight}
    \end{subfigure}
    \begin{subfigure}{\textwidth}
        \includegraphics[width=\textwidth]{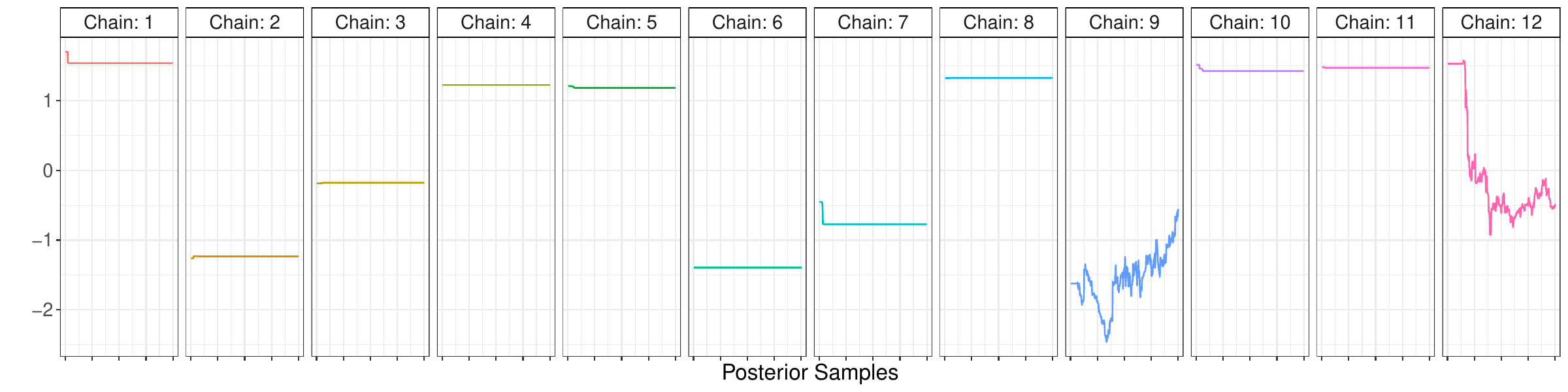}
        \caption{Random thrid layer weight}
    \end{subfigure}
    \begin{subfigure}{\textwidth}
        \includegraphics[width=\textwidth]{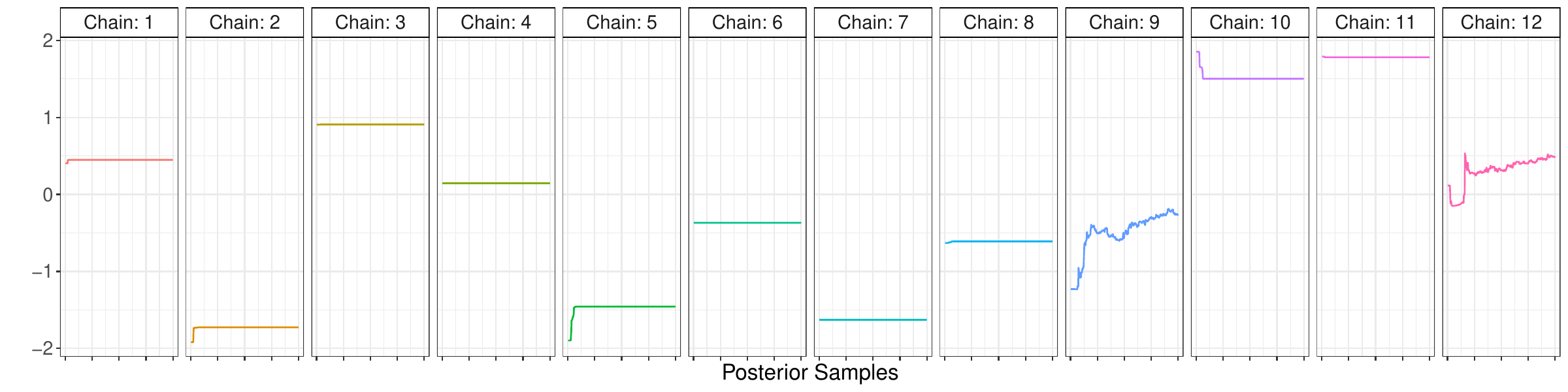}
        \caption{Random fourth (last) layer weight}
    \end{subfigure}
    \caption{
    Traceplots of a randomly selected parameter during the sampling phase of a dying NUTS sampler. The BNN has three hidden layer of ten neurons each, uses unit Gaussian priors, ReLU activation and is sampled for the \texttt{energy} data set with 1,000 NUTS warmup samples and 12 chains. The first 400 steps of the sampling are displayed (x-axis).}
    \label{fig:stuck_traces}
\end{figure}

\paragraph{Dying NUTS sampler details.}
In the cases where the NUTS sampler ends up in chains inducing utterly bad models--with, e.g., RMSE values far worse than a simple LM---a robust pattern emerges.
The sampler first tries larger step sizes but then, as the acceptance probability stays at practically zero, the step size and number of steps are quickly reduced. 
As a result, the sampler remains close to the starting values.
With the step size now contracted to a very small value, however, the acceptance probability rises again, in effect causing the sampler to move around the starting value in tiny steps.
Since this unfortunate state is attained fast during the warmup, the sampler only repeats samples from a fraction of the parameter space close to the initial values.
Obviously, those do not result in any meaningful uncertainty quantification and almost always fail to provide even close-to-good predictive capabilities. A visual example of such a warmup phase is provided in \cref{fig:stuck_traces}. Evidently most chains have contracted/died already after fewer than 100 warmup steps.

\begin{table}[H] 
\caption{The proportion of chains resulting in better RMSE performance than the LM. HMC is run with 10 warmup samples, NUTS with 5k warmup samples. Values are averaged across 12 chains and 3 replications. } \label{tab:comparison_nuts_hm}
   \vskip 0.15in
   \begin{center}
\begin{small}
\begin{sc}
\begin{tabular}{lllrr}
\toprule
\multicolumn{1}{l}{data set} & Activation & \multicolumn{1}{l}{Sampler} & Proportion & SD \\
\midrule\addlinespace[2.5pt]
\multirow{6}{*}{HMC} & \multirow{2}{*}{ReLU} & concrete & $0.03$ & $0.05$ \\
                     &                      & energy & $0.03$ & $0.05$ \\
                     & \multirow{2}{*}{SiLU} & concrete & $0.03$ & $0.05$ \\
                     &                      & energy & $0.00$ & $0.00$ \\
                     & \multirow{2}{*}{tanh} & concrete & $0.00$ & $0.00$ \\
                     &                      & energy & $0.00$ & $0.00$ \\ \hline
\addlinespace
\multirow{6}{*}{NUTS} & \multirow{2}{*}{ReLU} & concrete & $0.14$ & $0.13$ \\
                      &                      & energy & $0.31$ & $0.13$ \\
                      & \multirow{2}{*}{SiLU} & concrete & $0.14$ & $0.10$ \\
                      &                      & energy & $0.28$ & $0.05$ \\
                      & \multirow{2}{*}{tanh} & concrete & $0.58$ & $0.25$ \\
                      &                      & energy & $0.67$ & $0.08$ \\
\bottomrule
\end{tabular}
\end{sc}
\end{small}
\end{center}
\end{table}

\clearpage

\subsection{Mulimodality of Posteriors}

In the following the experiments for the analyses reported in Figures~\ref{fig:disconnectedmodes}, \ref{fig:slopes} and \ref{fig:withinvariance} are described in detail. The architecture is always a network with six hidden layers of eight neurons each with tanh activation. Unit Gaussian priors are applied and the used NUTS sampler is configured with a 10,000 step warmup phase and 8,000 samples for each of the 12 chains. Again, the few chains (always $\leq 2$) that cannot beat a LM with respect to the RMSE are removed from the analysis. 

\begin{figure}[!ht]
    \centering
    \vskip 0.1in
    \includegraphics[width = 0.5\columnwidth]{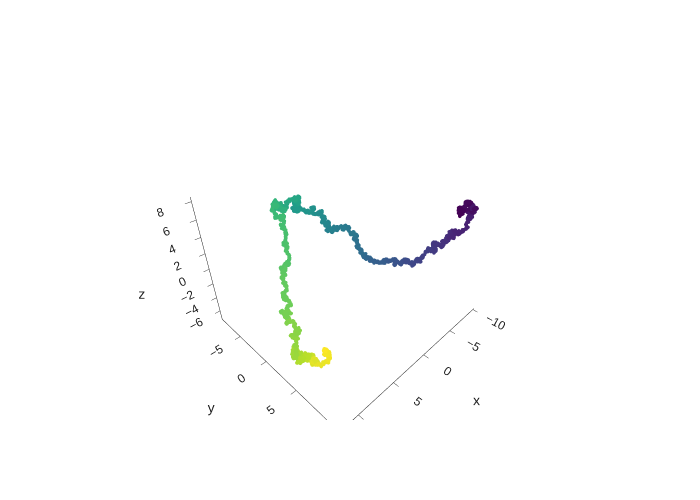}
    \vskip -0.2in
    \caption{Example sampling path of the first chain of the energy dataset projected in three dimensions using PCA. Sampler moved from dark to bright.}
    \label{fig:samp_path_energy}
\end{figure}

\begin{figure}[!ht]
    \centering
    \vskip 0.1in
    \includegraphics[width = 0.7\columnwidth]{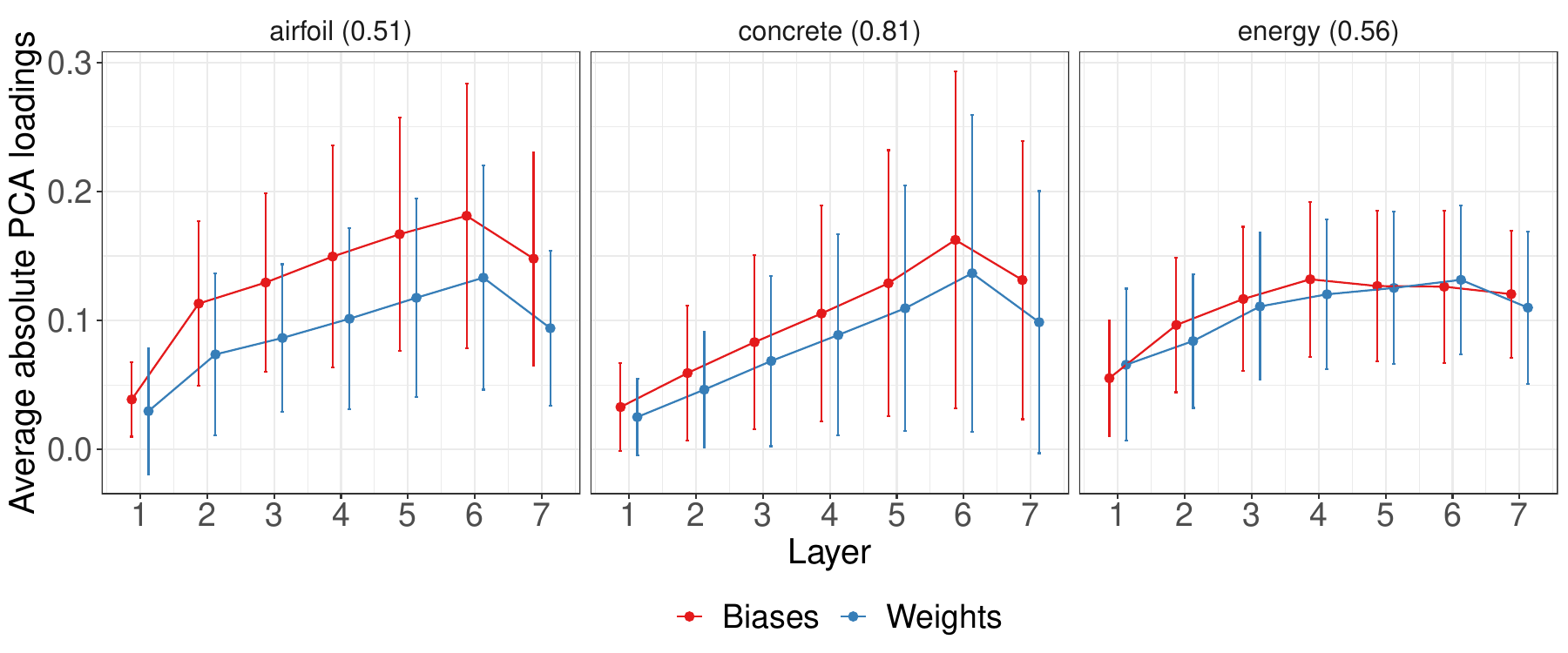}
    \vskip -0.2in
    \caption{Mean and standard deviation of the average absolute PCA loadings separated by layer (x-axis) for different data sets (columns) of a seven-layer BNN. Average explained variance by the first three principal components in brackets.}
    \label{fig:pcavariance}
\end{figure}

In addition to the analyses reported in Figures~\ref{fig:disconnectedmodes}, \ref{fig:slopes} and \ref{fig:withinvariance}, we investigate the sampling path for each chain. 
Here, instead of aggregating insights from marginal parameter traces, we take a joint view on the sampling path.
In order reduce the dimensionality of this problem, we perform a principal component analysis \citep[PCA;][]{WOLD198737} given $S$ posterior samples from the $k$-th chain $\bm{\Theta} = (\bm{\theta}^{(k, 1)}, \bm{\theta}^{(k, 2)}, \dots, \bm{\theta}^{(k, S)})^\top$ . Taking the three first principal components (each of dimension $|\bm{\theta}^{(k,s)}|$), we can project the sampling path into three dimensions. The first three principal components explain a substantial amount of variance of the sampling path ($>50\%$), which suggests that insights from the PCA-approximated path are indicative of the original, high-dimensional situation.
This is confirmed visually, e.g., in \cref{fig:samp_path_energy}, which is a representative example for the first chain of the energy dataset. Summing up the element-wise absolute values of the orthonormal vectors yields a proxy of how important the particular parameter is for the movement of the sampler. Applying this methodology and grouping the PCA factor loadings layer-wise, as well as averaging across experiments and chains, we obtain \cref{fig:pcavariance}. The same pattern as in \cref{fig:withinvariance} emerges, i.e., deeper layers are more important for the movement of the chain in the parameter space. 
We conclude from this that our findings pertain to a robust pattern.

\begin{figure}[!ht]
    \centering
    \begin{subfigure}{\textwidth}
        \includegraphics[width=\textwidth]{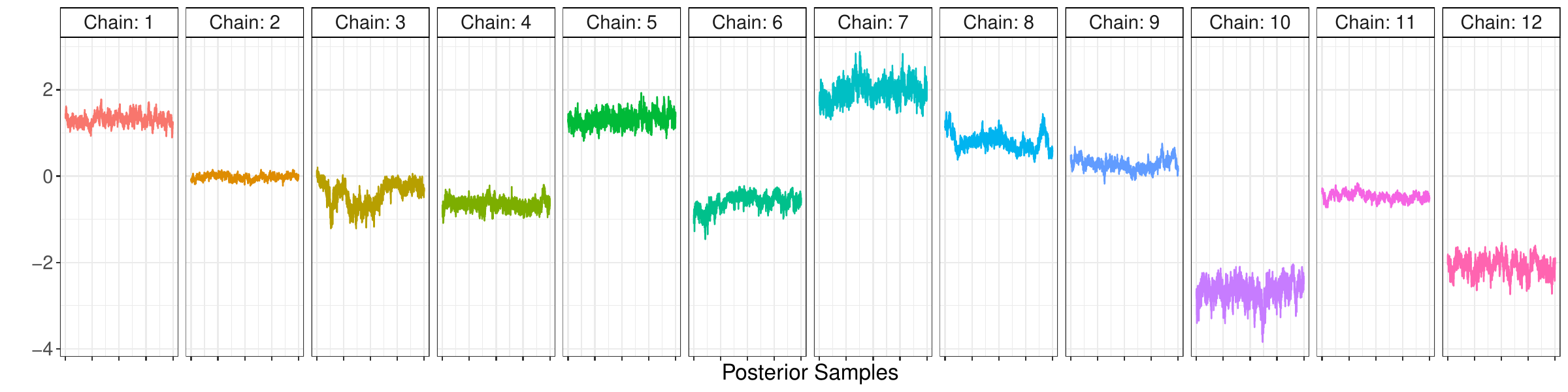}
        \caption{Random first layer weight}
    \end{subfigure}
    \begin{subfigure}{\textwidth}
        \includegraphics[width=\textwidth]{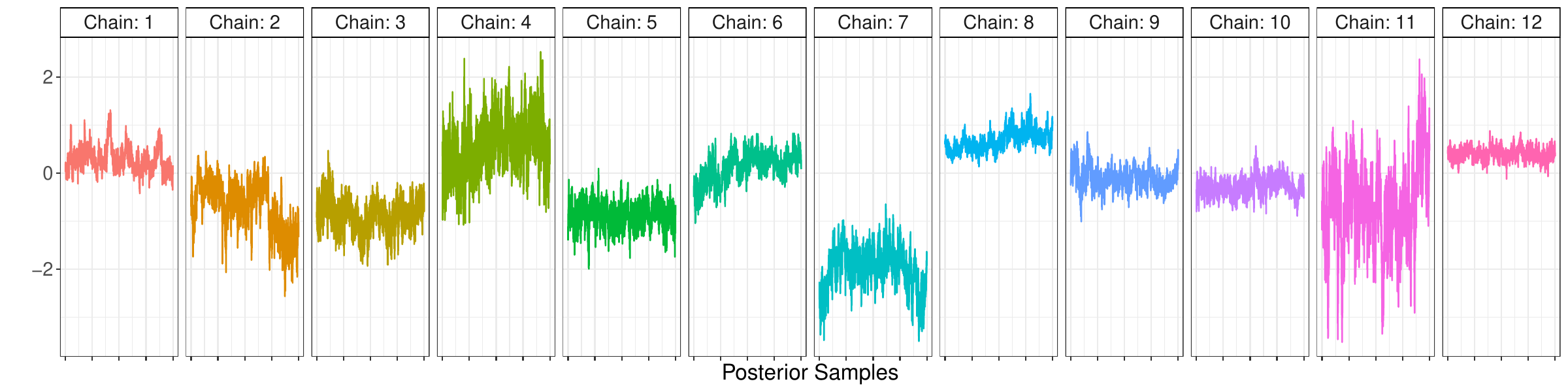}
        \caption{Random second layer weight}
    \end{subfigure}
    \begin{subfigure}{\textwidth}
        \includegraphics[width=\textwidth]{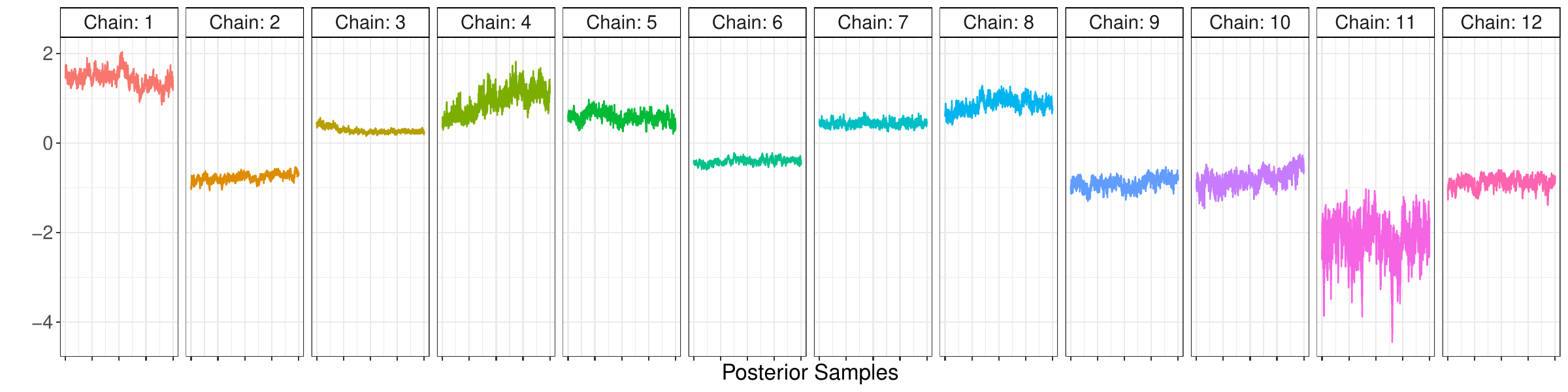}
        \caption{Random thrid (last) layer weight}
    \end{subfigure}
    \caption{
    Traceplots of a randomly selected parameter during the sampling phase of a NUTS sampler that performs well for most chains. The BNN has two hidden layer of 16 neurons each, uses unit Gaussian priors, tanh activation and is sampled for the \texttt{airfoil} data set with 10,000 NUTS warmup samples and 12 chains. The entire sampling phase, collecting 8,000 samples, is displayed (x-axis).
    }
    \label{fig:good_traces}
\end{figure}

\begin{figure}[!ht]
    \centering
    \begin{subfigure}{\textwidth}
        \includegraphics[width=\textwidth]{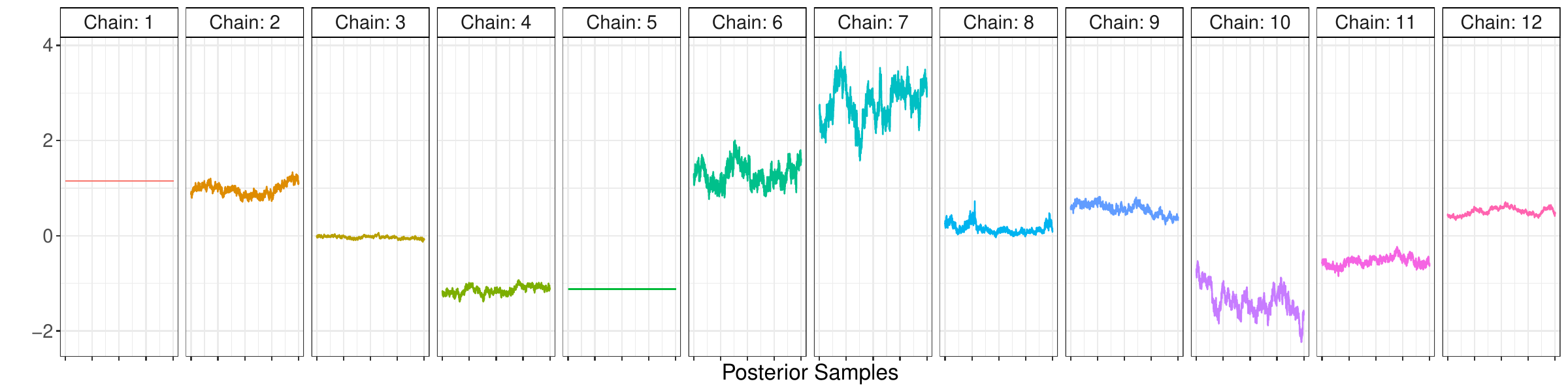}
        \caption{Random first layer weight}
    \end{subfigure}
    \begin{subfigure}{\textwidth}
        \includegraphics[width=\textwidth]{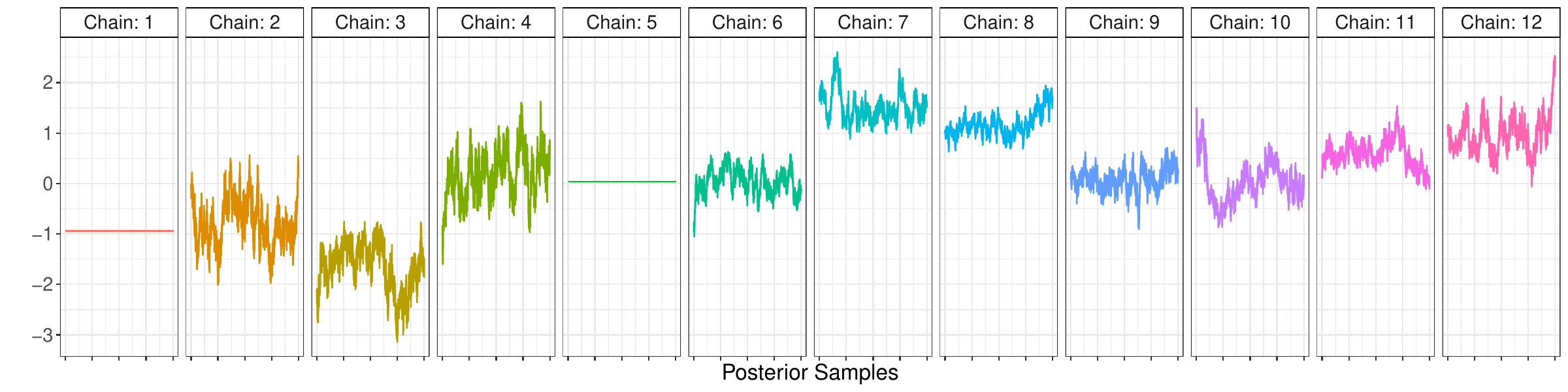}
        \caption{Random third layer weight}
    \end{subfigure}
    \begin{subfigure}{\textwidth}
        \includegraphics[width=\textwidth]{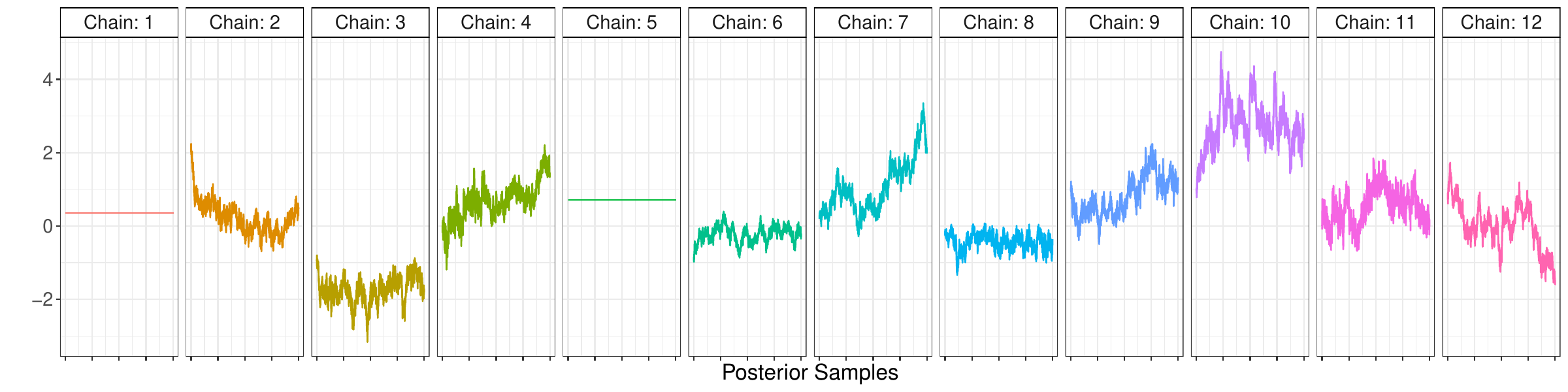}
        \caption{Random fifth layer weight}
    \end{subfigure}
    \begin{subfigure}{\textwidth}
        \includegraphics[width=\textwidth]{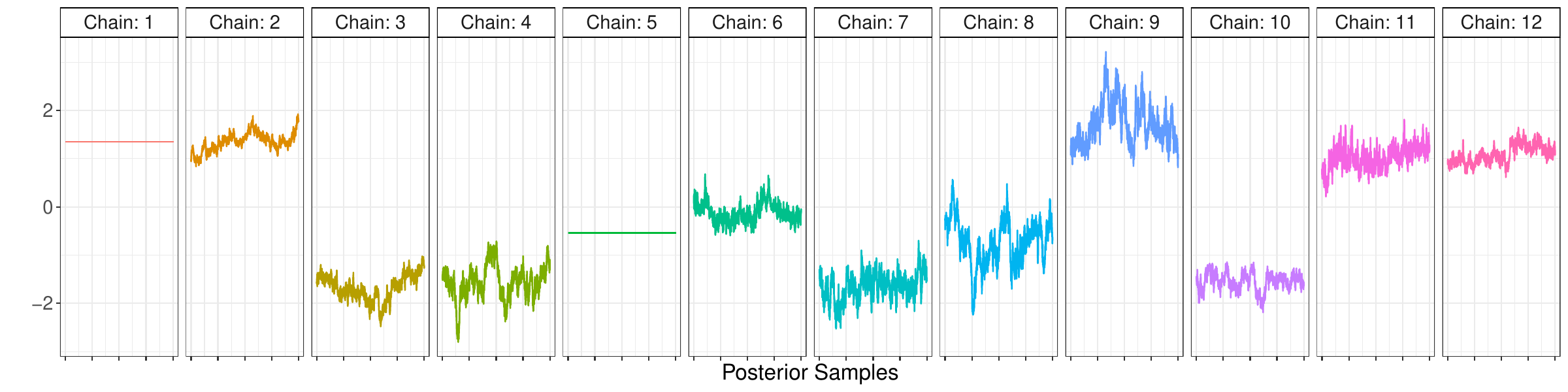}
        \caption{Random seventh (last) layer weight}
    \end{subfigure}
    \caption{Traceplots of a randomly selected parameter during the sampling phase of a NUTS sampler that performs well for most chains. The BNN has six hidden layer of eight neurons each, uses unit Gaussian priors, tanh activation and is sampled for the \texttt{airfoil} data set with 10,000 NUTS warmup samples and 12 chains. The entire sampling phase, collecting 8,000 samples, is displayed (x-axis).}
    \label{fig:deep_traces}
\end{figure}

The traceplots displayed in Figures \ref{fig:good_traces} and \ref{fig:deep_traces} are both examples of the fact that symmetry-induced multimodality exists and also illustrate how the within-chain variance increases in deeper layers. Moreover, Figure \ref{fig:deep_traces} shows how for the same setting some (in this case two) chains are ``dying", i.e., have a near-constant trace plot, while others explore the space in a meaningful way (implied by the good performance of the induced BNN).

\clearpage

\subsection{Practical SBI} \label{app:experiments_practical}

\textbf{Prior choice.} Here, we report detailed results for the influence of different prior choices.
Tables~\ref{tab:prior_comparison} and \ref{tab:prior_comparison2} show the test RMSE and the share of chains outperforming the LM, respectively, for the \texttt{airfoil} and \texttt{concrete} data sets. We employ NUTS with 10,000 warmup steps and 12 chains with 8,000 samples each. Each experiment is replicated on three random train-test splits of the data. All other tested configurations are displayed in the tables themselves.

\begin{table}[H] 
\caption{RMSE hold-out performance across various prior distributions and strengths (standard deviation in parentheses)}
\label{tab:prior_comparison}
   \vskip 0.15in
   \begin{center}
\begin{small}
\begin{sc}
\resizebox{1.0\columnwidth}{!}{
\begin{tabular}{l|l|rrrrrrrr}
\toprule
\multicolumn{2}{l}{} & \multicolumn{2}{c}{2} & \multicolumn{2}{c}{8} & \multicolumn{2}{c}{64} & \multicolumn{2}{c}{32-32-32} \\
\cmidrule(lr){3-4} \cmidrule(lr){5-6} \cmidrule(lr){7-8} \cmidrule(lr){9-10}
\multicolumn{2}{l}{data set} & ReLU & tanh & ReLU & tanh & ReLU & tanh & ReLU & tanh \\
\midrule\addlinespace[2.5pt]
airfoil & Laplace (0.1) & 0.5997 (0.0321) & 0.6072 (0.0287) & 0.3839 (0.0203) & 0.4413 (0.0534) & 0.3342 (0.0102) & 0.3557 (0.0109) & - & 0.2221 (0.0094) \\
 & Laplace (1) & 0.5876 (0.0328) & 0.5924 (0.0326) & 0.3349 (0.0107) & 0.3208 (0.0151) & 0.2288 (0.0066) & 0.2207 (0.0122) & - & 0.1609 (0.0189) \\
 & Laplace (10) & 0.5931 (0.0436) & 0.5916 (0.0297) & 0.3346 (0.0080) & 0.3289 (0.0022) & 0.2467 (0.0241) & 0.2209 (0.0121) & - & 0.2498 (0.0302) \\
 & Normal (0.1) & 0.6420 (0.0428) & 0.6703 (0.0188) & 0.5831 (0.0254) & 0.6475 (0.0295) & 0.5841 (0.0287) & 0.6558 (0.0276) & - & 0.6340 (0.0368) \\
 & Normal (1) & 0.5897 (0.0342) & 0.5949 (0.0303) & 0.3385 (0.0234) & 0.3333 (0.0029) & 0.2325 (0.0102) & 0.2282 (0.0069) & - & 0.1432 (0.0125) \\
 & Normal (10) & 0.5833 (0.0377) & 0.5982 (0.0312) & 0.3506 (0.0131) & 0.3468 (0.0157) & 0.2467 (0.0130) & 0.2189 (0.0122) & - & 0.2330 (0.0130) \\
\midrule\addlinespace[2.5pt]
concrete & Laplace (0.1) & 0.4323 (0.0026) & 0.4308 (0.0045) & 0.3615 (0.0070) & 0.3693 (0.0080) & - & 0.3610 (0.0124) & - & 0.2745 (0.0000)* \\
 & Laplace (1) & 0.4309 (0.0040) & 0.4271 (0.0019) & 0.3318 (0.0056) & 0.3241 (0.0047) & - & 0.2676 (0.0278) & - & 0.3045 (0.0119)\phantom{*} \\
 & Laplace (10) & 0.4323 (0.0030) & 0.4188 (0.0216) & 0.3334 (0.0063) & 0.3232 (0.0025) & - & 0.2803 (0.0199) & - & 0.3450 (0.0000)* \\
 & Normal (0.1) & 0.5012 (0.0208) & 0.5021 (0.0110) & 0.4439 (0.0079) & 0.4674 (0.0077) & - & 0.4718 (0.0041) & - & 0.4444 (0.0000)* \\
 & Normal (1) & 0.4323 (0.0046) & 0.4282 (0.0060) & 0.3317 (0.0057) & 0.3224 (0.0085) & - & 0.2844 (0.0262) & - & 0.3010 (0.0169)\phantom{*} \\
 & Normal (10) & 0.4321 (0.0020) & 0.4239 (0.0228) & 0.3406 (0.0107) & 0.3253 (0.0074) & - & 0.2762 (0.0249) & - & 0.3767 (0.0000)* \\
\bottomrule
 \multicolumn{1}{p{.12\textwidth}}{* Only one replication.}
\end{tabular}
}
\end{sc}
\end{small}
\end{center}
\end{table}

\begin{table}[!ht]
\caption{Proportion of better-than-LM chains across various prior distributions and strengths (standard deviation in parentheses)}
\label{tab:prior_comparison2}
   \vskip 0.15in
   \begin{center}
\begin{small}
\begin{sc}
\resizebox{1.0\columnwidth}{!}{
\begin{tabular}{l|l|rrrrrrrr}
\toprule
\multicolumn{2}{l}{} & \multicolumn{2}{c}{2} & \multicolumn{2}{c}{8} & \multicolumn{2}{c}{64} & \multicolumn{2}{c}{32-32-32} \\
\cmidrule(lr){3-4} \cmidrule(lr){5-6} \cmidrule(lr){7-8} \cmidrule(lr){9-10}
\multicolumn{2}{l}{data set} & ReLU & tanh & ReLU & tanh & ReLU & tanh & ReLU & tanh \\
\midrule\addlinespace[2.5pt]
airfoil & Laplace (0.1) & 0.97 (0.05) & 1.00 (0.00) & 0.89 (0.10) & 1.00 (0.00) & 0.36 (0.10) & 0.64 (0.13) & 0.00 (0.00) & 0.67 (0.25) \\
 & Laplace (1) & 0.97 (0.05) & 1.00 (0.00) & 0.89 (0.10) & 1.00 (0.00) & 0.36 (0.10) & 0.64 (0.13) & 0.00 (0.00) & 0.67 (0.25) \\
 & Laplace (10) & 0.97 (0.05) & 1.00 (0.00) & 0.89 (0.10) & 1.00 (0.00) & 0.33 (0.08) & 0.58 (0.17) & 0.00 (0.00) & 0.56 (0.21) \\
 & Normal (0.1) & 0.97 (0.05) & 1.00 (0.00) & 0.89 (0.10) & 1.00 (0.00) & 0.36 (0.10) & 0.64 (0.13) & 0.00 (0.00) & 0.64 (0.29) \\
 & Normal (1) & 0.97 (0.05) & 1.00 (0.00) & 0.89 (0.10) & 1.00 (0.00) & 0.36 (0.10) & 0.64 (0.13) & 0.00 (0.00) & 0.67 (0.25) \\
 & Normal (10) & 0.97 (0.05) & 1.00 (0.00) & 0.89 (0.10) & 1.00 (0.00) & 0.31 (0.05) & 0.64 (0.13) & 0.00 (0.00) & 0.58 (0.25) \\
\midrule\addlinespace[2.5pt]
concrete & Laplace (0.1) & 0.89 (0.10) & 1.00 (0.00) & 0.69 (0.10) & 1.00 (0.00) & 0.08 (0.08) & 0.56 (0.05) & 0.00 (0.00) & 0.33 (0.00)* \\
 & Laplace (1) & 0.86 (0.13) & 1.00 (0.00) & 0.69 (0.10) & 1.00 (0.00) & 0.08 (0.08) & 0.56 (0.05) & 0.00 (0.00) & 0.38 (0.06)\phantom{*} \\
 & Laplace (10) & 0.83 (0.08) & 1.00 (0.00) & 0.64 (0.13) & 1.00 (0.00) & 0.08 (0.08) & 0.56 (0.05) & 0.00 (0.00) & 0.33 (0.00)* \\
 & Normal (0.1) & 0.89 (0.10) & 1.00 (0.00) & 0.69 (0.10) & 1.00 (0.00) & 0.08 (0.08) & 0.56 (0.05) & 0.00 (0.00) & 0.33 (0.00)* \\
 & Normal (1) & 0.89 (0.10) & 1.00 (0.00) & 0.69 (0.10) & 1.00 (0.00) & 0.08 (0.08) & 0.56 (0.05) & 0.00 (0.00) & 0.38 (0.06)\phantom{*} \\
 & Normal (10) & 0.89 (0.10) & 1.00 (0.00) & 0.69 (0.10) & 0.97 (0.05) & 0.08 (0.08) & 0.56 (0.05) & 0.00 (0.00) & 0.33 (0.00)* \\
\bottomrule
 \multicolumn{1}{p{.12\textwidth}}{* Only one replication.}
\end{tabular}
}
\end{sc}
\end{small}
\end{center}
\end{table}

\begin{figure}[!ht]
    \centering
    \begin{subfigure}{0.35\textwidth}
        \includegraphics[width=\textwidth]{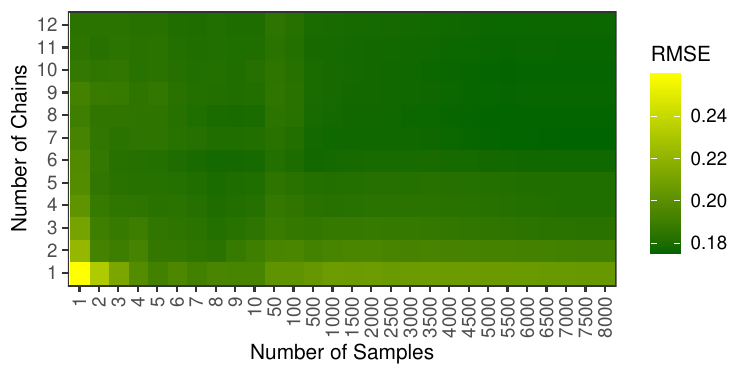}
        \caption{RMSE Airfoil}
    \end{subfigure}
    \begin{subfigure}{0.35\textwidth}
        \includegraphics[width=\textwidth]{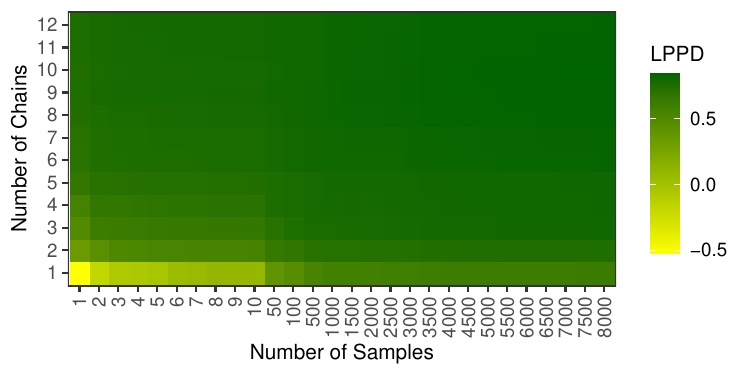}
        \caption{LPPD Airfoil}
    \end{subfigure}

    \begin{subfigure}{0.35\textwidth}
        \includegraphics[width=\textwidth]{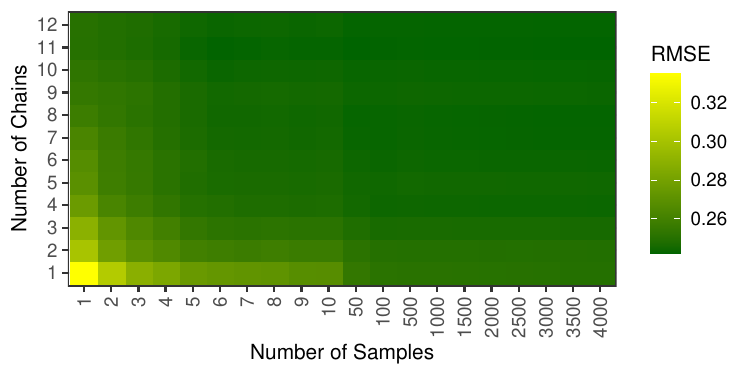}
        \caption{RMSE Bikesharing}
    \end{subfigure}
    \begin{subfigure}{0.35\textwidth}
        \includegraphics[width=\textwidth]{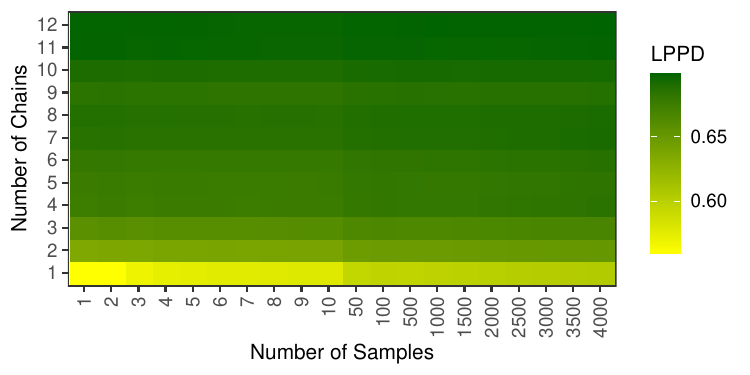}
        \caption{LPPD Bikesharing}
    \end{subfigure}

    \begin{subfigure}{0.35\textwidth}
        \includegraphics[width=\textwidth]{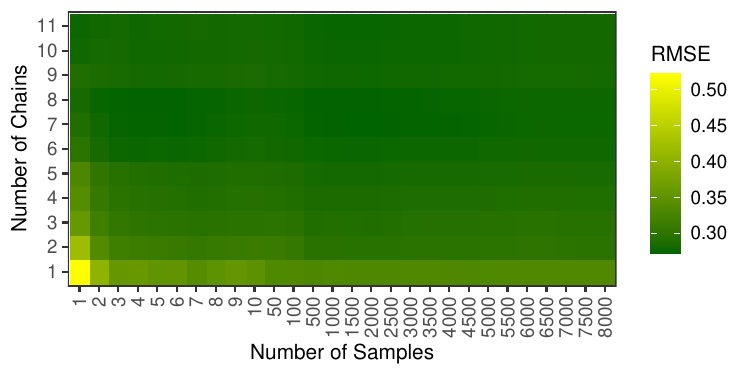}
        \caption{RMSE Concrete}
    \end{subfigure}
    \begin{subfigure}{0.35\textwidth}
        \includegraphics[width=\textwidth]{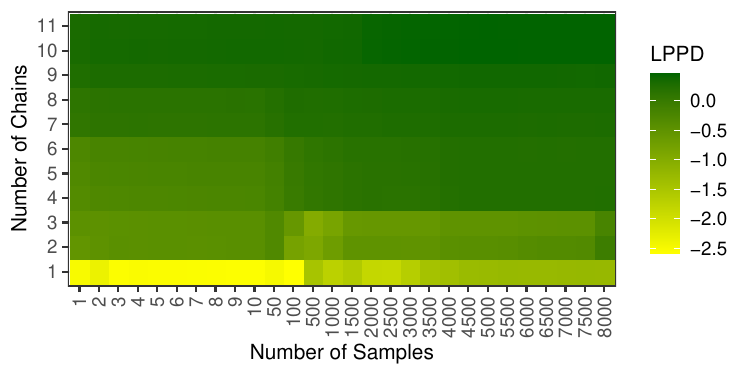}
        \caption{LPPD Concrete}
    \end{subfigure}

    \begin{subfigure}{0.35\textwidth}
        \includegraphics[width=\textwidth]{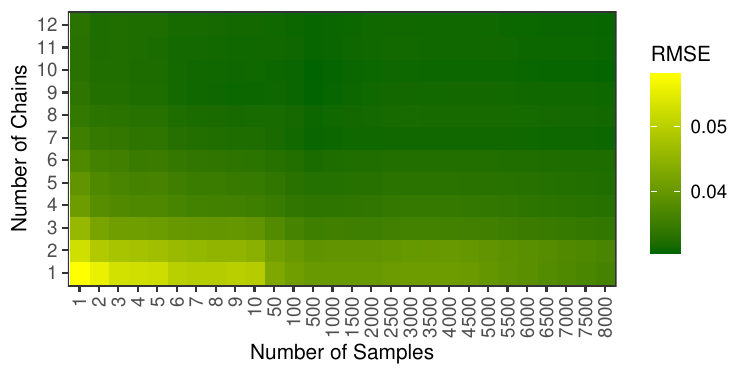}
        \caption{RMSE Energy}
    \end{subfigure}
    \begin{subfigure}{0.35\textwidth}
        \includegraphics[width=\textwidth]{icml2024/img/lppd_grid_energy.pdf}
        \caption{LPPD Energy}
    \end{subfigure}

    \begin{subfigure}{0.35\textwidth}
        \includegraphics[width=\textwidth]{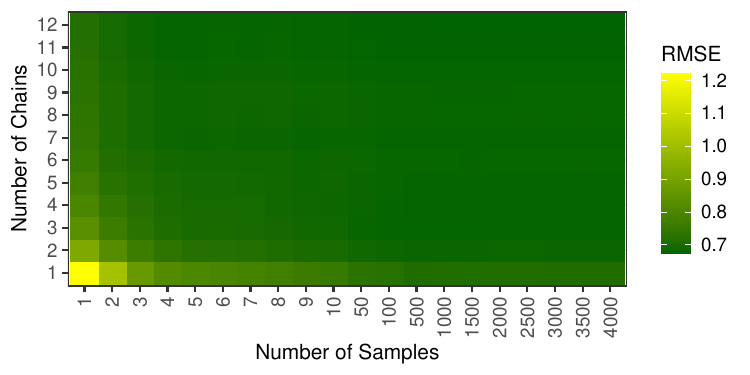}
        \caption{RMSE Protein}
    \end{subfigure}
    \begin{subfigure}{0.35\textwidth}
        \includegraphics[width=\textwidth]{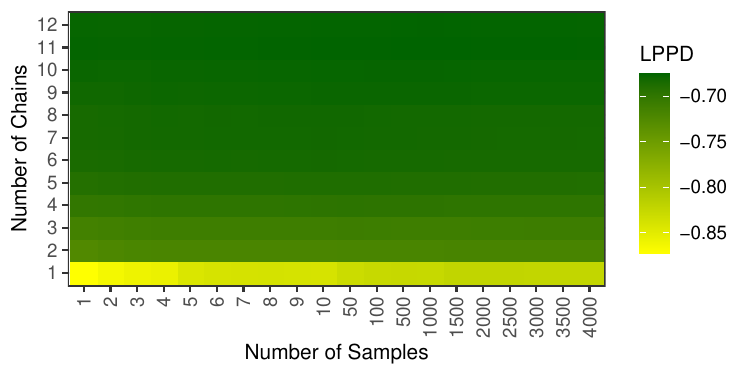}
        \caption{LPPD Protein}
    \end{subfigure}

    \begin{subfigure}{0.35\textwidth}
        \includegraphics[width=\textwidth]{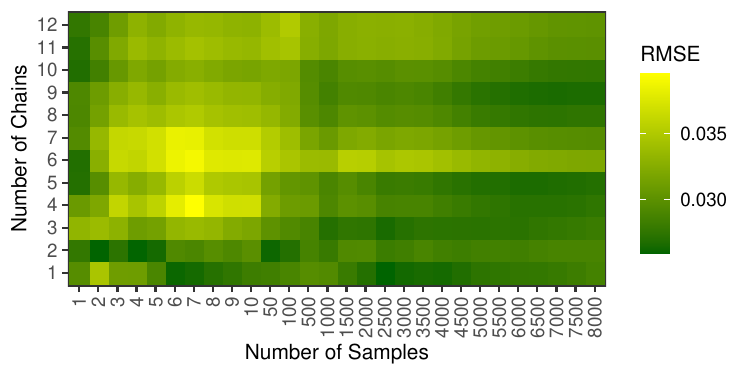}
        \caption{RMSE Yacht}
    \end{subfigure}
    \begin{subfigure}{0.35\textwidth}
        \includegraphics[width=\textwidth]{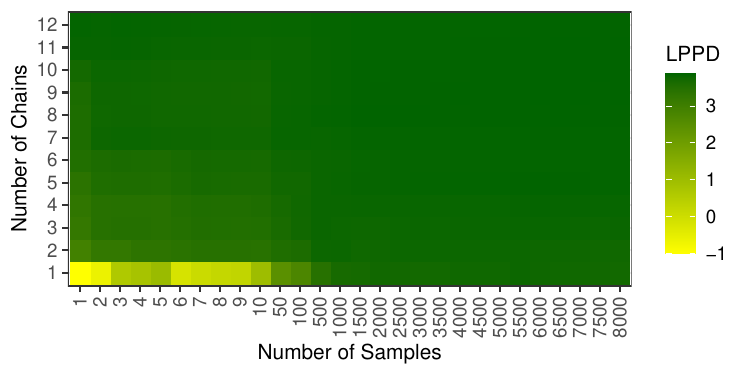}
        \caption{LPPD Yacht}
    \end{subfigure}
    \caption{Improvement across samples and chains of RMSE (left column) and LPPD (right column) for different data sets (rows).}
    \label{fig:grids}
\end{figure}

\textbf{Performance and uncertainty.} In accordance to \cref{fig:chainssamples}, \cref{fig:grids} visualizes the RMSE (left column) and LPPD (right column) performance for all data sets.
Except for the \texttt{yacht} data, we observe the same pattern of improvement along the directions of both chains and samples, where the former tends to have more effect. In all cases, the BNN consists of two hidden layers with 16 neurons each and tanh activation. NUTS with a warmup phase of 10,000 steps and 8,000 samples per chain (4,000 for the two larger data sets) as well as unit Gaussian priors are used. Only for \texttt{concrete}, one of the 12 chains is dropped as it underperformed the weak LM baseline.

\clearpage

\subsection{Dying Sampler and Deep Ensemble Initialization}

DEI-MCMC reported in  \cref{tab:bdes} uses 12 chains each with 8,000 samples (4,000 for the larger data sets) using the NUTS sampler with a short warmup phase of 100 steps. The table reports the first 10, 100 and 1,000 samples from these chains. Each chain's initial proposal is a parameter set of the converged DE member trained on the same architecture, which is a network with two hidden layers of 16 neurons each. The priors on the weights are again unit Gaussians, leading to a valid initialization of the chains. The activation is ReLU.

Moreover, the comparison presented in \cref{tab:9hidden} of the LPPD between DEI-MCMC and Deep Ensembles indicates a significant enhancement in uncertainty quantification also for deeper architectures.

\begin{table}[!h]
\begin{small}
\begin{sc}
\caption{Average LPPD values (± their standard deviations across replications) of a DE, and DEI-MCMC (i.e., a BNN
with DNN warm starts) using 1000 samples for four of the benchmark data sets. All networks have 9 hidden layers with 8 neurons each and ReLU activations. The best method per data set is highlighted in
bold}
\vskip 0.1in
\begin{center}

    \begin{tabular}{lrr}
    \toprule
    Dataset & DEI-MCMC & Deep Ensemble \\
    \midrule
    Airfoil & $\boldsymbol{0.5272} \pm 0.1210$ & $-0.1677 \pm 0.0766$ \\
    Bikesharing & $\boldsymbol{0.6799} \pm 0.0530$ & $0.4145 \pm 0.0422$ \\
    Concrete & $\boldsymbol{0.1683} \pm 0.0528$ & $-0.2794 \pm 0.0622$ \\
    Energy & $\boldsymbol{2.1147} \pm 0.0907$ & $0.8019 \pm 0.2090$ \\
    \bottomrule
    \end{tabular}%
  \label{tab:9hidden}%
\end{center}
\end{sc}
\end{small}
\end{table}%


\section{Experimental Settings and Further Details} \label{app:setup}

\paragraph{Benchmark data.}

\cref{tab:further} describes the data characteristics for our benchmark data. Note that we have normalized the features and outcome values for all data sets.

\begin{table}[ht]
\begin{small}
\begin{sc}
\begin{center}
\caption{Data set characteristics and references.} \label{tab:further}
\vskip 0.1in
\begin{tabular}{lrrl}
Data set & \# Obs. & \# Feat. & Reference \\ \hline 
Airfoil & 1503 & 5 &  \citet{Dua.2019}  \\
Bikesharing & 17379 & 13 & \citet{misc_bike_sharing_dataset_275}  \\
Concrete   & 1030 & 8 &  \citet{Yeh.1998} \\
Energy  & 768 & 8           &  \citet{Tsanas.2012} \\
Protein & 45730 &  9   & \citet{Dua.2019} \\
Yacht  & 308 & 6 & \citet{Ortigosa.2007,Dua.2019} \\ \hline
\end{tabular}
\end{center}
\end{sc}
\end{small}
\end{table}

\paragraph{Software.}
Our software is implemented in Python and mainly relies on the \texttt{jax} \citep{jax2018github}, \texttt{numpyro} \citep{phan2019composable} and \texttt{pytorch} \citep{paszke_2019_PyTorchImperative} libraries.
We further use \texttt{Docker} for a reproducible experimental setup.
Our code is available at \url{https://github.com/EmanuelSommer/bnn_connecting_the_dots}.

\paragraph{Baselines.}
For the DNN and DE baselines, we train all models full-batch for 5,000 epochs with a Gaussian negative log-likelihood loss function.
We use the \texttt{Adam} optimizer with an initial learning rate of $10^{-2}$ and weight decay equal to $10^{-2}$.
Note that the single-DNN performance is computed as the expected value of using one ensemble member, i.e., averaging the performance of individual members.

We also explored further settings in a grid search over the strength of weight decay ($10^{-3}$, $10^{-2}$, $10^{-1}$) as well as batch sizes (32, 64), allowing the training process to terminate if no improvement in validation loss was reached after 30 consecutive epochs.
For the latter, we set aside 10\% of the data as a validation set (amounting to a 70\%/10\%/20\% split for training/validation/test).
The results for training with tanh activation are displayed in \cref{tab:gridsearch}.
However, we could not detect any systematic improvement in RMSE performance (cf. \cref{tab:rmse_bench}) as compared to the above configurations.

\begin{table}[!ht]
    \centering
    \footnotesize
    \begin{sc}
    \begin{tabular}{lrrll}
        \toprule
        Data set & Weight decay & Batch Size & RMSE\\
        \midrule
        airfoil & 0.0001 & 32 & 0.2712 \\
        airfoil & 0.0001 & 64 & 0.3027 \\
        airfoil & 0.0010 & 32 & 0.2853 \\
        airfoil & 0.0010 & 64 & 0.2858 \\
        airfoil & 0.0100 & 32 & 0.2951 \\
        airfoil & 0.0100 & 64 & 0.3183 \\
        bikesharing & 0.0001 & 32 & 0.2779 \\
        bikesharing & 0.0001 & 64 & 0.2899 \\
        bikesharing & 0.0010 & 32 & 0.2832 \\
        bikesharing & 0.0010 & 64 & 0.2886 \\
        bikesharing & 0.0100 & 32 & 0.3289 \\
        bikesharing & 0.0100 & 64 & 0.3392 \\
        concrete & 0.0001 & 32 & 0.3572 \\
        concrete & 0.0001 & 64 & 0.3646 \\
        concrete & 0.0010 & 32 & 0.3566 \\
        concrete & 0.0010 & 64 & 0.3652 \\
        concrete & 0.0100 & 32 & 0.3592 \\
        concrete & 0.0100 & 64 & 0.3609 \\
        energy & 0.0001 & 32 & 0.2123 \\
        energy & 0.0001 & 64 & 0.2140 \\
        energy & 0.0010 & 32 & 0.2156 \\
        energy & 0.0010 & 64 & 0.2134 \\
        energy & 0.0100 & 32 & 0.2123 \\
        energy & 0.0100 & 64 & 0.2165 \\
        protein & 0.0001 & 32 & 0.7170 \\
        protein & 0.0001 & 64 & 0.7210 \\
        protein & 0.0010 & 32 & 0.7280 \\
        protein & 0.0010 & 64 & 0.7281 \\
        protein & 0.0100 & 32 & 0.7943 \\
        protein & 0.0100 & 64 & 0.7937 \\
        yacht & 0.0001 & 32 & 0.6096 \\
        yacht & 0.0001 & 64 & 0.6072 \\
        yacht & 0.0010 & 32 & 0.6188 \\
        yacht & 0.0010 & 64 & 0.5277 \\
        yacht & 0.0100 & 32 & 0.5355 \\
        yacht & 0.0100 & 64 & 0.6066 \\
        \bottomrule
    \end{tabular}
    \caption{RMSE performance for non-Bayesian DE with 12 members across different settings of weight decay and batch size. All networks have two hidden layers of 16 neurons each and tanh activation.}
    \label{tab:gridsearch}
    \end{sc}
\end{table}

\paragraph{Computing environment and times.} The experiments were run on 4 CPU instances with 32 cores each and 64GB RAM. Sampling 12 chains for most experiments allowed to parallelize the sampling such that at all times 2 experiments with 12 chains each can be run. The sampling with NUTS of 12 chains, 10,000 warmup steps and 8,000 samples (4,000 for the larger data sets) required on average three hours for the smaller data sets, and 30 hours for the larger ones on a two-hidden-layer network with 16 neurons in each hidden layer and tanh activation. This results in about half a second and 8 seconds for 12 parallel samples for the small and large data sets, respectively. Decreasing the number of warmup samples as proposed in DEI-MCMC reduces the number of required samples per chain by more than half, resulting in much faster model fits.

\end{document}